\def\doctype{2}
\def\year{2021}\relax
\documentclass[letterpaper]{article} % DO NOT CHANGE THIS
\usepackage{aaai21}  % DO NOT CHANGE THIS
\usepackage{times}  % DO NOT CHANGE THIS
\usepackage{helvet} % DO NOT CHANGE THIS
\usepackage{courier}  % DO NOT CHANGE THIS
\usepackage[hyphens]{url}  % DO NOT CHANGE THIS
\usepackage{graphicx} % DO NOT CHANGE THIS
\usepackage{graphicx}  % DO NOT CHANGE THIS

\usepackage{natbib}  % DO NOT CHANGE THIS AND DO NOT ADD ANY OPTIONS TO IT

\usepackage{caption} % DO NOT CHANGE THIS AND DO NOT ADD ANY OPTIONS TO IT
\usepackage[switch]{lineno}
\usepackage{array}
\usepackage{floatrow}
\usepackage{cite}
\usepackage{microtype}
\usepackage{inconsolata}
\usepackage{tipa}
\usepackage{mathtools}
\restylefloat{figure}
\usepackage{amsmath}
\usepackage{amssymb}  
\usepackage{mathtools}
\usepackage{fancyhdr}
\usepackage{kantlipsum}
\usepackage{eso-pic}
\usepackage{nomencl}
\usepackage{etoolbox}
\usepackage{balance}
\usepackage{dsfont}
\usepackage{tikz}
\usepackage{eucal}
\usepackage{nccmath}
\tikzset{>=latex}
\usetikzlibrary{automata,positioning}
\usepackage[algo2e,linesnumbered,ruled,lined]{algorithm2e}
\usepackage{listings}
\usepackage{multirow}
\usepackage{pgf}
\usepackage{pgfplots}
\pgfplotsset{compat=1.14}
\usetikzlibrary{shapes.geometric,arrows,automata}
\usepackage{subcaption}
\newtheorem{definition}{\textbf{Definition}}[section]
\newtheorem{theorem}{\textbf{Theorem}}[section]

\newcommand{\NewEdit}{\textcolor{black}}

\if\doctype0
\author{Submission 8323}
\fi
\if\doctype2
\title{DeepSynth: Automata Synthesis for Automatic Task Segmentation \\ in Deep Reinforcement Learning \\ {}[Extended Version]}
\else
\title{DeepSynth: Automata Synthesis for Automatic Task Segmentation \\ in Deep Reinforcement Learning}
\fi
\if\doctype2
\nocopyright
\author {
	Mohammadhosein Hasanbeig,
	Natasha Yogananda Jeppu, \\
	Alessandro Abate, 
	Tom Melham,
	Daniel Kroening\thanks{The work in this paper was done prior to joining Amazon.} \\
}
\affiliations {
	Computer Science Department, University of Oxford, Parks Road, Oxford, United Kingdom, OX1 3QD \\
	\{hosein.hasanbeig, natasha.yogananda.jeppu, alessandro.abate, tom.melham\}@cs.ox.ac.uk
        and daniel.kroening@magd.ox.ac.uk
}
\fi
\if\doctype3
\author {
	Mohammadhosein Hasanbeig,
	Natasha Yogananda Jeppu, \\
	Alessandro Abate, 
	Tom Melham,
	Daniel Kroening\thanks{The work in this paper was done prior to joining Amazon.} \\
}
\affiliations {
	Computer Science Department, University of Oxford, Parks Road, Oxford, United Kingdom, OX1 3QD \\
	\{hosein.hasanbeig, natasha.yogananda.jeppu, alessandro.abate, tom.melham\}@cs.ox.ac.uk
        and daniel.kroening@magd.ox.ac.uk
}
\fi
\begin{document}
\if\doctype1
\appendix
\section*{Appendix}

\if\doctype1
\section{Background on RL}
\begin{definition} [Expected Discounted Return] 
	\label{expectedut}
	For a policy $\pi$ on an MDP $\mathfrak{M}$, the expected discounted return for a Markovian reward $R$ is defined as~\cite{sutton}:
	\begin{equation}
	\label{expecteduteq}
	{U}^{\pi}(s)=\mathds{E}^{\pi} [\sum\limits_{n=0}^{\infty} \gamma^n~ r_n|s_0=s],
	\end{equation}
	where $\mathds{E}^{\pi} [\cdot]$ denotes the expected value given that the agent follows policy $\pi$, $[0,1)$ ($\gamma\in [0,1]$ when episodic) is a discount factor.
	%, $R:\mathcal{S}\times\mathcal{A}\rightarrow \mathds{R}$ is the reward, and $s_0,...,s_n$ is the sequence of states generated by policy $\pi$ up to time step $n$. 
\end{definition}

% \begin{definition}[Optimal Policy]
% 	\label{optimal_policy}
% 	Optimal policy $\pi^*$ is the best policy such that the expected discounted reward is maximized, i.e:
% 	$$
% 	\pi^*(s)=\arg\sup\limits_{\pi \in \mathcal{D}}~ {U}^{\pi}(s),
% 	$$
% 	where $\mathcal{D}$ is the set of all stationary deterministic policies over the state space $S$.
% \end{definition}
%
The expected return is also known as the \emph{value function} in the literature.  For any state-action pair $(s,a)$ we can also define an
action-value function that assigns a quantitative measure $Q:S\times
A\rightarrow \mathds{R}$ as follows:
\begin{equation}
\label{q_value}
{Q}^{\pi}(s,a)=\mathds{E}^{\pi} [\sum\limits_{n=0}^{\infty} \gamma^n~ r_n|s_0=s, a_0=a].
\end{equation}
Q-Learning (QL)~\cite{watkins} 
%is arguably the most extensively used RL algorithm. It 
employs the action-value function and updates state-action pair values upon
visitation, as shown in \eqref{ql_update_rule}.  QL is \emph{off-policy}, which
means that $\pi$ has no effect on the convergence of the Q-function,
as long as every state-action pair is visited infinitely many times.
Thus, for simplicity, we may use~$Q$ only as
\begin{equation}
\label{ql_update_rule}
Q(s,a)\!\leftarrow\!Q(s,a)+\alpha
[R(s,a)+\gamma \max\limits_{a' \in \mathcal{A}}(Q(s',a'))-Q(s,a)],
\end{equation}
where $0<\alpha\leq 1 $ is the learning rate, $\gamma$ is the discount factor,
and $s'$ is the state reached after performing action~$a$.  The learning
rate and discount factor in general can be state-dependant.  Under mild
assumptions, QL converges to a unique limit $Q^*$, as long as every state
action pair is visited infinitely many times~\cite{watkins}.  Once QL
converges, an optimal policy can be distilled from the action-value function:
$$
\pi^*(s)=\arg\max\limits_{a \in \mathcal{A}}~Q^*(s,a),
$$
where $ \pi^* $ is the same optimal policy that can be alternatively generated
with Bellman iterations~\cite{NDP} if the MDP was fully known, maximising the expected return
\eqref{expecteduteq} at any given state.  Thus, the main goal is to
synthesise $ \pi^* $ when the MDP is essentially a black box. We denote a non-Markovian optimal policy by $\widehat{\pi}^*$ which optimises a memory-dependent Q-function $\widehat{Q}^*$. 

In many problems, the MDP can have a continuous or large state space, and thus
the recursion in \eqref{ql_update_rule} has to be approximated by
parameterising $Q$ using a parameter set $\theta^Q$. The parameters are updated by minimizing the following loss
function~\cite{nfq}:
\begin{equation}\label{loss_function}
\mathfrak{L}(\theta^Q)= \mathds{E}_{s \sim {pr}^\beta}[(Q(s, a|\theta^Q)-y)^2], 
\end{equation}
where ${pr}^\beta$ is the probability distribution of state visit over
$\mathcal{S}$ under an arbitrary stochastic policy $\beta$, and $$y =
R(s,a)+\gamma \max\limits_{a'}Q(s',a'|\theta^Q).$$
The function $Q$ can then be
approximated via a deep neural network architecture where the parameter set
$\theta^Q$ represents the weights of the neural network.
\fi
\section{The Automata Synthesis Framework vs.~State Merge}\label{sec:lrmvssynth}

In this section we compare the automata synthesis algorithm we use to
algorithms based on state merging.  State merge is the
established approach for model generation from traces.  Traces are first
converted into a Prefix Tree Acceptor (PTA).  Model inference techniques are
then used to identify pairs of equivalent states to be merged in the
hypothesis model.  Starting from the traditional kTails~\cite{ktails}
algorithm for state merging, several alternatives to determine state
equivalence have been proposed over the years~\cite{state_merge}.  For our
experiment we used the MINT (Model INference Technique)~\cite{mint} tool,
which implements different variants of the state merge algorithm, including
data classifiers~\cite{Walkinshaw2016} to check state equivalence for
merging.

We generated models using MINT for all seven tasks for the Minecraft
environment and explored different tool configurations to generate a model
that best fits the input trace.  We observed that although MINT is faster,
the automata generated by the tool are either too big (meaning they have a
large number of states) or are over-generalised (in extreme cases they only
have a single state), depending on the tool configurations.  For instance,
the smallest model that best fits the traces for Task 5 includes 49 states
(Fig.~\ref{task5}), and 14 states for Task 6 (Fig.~\ref{task6}).  Here, the
`start' label signifies the beginning of a new trace obtained from another
instance of random exploration.  DeepSynth requires automata that are
succinct and accurately represent sequential behaviour observed in the
exploration trace to ensure fast and efficient learning.  Since state merge
algorithms do not produce the most succinct model that fits a given trace,
we prefer using the automata synthesis algorithm described in
Section~\ref{sec:synth}.

\begin{figure*}[btp]\centering
	\includegraphics[width=\textwidth]{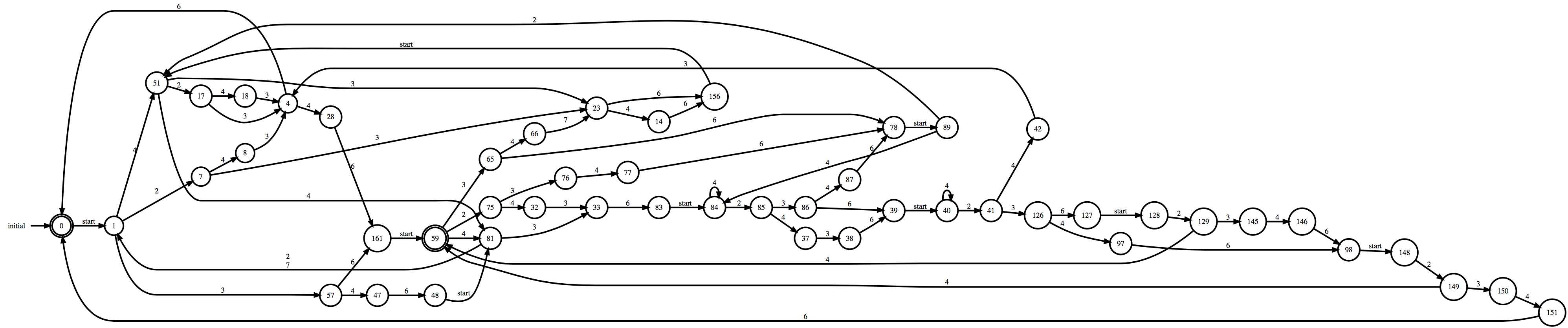}
	\caption{Best fit model for Task 5 generated by the MINT tool.}
	\label{task5}
\end{figure*}

\begin{figure*}[btp]\centering
	\includegraphics[width=\textwidth]{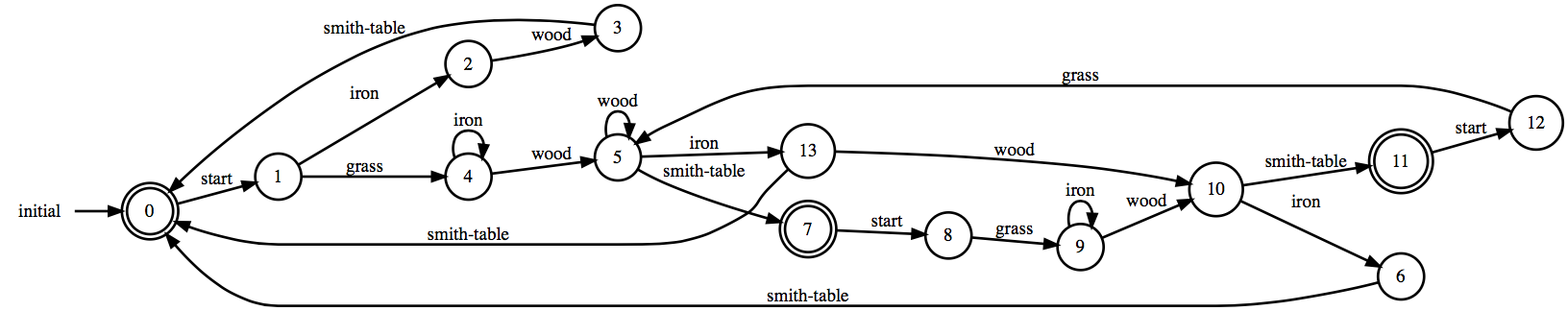}
	\caption{Best fit model for Task 6 generated by the MINT tool.}
	\label{task6}
\end{figure*}

\pgfplotsset{width = 9cm,compat=1.8,every axis/.append style={
		xlabel={Trace Length},          % default put x on x-axis
		%label style={font=\scriptsize},
		scaled x ticks=false,
		x tick label style = { text width = 1.7cm, align = center} ,
		mark size=1pt
}}
\begin{figure*}[btp]
	\begin{minipage}{0.3\columnwidth}
		\centering
		\scriptsize
		\begin{tikzpicture}
		\begin{axis}[scale=0.6,
		xmin=0, xmax=9000,
		ymin=0, ymax=2.0,
		legend style={at={(0.4,0.95)},
			anchor=north,legend columns=-1},
		ylabel={Runtime (min)},
		ytick = {0,0.2,0.4,0.6,0.8,1,1.2,1.4,1.6,1.8,2,2.2}
		]
		
		\addplot coordinates {(228,0.087766882)
			(449,0.103931801)
			(620,0.127841866)
			(889,0.123906235)
			(1066,0.140644634)
			(1281,0.154280917)
			(1458,0.19258105)
			(1741,0.183081547)
			(2012,0.184001696)
			(2227,0.145450934)
			(2458,0.147593451)
			(2657,0.141112665)
			(2898,0.15106045)
			(3129,0.131965951)
			(3346,0.140558966)
			(3543,0.147761635)
			(3779,0.144028866)
			(3958,0.133367936)
			(4133,0.160394204)
			(4365,0.167770513)
			(4564,0.148206266)
			(4782,0.155960333)
			(5029,0.170788435)
			(5308,0.166628035)
			(5479,0.157477065)
			(5748,0.155062751)
			(5932,0.161589833)
			(6161,0.138700855)
			(6396,0.144631986)
			(6607,0.134129512)
			(6838,0.135662834)
			(7086,0.134460231)
			(7343,0.129717549)
			(7606,0.137405487)
			(7793,0.170044684)
			(7952,0.134234265)
			(8134,0.13628362)
			(8345,0.136256437)
			(8544,0.173528687)
			(8707,0.180989397)};
		\addplot coordinates {(228,0.15)
			(449,0.17)
			(620,0.21)
			(889,0.23)
			(1066,0.26)
			(1281,0.28)
			(1458,0.29)
			(1741,0.32)
			(2012,0.37)
			(2227,0.37)
			(2458,0.4)
			(2657,0.43)
			(2898,0.44)
			(3129,0.48)
			(3346,0.51)
			(3543,0.5)
			(3779,0.53)
			(3958,0.54)
			(4133,0.6)
			(4365,0.59)
			(4564,0.61)
			(4782,0.64)
			(5029,0.68)
			(5308,0.68)
			(5479,0.7)
			(5748,0.74)
			(5932,0.73)
			(6161,0.75)
			(6396,0.79)
			(6607,0.79)
			(6838,0.82)
			(7086,0.86)
			(7343,0.91)
			(7606,0.94)
			(7793,0.95)
			(7952,0.94)
			(8134,0.96)
			(8345,0.99)
			(8544,0.96)
			(8707,1.01)};
		\legend{Synth,LRM}
		\end{axis}
		\end{tikzpicture}
		\subcaption{Cookie Domain}
	\end{minipage}
	\quad
	\begin{minipage}{0.3\columnwidth}
		\centering
		\scriptsize
		\begin{tikzpicture}
		\begin{axis}[scale=0.6,
		xmin=0, xmax=6000,
		ymin=0, ymax=2.0,
		legend style={at={(0.4,0.95)},
			anchor=north,legend columns=-1},
		ylabel={Runtime (min)},
		ytick = {0,0.2,0.4,0.6,0.8,1,1.2,1.4,1.6,1.8,2,2.2}
		]
		
		\addplot coordinates {(37,0.048800468)
			(129,0.156992233)
			(267,0.206101632)
			(377,0.239952501)
			(485,0.301804133)
			(629,0.331270635)
			(813,0.355126965)
			(905,0.35914677)
			(1085,0.364988101)
			(1229,0.363674684)
			(1369,0.398797997)
			(1535,0.446300117)
			(1690,0.447166634)
			(1869,0.434898233)
			(1971,0.446195332)
			(2087,0.450742368)
			(2199,0.4519798)
			(2283,0.441446082)
			(2377,0.446357361)
			(2505,0.503604186)
			(2647,0.477986666)
			(2827,0.429873498)
			(2909,0.456357733)
			(3013,0.39833333)
			(3109,0.493936284)
			(3245,0.458187226)
			(3385,0.460330049)
			(3521,0.448055836)
			(3689,0.455454934)
			(3803,0.39092445)
			(3957,0.406011951)
			(4081,0.420050748)
			(4249,0.455326235)
			(4383,0.462840037)
			(4513,0.464638555)
			(4679,0.453105732)
			(4819,0.434624036)
			(4921,0.439028903)
			(5007,0.395128894)
			(5111,0.390294619)
			(5201,0.393531048)
			(5398,0.401812681)};
		\addplot coordinates {(37,0.16)
			(129,0.16)
			(267,0.2)
			(377,0.22)
			(485,0.26)
			(629,0.3)
			(813,0.34)
			(905,0.37)
			(1085,0.41)
			(1229,0.44)
			(1369,0.47)
			(1535,0.5)
			(1690,0.54)
			(1869,0.56)
			(1971,0.6)
			(2087,0.63)
			(2199,0.68)
			(2283,0.66)
			(2377,0.71)
			(2505,0.72)
			(2647,0.72)
			(2827,0.76)
			(2909,0.8)
			(3013,0.81)
			(3109,0.82)
			(3245,0.87)
			(3385,0.92)
			(3521,0.89)
			(3689,0.97)
			(3803,0.95)
			(3957,1.02)
			(4081,1.01)
			(4249,1.04)
			(4383,1.12)
			(4513,1.11)
			(4679,1.17)
			(4819,1.21)
			(4921,1.17)
			(5007,1.16)
			(5111,1.23)
			(5201,1.36)
			(5398,1.31)};
		\legend{ Synth,LRM}
		\end{axis}
		
		\end{tikzpicture}
		\subcaption{Symbol Domain}
	\end{minipage}
	\quad
	\begin{minipage}{0.3\columnwidth}
		\centering
		\scriptsize
		\begin{tikzpicture}
		\begin{axis}[scale=0.6,
		xmin=0, xmax=8000,
		ymin=0, ymax=2.0,
		legend style={at={(0.4,0.95)},
			anchor=north,legend columns=-1},
		ylabel={Runtime (min)},
		ytick = {0,0.2,0.4,0.6,0.8,1,1.2,1.4,1.6,1.8,2,2.2}
		]
		
		\addplot coordinates {(121,0.396799751)
			(280,0.557217737)
			(350,0.652568066)
			(540,0.583352745)
			(692,0.556553268)
			(816,0.558721916)
			(934,0.620674515)
			(1046,0.543628919)
			(1237,0.61729635)
			(1427,0.6806928)
			(1501,0.660136918)
			(1679,0.627433721)
			(1749,0.632444215)
			(1817,0.589973565)
			(1949,0.633822453)
			(2051,0.606950434)
			(2266,0.681478703)
			(2436,0.74896034)
			(2542,0.605576086)
			(2700,0.605466497)
			(2914,0.579354799)
			(3012,0.50577658)
			(3206,0.565842581)
			(3374,0.609957846)
			(3549,0.544571849)
			(3683,0.567325819)
			(3801,0.572060466)
			(4006,0.596386067)
			(4166,0.518547646)
			(4325,0.537743179)
			(4482,0.586543381)
			(4616,0.591015768)
			(4736,0.596785434)
			(4818,0.592066987)
			(4943,0.595160651)
			(5055,0.619921354)
			(5231,0.598163434)
			(5399,0.586950564)
			(5611,0.599764466)
			(5798,0.581244687)
			(5918,0.553741264)
			(6057,0.612412616)
			(6261,0.64431682)
			(6403,0.752674635)
			(6567,0.651543387)
			(6767,0.70214895)
			(6937,0.708881783)
			(7079,0.547331734)};
		\addplot coordinates {(121,0.18)
			(280,0.24)
			(350,0.25)
			(540,0.29)
			(692,0.32)
			(816,0.35)
			(934,0.37)
			(1046,0.39)
			(1237,0.44)
			(1427,0.47)
			(1501,0.48)
			(1679,0.51)
			(1749,0.53)
			(1817,0.55)
			(1949,0.58)
			(2051,0.62)
			(2266,0.64)
			(2436,0.66)
			(2542,0.68)
			(2700,0.72)
			(2914,0.75)
			(3012,0.77)
			(3206,0.79)
			(3374,0.83)
			(3549,0.85)
			(3683,0.87)
			(3801,0.94)
			(4006,0.94)
			(4166,0.95)
			(4325,1)
			(4482,1.03)
			(4616,1.07)
			(4736,1.1)
			(4818,1.11)
			(4943,1.13)
			(5055,1.14)
			(5231,1.18)
			(5399,1.21)
			(5611,1.26)
			(5798,1.29)
			(5918,1.35)
			(6057,1.32)
			(6261,1.36)
			(6403,1.39)
			(6567,1.46)
			(6767,1.46)
			(6937,1.52)
			(7079,1.57)};
		\legend{ Synth,LRM}
		\end{axis}
		
		\end{tikzpicture}
		\subcaption{2-Keys Domain}
	\end{minipage}
	\caption{Runtime comparison of Tabu search vs. the Synth algorithm.}
	\label{fig:lrmvssynth}
\end{figure*}

\section{The Automata Synthesis Framework vs.~Learning Reward Machines
Using Tabu Search}\label{sec:synth_vs_tabu}

The line of work presented in~\cite{toronto}, referred to as LRM here, uses
reward machines learnt from trace data to guide the learning process and is
very closely related to our work.  In this section we compare the two
methodologies.  The automata synthesis algorithm (Synth) used in this paper
implements online learning.  Synth converts the model learning problem into
a Boolean Satisfiability (SAT) problem and uses a SAT solver to generate
models.  SAT solvers, in turn, use a backtracking search algorithm,
DPLL~\cite{DPLL}, to look for a satisfying assignment to the variables in
the Boolean formula.  LRM, on the other hand, uses Tabu search to explore
and find reward machines from trace data.  DPLL is complete and explores the
entire search space, as opposed to Tabu search, which is
local~\cite{DPLLvsTabu}.  Model completeness and accuracy is important in
DeepSynth as the agent performance relies on the automaton to guide
learning.  The automata generated by Synth are complete in the sense that
they capture all sequential behaviour seen in the trace.

DeepSynth uses traces from random exploration to generate an automaton.  In
each episode sequential behaviours observed over iterations of exploration
are incorporated into the automaton without changing the structure of the
automaton inferred in previous episodes.  More specifically, when the
automata is updated, it will always be just a superset of the previous
automata.  As such, the Q-networks can continue to be used and updated after
a automata is updated.  Namely, the only required change is that more
Q-networks need to be added and old Q networks can be reused.  This is not
the case in LRM where Q-networks must be reset and relearned from scratch
each time the automata changes.  Also, in contrast to LRM (generated using
Tabu search) Synth does not require an initial model.

Note that in DeepSynth we only care about positive trace samples
\cite{Gold1978ComplexityOA}.  This essentially allows us to add new traces
incrementally to the existing automaton.  Thus, whenever the agent discovers
a new label, it is added to $\Sigma_\mathfrak{A}$ and the associated trace
is considered as the new accepting trace.  So as the agent explores the
environment over multiple iterations of DeepSynth, we get longer trace
sequences with newly discovered labels.  These new labels are added
incrementally to the partial model generated in previous iterations.  Thus,
the model generated represents the sequence of labels (or intrinsic goals)
that composes extrinsic rewarding tasks.  This allows to incrementally build
up the DFA from intrinsically accepting traces to extrinsically accepting
traces as in Fig \ref{fig:montezuma_dfa}.

%For the cookie example in Toro Icarte et al we performed a runtime comparison of the two approaches, generating automata from traces of increasing lengths

The graph plots in Fig.\ref{fig:lrmvssynth} provide a runtime comparison of
the Synth algorithm and LRM for increasing trace length in different
benchmarks.  To ensure a fair comparison, we use the same three examples
used to benchmark the LRM approach here: Cookie, 2-Keys and Symbol Domain. 
Details on the benchmarks can be found in~\cite{toronto}.  The graph plots
show a linear increase in runtime for LRM as trace length increases.  The
runtime for Synth on the other hand saturates after an initial increase in
runtime, and is much lower compared to LRM.  Given a set of observed traces,
Synth generates a model from the first trace in the set, then incrementally
adds any new behaviour seen in subsequent traces.  For smaller trace sets,
we see new behaviours being added to the model with each trace sequence in
the set, but for larger trace sets subsequent trace sequences do not show
any new behaviours that are not already captured by the automaton. 
Therefore, the runtime saturates once all behaviours are captured by the
generated automaton.  The minimal noise observed in the runtime plots can be
attributed to trace processing and plotting the generated automaton.

\newpage
\section{Implementation Details}

All simulations have been carried out on a machine with an Intel Xeon
3.5\,GHz processor, Nvidia Tesla V100 GPU and 16\,GB of RAM, running
Ubuntu~18.

%\footnote{all the source code can be found here: \url{https://drive.google.com/drive/folders/1qpgw0Jou4Wc7LUOKqhVytohxySsqKrDi?usp=sharing}}.

\subsection{Minecraft}

The Minecraft environment consists of 7 crafting tasks taken
from~\cite{pol-sketch} that requires the agent to execute optimal low-level
actions in order to accomplish high-level objectives in proper order.  The
extrinsic reward of $+1$ is given to the agent only when the whole task is
achieved and the tasks are fully unknown from the beginning.  A number of
tasks include a long sequence of high-level objectives, and thus the
associated reward is extremely sparse and non-Markovian
(Table~\ref{tab:sketch}).  Note that, as shown in the experiments, DeepSynth
can handle a stochastic environment with continuous state spaces; the
example of the deterministic Minecraft environment is chosen for the sake of
exposition and to enable comparison with~\cite{pol-sketch}.

In this example, the agent location in the grid is the MDP state $s \in
\mathcal{S}$.  At each state $s \in \mathcal{S}$ the agent has a set of
actions
$\mathcal{A}=\{\mathit{left},\mathit{right},\mathit{up},\mathit{down}\}$ by
which it is able to move to a neighbouring state $s' \in \mathcal{S}$ unless
stopped by the boundary of $\mathcal{S}$ or by an obstacle.  Obstacle states
are the blue cells which represent a river in the game.  We assume the
elements of the vocabulary set {$\Sigma~=~\{ \mbox{\tt wood}, \mbox{\tt
grass}, \allowbreak\mbox{\tt iron}, \allowbreak\mbox{\tt craft-table},
\allowbreak\mbox{\tt smith-table}, \allowbreak\mbox{\tt gold}\}$} to be
known.

\begin{figure}[t]\centering
\includegraphics[width=0.8\columnwidth]{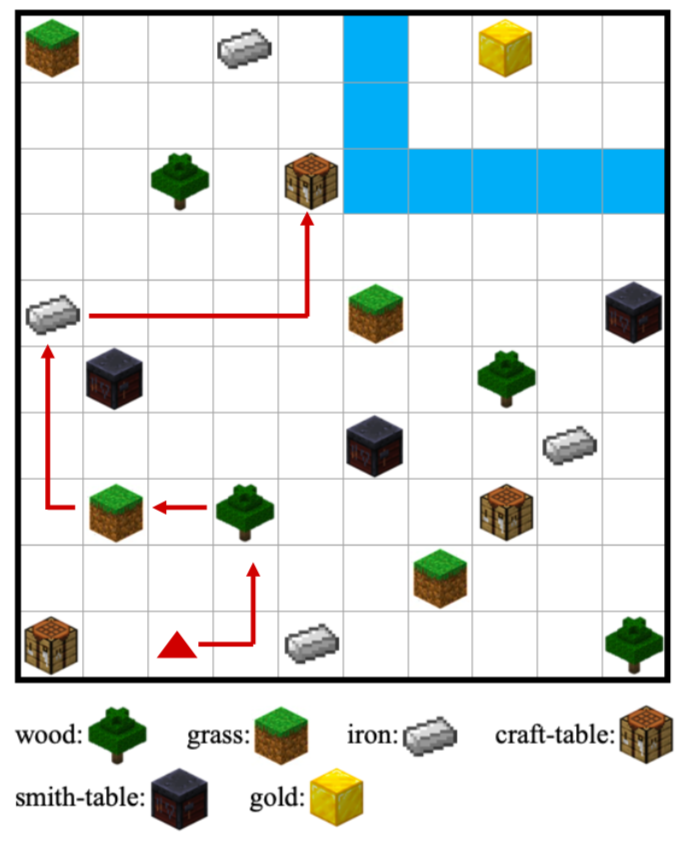}
\caption{Example policy learnt by DeepSynth for Task 3 of the Minecraft environment with given vocabulary
{\ttfamily $\Sigma=\{$wood, grass, iron, craft-table, smith-table, gold$\}$}}
\label{minecraft}
\end{figure}

Recall that the reward in this game is generally sparse and non-Markovian:
the agent will receive a positive extrinsic reward only when a correct
sequence is performed in each (high-level) task.  Namely, the reward
$\widehat{R}:(\mathcal{S}\times\mathcal{A})^*\rightarrow \mathds{R}$ is a
function over the set of finite paths.  A trace-dependent reward is
associated to the accomplishment of a given task: for example, performing a
high-level task
\begin{center}
	{\ttfamily
		Task 1: \textbf{wood}$~\rightarrow~$grass$~\rightarrow~$\textbf{craft-table}} 
\end{center}
results in an extrinsic reward
$\widehat{R_1}$, and for another high-level task such as
\begin{center}
	{\ttfamily
		Task 4: {grass}$~\rightarrow~$\textbf{wood}$~\rightarrow~${iron}$~\rightarrow~$\textbf{smith-table}} 
\end{center}
the extrinsic reward is $\widehat{R_4}$.  Further,  these temporal orderings
are initially unknown, and unlike \cite{pol-sketch} the agent is not
equipped with any instructions to accomplish them.  In these scenarios,
existing exploration schemes fail, and prior work such as~\cite{lcnfq,hahn}
requires the temporal ordering to be known in advance.

For each task in the Minecraft environment, as shown in
Table~\ref{tab:sketch}, we construct a DFA from trace sequences gathered by
intrinsically-motivated exploration.  The generated DFAs are presented in
Fig.~\ref{fig:task1}--\ref{fig:task6} and Fig.~\ref{fig:task7} where the
green state shows task satisfaction.

\begin{table*}[btp]
	\centering
	\ttfamily
	\small{
		\begin{tabular}{|l|l@{\hspace{3mm}}l@{\hspace{3mm}}l@{\hspace{3mm}}l@{\hspace{3mm}}l@{\hspace{3mm}}l@{\hspace{3mm}}l@{\hspace{3mm}}l|}
			\hline
			\hline
			\textrm{\colorbox{white}{\textbf{Task}}}~ & \multicolumn{8}{l|}{\textrm{\textbf{Sequence}}}                                      \\ \hline
			\colorbox{white}{Task1} & $\Sigma^*$ & wood  & $\Sigma^*$ & craft table & & & & \\
			\colorbox{white}{Task2} & $\Sigma^*$ & grass & $\Sigma^*$ & craft table & & & & \\
			\colorbox{white}{Task3} & $\Sigma^*$ & wood  & $\Sigma^*$ & grass & $\Sigma^*$ & iron & $\Sigma^*$ & craft table \\
			\colorbox{white}{Task4} & $\Sigma^*$ &  wood  & $\Sigma^*$ & smith table & & & & \\
			\colorbox{white}{Task5} & $\Sigma^*$ & grass & $\Sigma^*$ & smith table & & & & \\
			\colorbox{white}{Task6} & $\Sigma^*$ & iron  & $\Sigma^*$ & wood & $\Sigma^*$ & smith table & & \\
			\colorbox{white}{Task7} & $\Sigma^*$ & wood  & $\Sigma^*$ & iron & $\Sigma^*$ & craft table & $\Sigma^*$ & gold \\
			\hline
			\hline              
	\end{tabular}}
	\caption{High-level sequence for each task}
	\label{tab:sketch}
\end{table*}

\begin{figure*}[p]
\begin{tabular}{@{}l@{}l@{}}
\begin{minipage}{0.5\textwidth}\centering\small\ttfamily
	\begin{tikzpicture}[->,>=stealth',shorten >=1pt,auto,node distance=2.5cm, thick]
	\tikzstyle{every stateo}=[fill=white,draw=black,text=black]
	\node[initial,state] (A)                    {$q_1$};
	\node[state]         (B) [right of=A] {$q_2$};
	\node[state]         (D) [below of=A] {$q_4$};
	\node[state]         (E) [left of=D] {$q_5$};
	\node[state,accepting,fill=green] (F) [above of=E,yshift=-4mm] {$q_6$};
	\node[state]         (C) [right of=D] {$q_3$};

	\path (A) edge [right]      node[anchor=south]{iron} (B)
	(B) edge [right]            node[anchor=west]{grass} (C)
	(A) edge [right]                node [anchor=south,xshift=-5mm,sloped]{grass} (C)
	(B) edge [left]            node[anchor=north,sloped,xshift=-5mm]{wood} (D)
	(C) edge [left]            node[anchor=north]{wood} (D)
	(D) edge [left]            node[anchor=west,yshift=3mm,xshift=-3mm]{crafttable} (F)
	(D) edge [left]            node[anchor=south]{smithtable} (E)
	(E) edge [left]            node[anchor=east]{crafttable} (F)
	(A) edge [loop above]            node[anchor=south]{$\emptyset$} (A)
	(B) edge [loop above]            node[anchor=south]{$\emptyset$} (B)
	(C) edge [loop below]            node[anchor=north]{$\emptyset$} (C)
	(D) edge [loop below]            node[anchor=north]{$\emptyset$} (D)	
	(E) edge [loop left]            node[anchor=east]{$\emptyset$} (E)
	;
	\end{tikzpicture}
\subcaption{Task 1\label{fig:task1}}
\end{minipage} & \begin{minipage}{0.5\textwidth}\centering\small\ttfamily

	\begin{tikzpicture}[->,>=stealth',shorten >=1pt,auto,node distance=2.5cm, thick]
	\tikzstyle{every state}=[fill=white,draw=black,text=black]
	\node[initial,state] (A)                    {$q_1$};
	\node[state]         (B) [right of=A] {$q_2$};
	\node[state]         (E) [below left of=A] {$q_5$};
	\node[state,accepting,fill=green]         (F) [below of=E,yshift=10mm] {$q_6$};
	\node[state]         (D) [below right of=E,yshift=2mm] {$q_4$};
	\node[state]         (C) [right of=D] {$q_3$};

	\path (A) edge [right]      node[anchor=south,align=center]{smithtable,\\ iron} (B)
	(B) edge [right]            node[anchor=south,sloped]{wood} (C)
	(C) edge [bend right]            node[anchor=west]{smithtable} (B)
	(A) edge [right]                node [anchor=north,sloped,xshift=9mm]{wood} (C)
	(D) edge [left]            node[anchor=south,xshift=9mm,sloped]{\scriptsize smithtable} (B)
	(C) edge [left]            node[anchor=north]{iron} (D)
	(C) edge [left]            node[anchor=south,sloped,xshift=-8.7mm]{grass} (E)
	(D) edge [left]            node[anchor=north,sloped]{grass} (E)
	(B) edge [left]            node[anchor=north,sloped,xshift=-8mm]{grass} (E)
	(E) edge [left]            node[anchor=east]{crafttable} (F)
	(A) edge [left]            node[anchor=south,sloped]{grass} (E)

	(A) edge [loop above]            node[anchor=south]{$\emptyset$} (A)
	(B) edge [loop above]            node[anchor=south]{$\emptyset$} (B)
	(C) edge [loop below]            node[anchor=north]{$\emptyset$} (C)
	(D) edge [loop below]            node[anchor=north]{$\emptyset$} (D)	
	(E) edge [loop left]            node[anchor=east]{$\emptyset$} (E)
	;
	\end{tikzpicture}
\subcaption{Task 2\label{fig:task2}}
\end{minipage}\\
\\\\
\begin{minipage}{0.5\textwidth}\centering\small\ttfamily

	\begin{tikzpicture}[->,>=stealth',shorten >=1pt,auto,node distance=2cm, thick]
	\tikzstyle{every state}=[fill=white,draw=black,text=black]
	\node[initial,state] (A)                    {$q_1$};
	\node[state]         (B) [right of=A] {$q_2$};
	\node[state]         (D) [below of=A,yshift=4mm] {$q_4$};
	\node[state]         (C) [right of=D] {$q_3$};
	\node[state,accepting,fill=green]         (E) [left of=D,xshift=-6mm] {$q_5$};

	\path (A) edge [right]      node[anchor=south]{wood} (B)
	(B) edge [right]            node[anchor=west]{grass} (C)
	(C) edge [right]                node [anchor=south]{iron} (D)
	(D) edge [left]            node[anchor=north]{crafttable} (E)

	(A) edge [loop above]            node[anchor=south]{$\emptyset$} (A)
	(B) edge [loop above]            node[anchor=south]{$\emptyset$} (B)
	(C) edge [loop below]            node[anchor=north]{$\emptyset$} (C)
	(D) edge [loop below]            node[anchor=north]{$\emptyset$} (D)	
	;
	\end{tikzpicture}
\subcaption{Task 3\label{fig:task3}}
\end{minipage}&\begin{minipage}{0.5\textwidth}\centering\small\ttfamily

	\begin{tikzpicture}[->,>=stealth',shorten >=1pt,auto,node distance=2.5cm, thick]
	\tikzstyle{every state}=[fill=white,draw=black,text=black]
	\node[initial,state] (A)                    {$q_1$};
	\node[state]         (B) [right of=A] {$q_2$};
	\node[state,accepting,fill=green]         (C) [right of=B] {$q_3$};

	\path (A) edge       node[anchor=south]{wood} (B)
	(B) edge        node[anchor=south]{smithtable} (C)

	(A) edge [loop below]            node[anchor=north]{$\emptyset$} (A)
	(B) edge [loop below]            node[anchor=north]{$\emptyset$} (B)
	;
	\end{tikzpicture}
\subcaption{Task 4\label{fig:task4}}
\end{minipage}\\
\\\\
\begin{minipage}{0.5\textwidth}\centering\small\ttfamily

	\begin{tikzpicture}[->,>=stealth',shorten >=1pt,auto,node distance=3.5cm, thick]
	\tikzstyle{every state}=[fill=white,draw=black,text=black]
	\node[initial,state] (A)                    {$q_1$};
	\node[state]         (B) [right of=A] {$q_2$};
	\node[state]         (E) [below left of=A,yshift=-5mm] {$q_5$};
	\node[state,accepting,fill=green]         (F) [above of=E,yshift=-14mm] {$q_6$};
	\node[state]         (D) [below right of=E,yshift=9mm] {$q_4$};
	\node[state]         (C) [right of=D] {$q_3$};

	\path (A) edge [bend left]      node[anchor=south]{iron} (B)
	(B) edge [right]      node[anchor=south,sloped,xshift=9mm]{smithtable} (F)
	(B) edge [right]            node[anchor=south,sloped]{wood} (C)
	(C) edge [bend right]            node[anchor=west]{iron} (B)
	(A) edge [right]                node [anchor=south,sloped,xshift=15mm]{wood} (C)
	(D) edge [left]            node[anchor=north,sloped,xshift=-10mm]{iron} (B)
	(B) edge [bend left]            node[anchor=west,xshift=-11mm,yshift=-12mm]{grass} (D)
	(C) edge [left]            node[anchor=north]{grass} (D)
	(D) edge [left]            node[anchor=south,sloped,xshift=7mm]{smithtable} (F)
	(A) edge [left]            node[anchor=south,sloped,xshift=2mm]{grass} (D)
	(E) edge [left]            node[anchor=east]{grass} (D)
	(B) edge [left]            node[anchor=north,sloped,xshift=15mm]{crafttable} (E)
	(E) edge [left]            node[anchor=east]{smithtable} (F)

	(A) edge [loop above]            node[anchor=south]{$\emptyset$} (A)
	(B) edge [loop above]            node[anchor=south]{$\emptyset$} (B)
	(C) edge [loop below]            node[anchor=north]{$\emptyset$} (C)
	(D) edge [loop below]            node[anchor=north]{$\emptyset$} (D)	
	(E) edge [loop left]            node[anchor=east]{$\emptyset$} (E)
	;
	\end{tikzpicture}
\subcaption{Task 5\label{fig:task5}}
\end{minipage}&\begin{minipage}{0.5\textwidth}\centering\small\ttfamily

	\begin{tikzpicture}[->,>=stealth',shorten >=1pt,auto,node distance=2.5cm, thick]
	\tikzstyle{every state}=[fill=white,draw=black,text=black]
	\node[initial,state] (A)                    {$q_1$};
	\node[state]         (B) [right of=A] {$q_2$};
	\node[state,accepting,fill=green]         (E) [below right of=B,yshift=5mm,xshift=5mm] {$q_3$};
	\node[state]         (D) [below of=A] {$q_5$};
	\node[state]         (C) [right of=D] {$q_4$};

	\path (A) edge [right]      node[anchor=south]{grass} (B)
	(B) edge [right]            node[anchor=west]{wood} (C)
	(B) edge [right]            node[anchor=south,sloped]{iron} (D)
	(C) edge [bend left]                node [anchor=north]{iron} (D)
	(C) edge [left]            node[anchor=north,sloped]{smithtable} (E)
	(D) edge [right]                node [anchor=south]{wood} (C)
	(A) edge [left]            node[anchor=south,sloped]{iron} (D)

	(A) edge [loop above]            node[anchor=south]{$\emptyset$} (A)
	(B) edge [loop above]            node[anchor=south]{$\emptyset$} (B)
	(C) edge [loop below]            node[anchor=north]{$\emptyset$} (C)
	(D) edge [loop below]            node[anchor=north]{$\emptyset$} (D)	
	;
	\end{tikzpicture}
\subcaption{Task 6\label{fig:task6}}
\end{minipage}
\end{tabular}
\caption{Automata generated for the Minecraft Tasks 1 to 6}
\end{figure*}

\begin{figure*}
\centering\small\ttfamily

	\begin{tikzpicture}[->,>=stealth',shorten >=1pt,auto,node distance=3cm, thick]
	\tikzstyle{every state}=[fill=white,draw=black,text=black]
	\node[state] (E) 							{$q_5$};
	\node[state]         (D) [right of=E] {$q_4$};
	\node[state]         (C) [above right of=D] {$q_3$};
	\node[state]         (F) [above left of=E] {$q_6$};
	\node[initial,state] (A)  [above left of=F]                  {$q_1$};
	\node[state]         (B) [right of=A,xshift=2.5cm] {$q_2$};
	\node[state,accepting,fill=green]         (G) [below left of=F] {$q_7$};

	\path (A) edge [left]      node[anchor=south]{smithtable} (B)
	(B) edge [left]      node[anchor=south,sloped]{wood} (F)
	(A) edge [left]      node[anchor=north,sloped]{wood} (F)
	(B) edge [bend left]            node[anchor=south,sloped]{crafttable} (C)
	(C) edge [left]            node[anchor=north,sloped]{smithtable} (B)
	(D) edge [left]            node[anchor=north,sloped,xshift=-10mm]{smithtable} (B)
	(D) edge [left]            node[anchor=north,sloped]{crafttable} (C)
	(F) edge [left]            node[anchor=south,sloped]{iron} (D)
	(C) edge [left]            node[anchor=south,xshift=-2.5cm]{wood} (F)
	(D) edge [bend left]            node[anchor=north]{grass} (E)
	(E) edge [left]            node[anchor=south]{iron} (D)
	(F) edge [left]            node[anchor=south,sloped]{grass} (E)
	(E) edge [bend left]            node[anchor=north,sloped]{wood} (F)
	(C) edge [left]            node[anchor=south,sloped,xshift = 10mm]{grass} (E)
	(F) edge [left]            node[anchor=south,sloped]{gold} (G)
	(E) edge [left]            node[anchor=south,sloped,xshift=-5mm]{gold} (G)
	(D) edge [bend left=45]            node[anchor=south,sloped]{gold} (G)

	(A) edge [loop above]            node[anchor=south]{$\emptyset$} (A)
	(B) edge [loop above]            node[anchor=south]{$\emptyset$} (B)
	(C) edge [loop right]            node[anchor=west]{$\emptyset$} (C)
	(D) edge [loop below]            node[anchor=north]{$\emptyset$} (D)	
	(E) edge [loop below]            node[anchor=north]{$\emptyset$} (E)
	(F) edge [loop left]            node[anchor=east]{$\emptyset$} (F)
	;
	\end{tikzpicture}
\caption{Automaton generated for the Minecraft Task 7\label{fig:task7}}
\end{figure*}

As mentioned in the body of paper, for experiments where the state is
already in vector form we employ NFQ modules instead of DQN ones.  Similar
to DQN, NFQ uses experience replay in order to efficiently approximate the
$Q$-function in general MDPs with continuous state spaces.  Let $q_i \in
\mathcal{Q}$ be a state in the DFA $\mathfrak{A}$.  Then define
$\mathcal{E}_{q_i}$ as the projection of $\mathcal{E}$ onto $q_i$.  Each NFQ
module $B_{q_i}$ is trained by its associated experience set
$\mathcal{E}_{q_i}$.  At each iteration a pattern set $\mathcal{P}_{q_i}$ is
generated based on $\mathcal{E}_{q_i}$:
$$
\mathcal{P}_{q_i}=\{(\mathit{input}_l,\mathit{target}_l), l=1,...,|\mathcal{E}_{q_i}|)\},$$
where
$$input_l=({s_l},a_l),$$ and $$\mathit{target}_l \allowbreak =
\allowbreak R^T({s_l},a_l) + \allowbreak \gamma \allowbreak
\max_{a'\in\mathcal{A}} \allowbreak Q({{s_l}}',a'),$$
such that $\langle s_l,a_l,{s_l}',r^T,L(s_l') \rangle$.  This pattern set is
then used to train the neural net $B_{q_i}$ as in Algorithm~\ref{lcnfqal}.

We~use the Adam optimiser~\cite{adam} to update the weights in each module
(line~8).  Within each fitting epoch (lines 2--10), the training schedule
starts from networks that are associated with accepting states of the
automaton and goes backward until it reaches the networks that are
associated to the initial states.  In this way we back-propagate the
$Q$-value through the networks one by one.  Later, once the $Q$-value has
converged to the approximated optimal $\widehat{Q}^*$, the policy is
synthesised by maximising the $\widehat{Q}^*$.

The training progress for Task 1 and Task 3 in Minecraft is illustrated in
Fig.~\ref{fig:result_1} and Fig.~\ref{fig:result_3_a}.  In
Fig.~\ref{fig:result_3_a} the orange line shows the very first deep net
associated to the initial state of the DFA, the red and blue ones are of the
intermediate states in the DFA and the green line is associated to the final
state.  This shows an efficient back-propagation of extrinsic reward from
the final high-level state to the initial state, namely once the last deep
net converges the expected reward is back-propagated to the second and so
on.  Other networks associated with automaton states within non-optimal
paths remained unstable and are not shown.  In both tasks, each NFQ module
includes $2$ hidden layers and $128$ ReLUs.

\begin{algorithm2e}[!b]
	\DontPrintSemicolon
	\SetKw{return}{return}
	\SetKwRepeat{Do}{do}{while}
	%\SetKwFunction{assume}{assume}
	%\SetKwFunction{isf}{isFeasible}
	\SetKwData{conflict}{conflict}
	\SetKwData{safe}{safe}
	\SetKwData{sat}{sat}
	\SetKwData{unsafe}{unsafe}
	\SetKwData{unknown}{unknown}
	\SetKwData{true}{true}
	\SetKwInOut{Input}{input}
	\SetKwInOut{Output}{output}
	\SetKwFor{Loop}{Loop}{}{}
	\SetKw{KwNot}{not}
	\begin{footnotesize}
		\Input{automaton $\mathfrak{A}$ from the Synth step,\\ the set of transition samples $\mathcal{E}$}
		\Output{approximated optimal $Q$-function: $\hat{Q}^*$}
		initialize all neural nets $B_{q_i}$ with random weights\;
		\Repeat{end of trial}
		{
			\For{$q_i=|\mathcal{Q}|$ {\normalfont\textbf{to}} $1$}
			{
				$\mathcal{P}_{q_i}=\{(\mathit{input}_l,\mathit{target}_l),~l=1,...,|\mathcal{E}_{q_i}|)\}$\;
				~~~~~~~~~$\mathit{input}_l=(s_l,a_l)$\;
				~~~~~~~~~~\hspace{1mm}$\mathit{target}_l=R^T(s_l,a_l)+\gamma \max \limits_{a'} Q(s_l',a')$\;
				~~~~~~~~~where $(s_l,a_l,{{s_l}}',r^t,L(s'_L)) \in \mathcal{E}_{q_i}$\;
				$B_{q_i} \leftarrow$ Adam$(\mathcal{P}_{q_i})$
			}
		}
	\end{footnotesize}
	\caption{DeepSynth with Temporal NFQ}
	\label{lcnfqal}
\end{algorithm2e}

\begin{figure*}
\begin{minipage}{0.45\columnwidth}
	\centering
	\begin{tikzpicture}[scale=0.6]
	%	\definecolor{color0}{rgb}{0.75,0,0.75}
	\definecolor{oxford_blue}{rgb}{0,0.13,0.28}
	
	\begin{axis}[
	axis background/.style={fill=white},
	axis line style={black},
	tick align=outside,
	tick pos=left,
	x grid style={white!89.80392156862746!black},
	xlabel={Epoch Number},
	xmajorgrids,
	xmin=-4.95, xmax=103.95,
	xtick style={color=white!33.33333333333333!black},
	y grid style={white!89.80392156862746!black},
	ylabel={Loss},
	ymajorgrids,
	ymin=-0.04, ymax=1.05,
	ytick style={color=white!33.33333333333333!black},
	legend style={draw=black, fill=white}
	]
	\addplot [line width=1.64pt, red]
	table {%
		0 0.00226820191857362
		1 0.00105805164828321
		2 0.000558881349024778
		3 0
		4 0.000165771286353101
		5 0.000817121781852747
		6 0.00429518302561603
		7 0.0667162349070649
		8 0.341859389941351
		9 0.589113734738963
		10 0.759710356350914
		11 0.77723332351119
		12 0.799766319221119
		13 0.854060267196374
		14 0.859889008294507
		15 0.917566096389879
		16 0.913563978100334
		17 1
		18 0.752962752927602
		19 0.462219381131879
		20 0.267374975480813
		21 0.150107970987929
		22 0.0914537861406565
		23 0.0931537134259089
		24 0.082399380966962
		25 0.0797203914658563
		26 0.0890641435731393
		27 0.0618165823888686
		28 0.0520383032764583
		29 0.0422154687687773
		30 0.0412490063363559
		31 0.0378154548625864
		32 0.0324830666018465
		33 0.029223446628658
		34 0.0300000840590696
		35 0.0338729238143192
		36 0.0312334889531663
		37 0.0311422700062818
		38 0.0307297086297242
		39 0.0319415332423018
		40 0.0288479968475198
		41 0.0254069858420298
		42 0.0266001971764888
		43 0.0260691772066925
		44 0.0276815083733191
		45 0.0306366145127603
		46 0.0345870268669099
		47 0.032895536257686
		48 0.0342069795661373
		49 0.0318859723582158
		50 0.0315437468491293
		51 0.0329235066365446
		52 0.0311374957211055
		53 0.0290742336307973
		54 0.0279267837135545
		55 0.0270021574026716
		56 0.0232084589555048
		57 0.0222968295667244
		58 0.0225161471564269
		59 0.0232325975742775
		60 0.0232654654798825
		61 0.0239416385923915
		62 0.0275513406987658
		63 0.0233405170905749
		64 0.0195954779578157
		65 0.0219591571769032
		66 0.023748819214901
		67 0.0249180548385736
		68 0.0230461318791973
		69 0.0203053870105953
		70 0.0221974666908575
		71 0.0238897011994914
		72 0.0245128999277034
		73 0.021788764406605
		74 0.0202453536375935
		75 0.0167881861769009
		76 0.0204651529327461
		77 0.0181216120974995
		78 0.0227648113192656
		79 0.0253894831745682
		80 0.0275587861216278
		81 0.02881376654077
		82 0.0255519711643293
		83 0.0286512582624747
		84 0.0309583994220874
		85 0.0293513526003322
		86 0.0272948739535728
		87 0.0222878738405459
		88 0.0203323137933456
		89 0.0198721462236622
		90 0.0204280911763909
		91 0.0236435168482993
		92 0.0219423194543461
		93 0.0216611181267253
		94 0.0220093995491976
		95 0.0210033409760862
		96 0.0253310657261782
		97 0.0207202443645061
		98 0.0173870899599741
		99 0.0166497157249911
	};
	\addplot [line width=1.64pt, cyan]
	table {%
		0 0.00210265475304704
		1 0
		2 0.00039083246500324
		3 0.000745427843435552
		4 0.000709280781970173
		5 1
		6 0.886969450464399
		7 0.771253224559112
		8 0.513270753557443
		9 0.699641384581995
		10 0.588652913100766
		11 0.574749626139031
		12 0.435950244691162
		13 0.29101602531285
		14 0.214895379162386
		15 0.138465481473789
		16 0.0874708895816055
		17 0.0519005870372013
		18 0.0443095427114212
		19 0.0375389430897292
		20 0.0590992674693107
		21 0.0452980936415078
		22 0.0434231993867595
		23 0.035134840015014
		24 0.0334151990884742
		25 0.0350862809886211
		26 0.0341010921146968
		27 0.0332799914929146
		28 0.0295748769324761
		29 0.0284787734922108
		30 0.0271241491449599
		31 0.0277283595811327
		32 0.0304160138553115
		33 0.032651270130731
		34 0.0334976300243565
		35 0.0315041609801717
		36 0.0311404207015566
		37 0.029246713951078
		38 0.0257304932841175
		39 0.0266060792101161
		40 0.0298159141470878
		41 0.035556610368598
		42 0.0472848930818862
		43 0.0401483810855293
		44 0.033888832996931
		45 0.0343076595287094
		46 0.0394511206907146
		47 0.03902229892102
		48 0.0408632282725592
		49 0.0343027050825521
		50 0.0326043328288782
		51 0.0295963384774104
		52 0.0297011325506729
		53 0.0278758827027448
		54 0.0237939116633363
		55 0.0237839396576704
		56 0.0225842618876472
		57 0.0263345386974246
		58 0.0250898669553388
		59 0.028376086435144
		60 0.0273406705277329
		61 0.026839843381881
		62 0.0286228943034827
		63 0.0302781914652173
		64 0.027810429886514
		65 0.0299140409448514
		66 0.0336719991882117
		67 0.0343179539837265
		68 0.0346661381092681
		69 0.0346623324780783
		70 0.0341943322869113
		71 0.0325385428024473
		72 0.0296757132014913
		73 0.0275385000139287
		74 0.0269031344125832
		75 0.0266372034430106
		76 0.0314129770323087
		77 0.0318329501647136
		78 0.0317973572459226
		79 0.0326258982392173
		80 0.0330893419033527
		81 0.0307273574356487
		82 0.029614753071922
		83 0.0279875317932224
		84 0.0243405294297506
		85 0.0254117862110464
		86 0.0262994837538008
		87 0.0265927346572126
		88 0.0268682109202044
		89 0.0300405274887182
		90 0.0336933832638959
		91 0.0355830237454755
		92 0.0332720952535881
		93 0.0307953267349333
		94 0.0294410825480378
		95 0.0271731182109952
		96 0.0268852583755905
		97 0.0240877139490752
		98 0.0227658377041164
		99 0.0247925218288762
	};
	\addplot [line width=1.64pt, teal]
	table {%
		0 0.768542717028077
		1 0.884626425364041
		2 1
		3 0.433499829457181
		4 0.0184754052475033
		5 0.00716347873479202
		6 0.00436366066199172
		7 0.00288226876221181
		8 0.00381361241510824
		9 0.00146715096474405
		10 0.000945767780229779
		11 0.00132014592020236
		12 0.00056002942886993
		13 0.000376515031127124
		14 0.000364619888901333
		15 0.000437405297379484
		16 0.000336526849199241
		17 0.000240503117979751
		18 0.000429303730104743
		19 0.000762636742056556
		20 0.00078613577554094
		21 0.000255378624652557
		22 0.000304755015321148
		23 0.000371463568800679
		24 0.000287855783653524
		25 0.00055914866601107
		26 0.000321652063112734
		27 0.000195158365240837
		28 0.000267698508943969
		29 0.000203518016911023
		30 0.000233059450135243
		31 0.000187255623007543
		32 0.000265173813421246
		33 0.000266385122857072
		34 0.000283776279745666
		35 0.000258460216418974
		36 0.000454927415734982
		37 0.000649399754518231
		38 0.000236431863960898
		39 0.00027434035739629
		40 0.000763685948056904
		41 0.000897360662997071
		42 0.000806379032841575
		43 0.000564044752039369
		44 0.00121427100712671
		45 0.00073303798585431
		46 0.000645137249855237
		47 0.00121154207854676
		48 0.000984481511202037
		49 0.00056977791666008
		50 0.000500591615421037
		51 0.000464708851552634
		52 0.000267670700220398
		53 0.000217546663914987
		54 0.000203851566667079
		55 0.000239325955522512
		56 0.000293191983949712
		57 0.000210925392759634
		58 0.000488032960493743
		59 0.000196523633370345
		60 0.000300575308215343
		61 0.000128879331667904
		62 6.48660981407319e-05
		63 0.000135528753563674
		64 0.00011916605382844
		65 0.000166615304710764
		66 0.000614173696453143
		67 0.00023143898763658
		68 0.000159764238191835
		69 0.000135660562445872
		70 0.00012552529314335
		71 0.000106804704796755
		72 8.40934520321745e-05
		73 0.000526463716931263
		74 0.000198777205047085
		75 0.000253630624083082
		76 0.000314281609109747
		77 6.40077216281003e-05
		78 5.68918729727626e-05
		79 0.000123972655342604
		80 0.000104083972831247
		81 3.58852294603933e-05
		82 4.1395788001757e-05
		83 0.000180504606856285
		84 0.000145328342500006
		85 0.000155379037462677
		86 0.000238202850118587
		87 9.85323691516852e-05
		88 0.000149690092180958
		89 0.000228800879829998
		90 0.0001014815802716
		91 9.77799474003005e-05
		92 9.03764198376461e-05
		93 0.000145683396104438
		94 0
		95 0.000690460546363017
		96 0.000211169611282198
		97 0.000232641715847219
		98 0.000120568677399409
		99 0.000171010771272906
	};
	\legend{$q_1$, $q_2$, $q_4$}
	\end{axis}
	
	\end{tikzpicture}
\subcaption{Task1}
\label{fig:result_1}
\end{minipage}
\quad
\begin{minipage}{0.45\columnwidth}
%\begin{figure}[!t]
	\centering
	\begin{tikzpicture}[scale=0.6]
	%	\definecolor{color0}{rgb}{0.75,0,0.75}
	\definecolor{oxford_blue}{rgb}{0,0.13,0.28}
	\begin{axis}[
	axis background/.style={fill=white},
	axis line style={black},
	tick align=outside,
	tick pos=left,
	x grid style={white!89.80392156862746!black},
	xlabel={Steps},
	xmajorgrids,
	xmin=-4.95, xmax=103.95,
	xtick style={color=white!33.33333333333333!black},
	y grid style={white!89.80392156862746!black},
	ylabel={Loss},
	ymajorgrids,
	ymin=-0.04, ymax=1.05,
	ytick style={color=white!33.33333333333333!black},
	legend style={draw=black, fill=white}
	]
	\addplot [line width=1.64pt, orange]
	table {%
		0 0.0219974718391983
		1 0.00189497959765653
		2 0
		3 0.000906162101425027
		4 0.000872798931744083
		5 0.0019594045874813
		6 0.00311768502190989
		7 0.0188322896285152
		8 0.113640830067244
		9 0.282939709388352
		10 0.514819577410236
		11 0.902620645685925
		12 0.835450913744719
		13 0.849803306636028
		14 1
		15 0.620136526650263
		16 0.622401557756216
		17 0.421188243172349
		18 0.254522780763907
		19 0.183077500323138
		20 0.158314537471929
		21 0.13570968464083
		22 0.0900510701765381
		23 0.085301504654935
		24 0.0785587585289398
		25 0.076890180418157
		26 0.0667220483414757
		27 0.0640463432890918
		28 0.0610492970911719
		29 0.0531263653718212
		30 0.0575015767838292
		31 0.0658844327113103
		32 0.0647737209341279
		33 0.0625981330586676
		34 0.0603434366554289
		35 0.0623908516852282
		36 0.0483382486209102
		37 0.0531218923151253
		38 0.0494563516378463
		39 0.0506000958767886
		40 0.0453129112725503
		41 0.0499927982590995
		42 0.050977240388877
		43 0.0500230635977737
		44 0.0431982182136362
		45 0.0416703964844275
		46 0.0431192940499365
		47 0.0424342106581494
		48 0.0387003345091131
		49 0.0377931715028839
		50 0.0358631220976944
		51 0.0383032224140282
		52 0.0440072245381724
		53 0.0498154770339403
		54 0.0447095508602017
		55 0.0442563974061295
		56 0.0442055585940536
		57 0.0500565138153318
		58 0.050265637350969
		59 0.0498373975037875
		60 0.0494242453641012
		61 0.0452957448991756
		62 0.0450192555414657
		63 0.0428252525639273
		64 0.0389235560144666
		65 0.0387726646706917
		66 0.0444168791834289
		67 0.0431930307616336
		68 0.0501070771110487
		69 0.0467723423767068
		70 0.0442317177188028
		71 0.0448645843278006
		72 0.039080538460197
		73 0.0379157184875176
		74 0.0354189726036569
		75 0.0337190280188751
		76 0.0341470760773157
		77 0.0325897120668148
		78 0.0334643249921898
		79 0.032437826188372
		80 0.0340336762825764
		81 0.0368644868403279
		82 0.0351405806510906
		83 0.036124459297301
		84 0.0311453514655808
		85 0.0345026220561112
		86 0.0390174682162321
		87 0.0437615862893644
		88 0.0416413520697665
		89 0.0392125564706872
		90 0.0361700354931935
		91 0.0360068769353516
		92 0.0333918841504097
		93 0.0333089540562063
		94 0.0325783453240001
		95 0.035580800171766
		96 0.0313448835891514
		97 0.0321000403750567
		98 0.0330433178687179
		99 0.0288823756984654
	};
	\addplot [line width=1.64pt, red]
	table {%
		0 0.00864234024647059
		1 0
		2 0.000937900527175273
		3 0.00922995030990268
		4 0.035611850579697
		5 0.135163420985722
		6 0.419153623949563
		7 0.797534878193328
		8 0.99
		9 0.84053344959494
		10 0.826132022582378
		11 0.848511565640329
		12 0.790496145499474
		13 0.679491393605454
		14 0.612921872534113
		15 0.530914669353344
		16 0.465949783666173
		17 0.355167643437352
		18 0.248900471794269
		19 0.21685432690641
		20 0.149896711067824
		21 0.149158953727419
		22 0.128409165561266
		23 0.101953134496922
		24 0.0921271821795723
		25 0.0921534792251195
		26 0.0920660435225799
		27 0.112195245172139
		28 0.110496938063392
		29 0.11138539156858
		30 0.108497230657226
		31 0.0981153332465544
		32 0.0938279565379797
		33 0.108616140694634
		34 0.110051440548829
		35 0.110718485633159
		36 0.0898074189877533
		37 0.0884879710730735
		38 0.0787864620719034
		39 0.084078445379975
		40 0.0816682292388798
		41 0.0789294494301557
		42 0.0647601127137202
		43 0.0662696470795122
		44 0.0706504298529633
		45 0.0776342868176092
		46 0.0784252642778984
		47 0.0871079631583549
		48 0.0788834791209375
		49 0.0850482224868128
		50 0.0783825999225139
		51 0.0856720505435502
		52 0.0844467977589292
		53 0.0991792801183931
		54 0.106245202512411
		55 0.0991873544113814
		56 0.099099374540341
		57 0.102803356209641
		58 0.0920220078052647
		59 0.0874467054677075
		60 0.0834110781247223
		61 0.0817337649547796
		62 0.0945820923843558
		63 0.0943829889869492
		64 0.10399634536155
		65 0.0884138192904141
		66 0.078195921097295
		67 0.0736568764454254
		68 0.0808085678459566
		69 0.0751716044666618
		70 0.0766513465456401
		71 0.0765912558166009
		72 0.0731632840945614
		73 0.068455784149889
		74 0.067996245325212
		75 0.0708335143282486
		76 0.0740539184566304
		77 0.0803249977258467
		78 0.0861652102663627
		79 0.0919329168256159
		80 0.0986635241916202
		81 0.0944403171225853
		82 0.0979482767739276
		83 0.101701797152881
		84 0.0957601427903384
		85 0.0819880012734439
		86 0.0837920480593645
		87 0.0783633364135052
		88 0.0844907658862384
		89 0.0827970281234514
		90 0.0792853567596844
		91 0.0843023790409799
		92 0.0763227290105166
		93 0.0661412905736569
		94 0.0666261457088366
		95 0.0626032048565481
		96 0.0762272600580626
		97 0.0738764312223583
		98 0.0743838808890086
		99 0.0676258535514504
	};
	\addplot [line width=1.64pt, cyan]
	table {%
		0 0.00989081825592384
		1 0.608118612920054
		2 0.522941483267005
		3 0.45463353494612
		4 0.245923120731229
		5 0.9
		6 0.810314097877729
		7 0.565930363252504
		8 0.659335651707327
		9 0.797586895706883
		10 0.62024078121105
		11 0.47700583245478
		12 0.324409528312561
		13 0.213854183967685
		14 0.129046449372162
		15 0.0935290597160753
		16 0.0719829252422544
		17 0.0562216836156985
		18 0.0471376353986296
		19 0.0387763446321561
		20 0.0353787021512887
		21 0.034726254385767
		22 0.0266689832718646
		23 0.0251885955181299
		24 0.0206333531701117
		25 0.0136106951345889
		26 0.0123706503973877
		27 0.0139494861419395
		28 0.0157359859697341
		29 0.0202313821191468
		30 0.0212002705538863
		31 0.0183062540687078
		32 0.0134824477607028
		33 0.0125979356535329
		34 0.0131668445074561
		35 0.0092809721216348
		36 0.0129408207790234
		37 0.0106927354088025
		38 0.0128588424276185
		39 0.00674995595959702
		40 0.00813737618403789
		41 0.00536911451374637
		42 0.00492225530877994
		43 0.00543576464084643
		44 0.00914207994389291
		45 0.00223786547592836
		46 0.00879176901731022
		47 0.00629358249831715
		48 0.0193635147356342
		49 0.0136090813966036
		50 0.0095607846269669
		51 0.0135331140937718
		52 0.0132083405127988
		53 0.0142254317332874
		54 0.0115030255984427
		55 0.00854474947155487
		56 0.0145788964673339
		57 0.016380868766126
		58 0.0102973824187202
		59 0.0107746616594539
		60 0.014696133265908
		61 0.00787354312575404
		62 0.00262598901760689
		63 0.00404869712859842
		64 0.00934680487105591
		65 0.016721393227874
		66 0.0127269210966935
		67 0.00504876368822966
		68 0.00710362747504549
		69 0.00383818181063979
		70 0.00214456986933007
		71 0.00451895271285959
		72 0.00654656787065781
		73 0.00897447745494359
		74 0.00857421394047558
		75 0.0110199916779951
		76 0.0114552270734137
		77 0.00358699228935473
		78 0.00329501733227772
		79 0
		80 0.0139690145226309
		81 0.00863288814515055
		82 0.00992152834276661
		83 0.00701505053850568
		84 0.00783861322814526
		85 0.00978021067216244
		86 0.00739782690407003
		87 0.00970138440959947
		88 0.00528180082222758
		89 0.0104435037959008
		90 0.00385055852457489
		91 0.00583207251200344
		92 0.00617423896421866
		93 0.00790271773329749
		94 0.0146347393747345
		95 0.00773475553682871
		96 0.0164343387497172
		97 0.0150721640852567
		98 0.00916299668245827
		99 0.00480466064613712
	};
	\addplot [line width=1.64pt, teal]
	table {%
		0 0.846507738562942
		1 1
		2 0.381818709247165
		3 0.038548378129984
		4 0.0172899675734227
		5 0.0103963042932547
		6 0.00731274763888972
		7 0.00817566374033483
		8 0.00791815477350432
		9 0.00562333335600643
		10 0.00731512431319416
		11 0.00421180781308664
		12 0.00123195112056568
		13 0.000706751712891235
		14 0.000471383157830484
		15 0.000377287724407175
		16 0.000355932011933475
		17 0.000260288750414126
		18 0.000216204102242873
		19 0.00018761742996316
		20 0.000174779328373982
		21 0.000144983589979023
		22 0.000139048034425897
		23 0.000104955187376041
		24 9.1493267931788e-05
		25 8.64575285430004e-05
		26 9.32657015715194e-05
		27 7.33202772318895e-05
		28 6.78173526817374e-05
		29 5.39698998126336e-05
		30 3.82789409261336e-05
		31 4.70811726080277e-05
		32 3.2132875937684e-05
		33 2.87978734161124e-05
		34 2.7593481955694e-05
		35 3.40402072431342e-05
		36 4.25673553216888e-05
		37 1.26848260573879e-05
		38 8.24680312983383e-05
		39 9.91152154979975e-06
		40 7.81937718437213e-06
		41 5.0390191053606e-06
		42 0
		43 3.82760635910285e-06
		44 3.04929387083954e-05
		45 4.82001250033122e-05
		46 7.4038575138734e-05
		47 0.000191431021237572
		48 0.000235179750314184
		49 0.000151309344670031
		50 0.000124915176927642
		51 0.000151629824784591
		52 0.000231483813482166
		53 0.000154083922270951
		54 0.000166357214167888
		55 0.000199715285291695
		56 0.000146024954865104
		57 0.000263141281141484
		58 0.000289419916709104
		59 0.000576681229544678
		60 0.000370036492719829
		61 0.000272752453863325
		62 0.000634780909139024
		63 0.000689768957233849
		64 0.000685422449831545
		65 0.000274914337106199
		66 0.000118750024494032
		67 0.000208876720343558
		68 0.000282605947029629
		69 0.000297521587334832
		70 0.000273918001377953
		71 0.000295224086052632
		72 0.000365462700814418
		73 0.00132347766064677
		74 0.000303848436898333
		75 0.000118737475179447
		76 0.000145456721441961
		77 8.47324318426414e-05
		78 0.000311626251153642
		79 0.000382609482887321
		80 0.000296542790919266
		81 0.00069376176419825
		82 0.000485507255325707
		83 0.000420455206949497
		84 0.00047123621341189
		85 0.000537101044682343
		86 0.000156748806072103
		87 0.000478775344878417
		88 0.000431475095433162
		89 0.000189947028844631
		90 0.00028317853848082
		91 0.000243859434811655
		92 0.000164918203675997
		93 0.000312189307349088
		94 0.000498452973043782
		95 0.000514775483066015
		96 0.000426015343641151
		97 0.00014536498003537
		98 0.000394372971752186
		99 0.000220766211859343
	};
	\legend{$q_1$, $q_2$,$q_3$, $q_4$}
	\end{axis}
	\end{tikzpicture}
\subcaption{Task3}
\label{fig:result_3_a}
\end{minipage}
\caption{Training progress for Tasks 1 and 3 with three and four active hybrid deep NFQ modules coupled together, respectively.}
\end{figure*}

%\begin{figure}[!t]\centering
%	\includegraphics[width=0.6\textwidth]{minecraft_path.png}
%	\caption{Example policy learnt by DeepSynth for Task 3}
%	\label{path}
%\end{figure}

The crafting environment outputs a reward for Task~1 when the agent brings
``wood'' to the ``craft table''.  Fig.~\ref{fig:u_task_1} illustrates the
results of training for Task~1.  Note that with the very same training set
$\mathcal{E}$ of 4500 training samples DeepSynth is able to converge while
DQN fails.
%
%However, we allowed DQN to explore more and gather enough training samples to converge.  The larger training set required for this is denoted by $\mathcal{E}'$ and contains 5500 training samples.  The algorithm that employs this larger set is called DQN* in Fig.~\ref{fig:u_task_1}.
%
%The training
%progress for the same is illustrated in Fig.~\ref{fig:result_1}.  The red line in Fig.~\ref{fig:result_1} shows the
%very first deep net associated to~$q_1$, the blue one is of the intermediate
%state $q_2$ in the DFA and the green line is associated to~$q_4$.  This
%figure shows sequential back-propagation from $q_4$ to $q_1$, namely once
%the last deep net converges the expected reward is back-propagated to the
%second and so on.  Hence, the deep net associated to~$q_1$ converges at
%last.  
%
Task~3 has a more complicated sequential structure, as given in
Table~\ref{tab:sketch}.  An example policy learnt by DeepSynth
for Task~3 is provided in Fig.~\ref{tab:sketch}.  Fig.~\ref{fig:u_task_3} gives the
result of training for Task~3 using DeepSynth and DQN where the training set
$\mathcal{E}$ has $6000$ training samples.  However, for Task~3, DQN failed to
converge even after we increased the training set by an order of magnitude
to $60000$.

\begin{figure*}[!t]
	\begin{minipage}{0.45\columnwidth}
	\centering
	\begin{tikzpicture}[scale=0.6]
	\definecolor{oxford_blue}{rgb}{0,0.13,0.28}
	\begin{axis}[
	axis background/.style={fill=white},
	axis line style={black},
	tick align=outside,
	tick pos=left,
	x grid style={white!89.80392156862746!black},
	xlabel={Epoch Number},
	xmajorgrids,
	xmin=-4.95, xmax=103.95,
	xtick style={color=white!33.33333333333333!black},
	y grid style={white!89.80392156862746!black},
	ylabel={Expected Discounted Reward},
	ymajorgrids,
	ymin=-0.0276098847900357, ymax=0.85,
	ytick style={color=white!33.33333333333333!black},
	legend style={draw=black, fill=white},
	]
	\addplot [line width=1.64pt, orange]
	table {%
		0 0
		1 0.00431733950972557
		2 0.0106481453403831
		3 0.0138689950108528
		4 0.0195301696658134
		5 0.0229513458907604
		6 0.0255090575665236
		7 0.0296972654759884
		8 0.033820204436779
		9 0.0408200100064278
		10 0.0448088832199574
		11 0.067077711224556
		12 0.108686119318008
		13 0.183357551693916
		14 0.268040716648102
		15 0.402545928955078
		16 0.479633748531342
		17 0.487789750099182
		18 0.47241398692131
		19 0.4895339012146
		20 0.49701139330864
		21 0.504104673862457
		22 0.512821316719055
		23 0.513105869293213
		24 0.518888533115387
		25 0.525080621242523
		26 0.527886927127838
		27 0.528277695178986
		28 0.527631878852844
		29 0.523696660995483
		30 0.519606292247772
		31 0.521571636199951
		32 0.521705865859985
		33 0.522488653659821
		34 0.518547534942627
		35 0.523574888706207
		36 0.514267921447754
		37 0.518903911113739
		38 0.526214420795441
		39 0.526579558849335
		40 0.535552442073822
		41 0.539129555225372
		42 0.543335378170013
		43 0.538076937198639
		44 0.533085942268372
		45 0.530596017837524
		46 0.533137738704681
		47 0.538177371025085
		48 0.536502540111542
		49 0.531742215156555
		50 0.538953959941864
		51 0.535583198070526
		52 0.534995079040527
		53 0.525425970554352
		54 0.52646791934967
		55 0.527913331985474
		56 0.531678974628448
		57 0.533949255943298
		58 0.541063845157623
		59 0.535906434059143
		60 0.538173198699951
		61 0.539213716983795
		62 0.54160088300705
		63 0.541565716266632
		64 0.540591239929199
		65 0.548640847206116
		66 0.549121916294098
		67 0.544667184352875
		68 0.544145703315735
		69 0.539474725723267
		70 0.53059321641922
		71 0.523718118667603
		72 0.520298421382904
		73 0.523507237434387
		74 0.529160022735596
		75 0.532530963420868
		76 0.532611906528473
		77 0.53290867805481
		78 0.531335711479187
		79 0.532177269458771
		80 0.531749427318573
		81 0.536367535591125
		82 0.538284659385681
		83 0.533191561698914
		84 0.53061580657959
		85 0.52616947889328
		86 0.521397173404694
		87 0.522711515426636
		88 0.525781333446503
		89 0.526919424533844
		90 0.529275476932526
		91 0.531123399734497
		92 0.532268345355988
		93 0.532204747200012
		94 0.539092838764191
		95 0.544586598873138
		96 0.546326458454132
		97 0.547029495239258
		98 0.550163745880127
		99 0.552973866462708
	};
	\addplot [line width=1.64pt, oxford_blue]
	table {%
		0 0
		1 0.0169486747387447
		2 0.0152754923795323
		3 0.00510138051478843
		4 0.00990870174183882
		5 0.00898982129577476
		6 0.0130318594544553
		7 0.0157744670227103
		8 0.0018771917354847
		9 0.000566949530440126
		10 0.0167153020783974
		11 0.00865534135810107
		12 0.0152456016491588
		13 4.21210670222139e-05
		14 0.00890774388109603
		15 0.0144308006468157
		16 0.00457524442540905
		17 0.0189054139110784
		18 0.0180285491522297
		19 0.000611799660671071
		20 0.000508917219869216
		21 0.0108282494558699
		22 0.0187829832555702
		23 0.00762408475376425
		24 0.00433198794261227
		25 0.00844233151165435
		26 0.000580815751497359
		27 0.0044338333254607
		28 0.00875775187301144
		29 0.00991624482763701
		30 0.00466168900515145
		31 0.00461733083081969
		32 0.00437562074675377
		33 0.00919206931475467
		34 0.00579563229180971
		35 0.000429794105318178
		36 0.0167515595132515
		37 0.0111290864530487
		38 0.0128458872586489
		39 0.00371812531789435
		40 0.0198508682435213
		41 0.0171989305759058
		42 0.00241779919611613
		43 0.00665390370720258
		44 0.0144296881516654
		45 0.0142238353939056
		46 0.0187288117359892
		47 0.0084421399992283
		48 0.0166007138654865
		49 0.0134061113282814
		50 0.00606737021865835
		51 0.0117516121228712
		52 0.0176495800166372
		53 0.0169239483685663
		54 0.010105676411592
		55 0.011780045159651
		56 0.000690516603026832
		57 0.00485479947086135
		58 0.0159480849510861
		59 0.00828627998601549
		60 0.00346014803158102
		61 0.0109759752277631
		62 0.0140608152413126
		63 0.0134897166100465
		64 0.00749406041003281
		65 0.00877923260089126
		66 0.0101685297649996
		67 0.0155688523000029
		68 0.0104187683522629
		69 0.00786510189928452
		70 0.00979387040924517
		71 0.000591499279338141
		72 0.000869745807130549
		73 0.0140676417720767
		74 0.0196637543461935
		75 0.0118636746076012
		76 0.00787199372755828
		77 0.00340698393711363
		78 0.0100447711686697
		79 0.0196415327507707
		80 0.015410462796616
		81 0.0107923489689956
		82 0.017205795578411
		83 0.00464352256126029
		84 0.0102754332637527
		85 0.0190493477653654
		86 0.0115558961560241
		87 0.00918263463821337
		88 0.00538558954882842
		89 0.010959926189325
		90 0.0191423256292045
		91 0.000114182589007858
		92 0.0156731046523078
		93 0.0164097182385096
		94 0.0177235916165202
		95 0.0148100682366639
		96 0.0161827980174496
		97 0.01037356567046
		98 0.0112271572955676
		99 0.008521813593763
	};
	\addplot [line width=1.64pt, blue]
	table {%
		0 0.56
		1 0.56
		2 0.56
		3 0.56
		4 0.56
		5 0.56
		6 0.56
		7 0.56
		8 0.56
		9 0.56
		10 0.56
		11 0.56
		12 0.56
		13 0.56
		14 0.56
		15 0.56
		16 0.56
		17 0.56
		18 0.56
		19 0.56
		20 0.56
		21 0.56
		22 0.56
		23 0.56
		24 0.56
		25 0.56
		26 0.56
		27 0.56
		28 0.56
		29 0.56
		30 0.56
		31 0.56
		32 0.56
		33 0.56
		34 0.56
		35 0.56
		36 0.56
		37 0.56
		38 0.56
		39 0.56
		40 0.56
		41 0.56
		42 0.56
		43 0.56
		44 0.56
		45 0.56
		46 0.56
		47 0.56
		48 0.56
		49 0.56
		50 0.56
		51 0.56
		52 0.56
		53 0.56
		54 0.56
		55 0.56
		56 0.56
		57 0.56
		58 0.56
		59 0.56
		60 0.56
		61 0.56
		62 0.56
		63 0.56
		64 0.56
		65 0.56
		66 0.56
		67 0.56
		68 0.56
		69 0.56
		70 0.56
		71 0.56
		72 0.56
		73 0.56
		74 0.56
		75 0.56
		76 0.56
		77 0.56
		78 0.56
		79 0.56
		80 0.56
		81 0.56
		82 0.56
		83 0.56
		84 0.56
		85 0.56
		86 0.56
		87 0.56
		88 0.56
		89 0.56
		90 0.56
		91 0.56
		92 0.56
		93 0.56
		94 0.56
		95 0.56
		96 0.56
		97 0.56
		98 0.56
		99 0.56
	};
	\legend{DeepSynth, DQN, optimal}
	\end{axis}
	\end{tikzpicture}
	\subcaption{Task1}
%	\caption{Training progress for Task 1 with DeepSynth and DQN on the same training set $\mathcal{E}$. The expected return is over state $s_0=[4,4]$ with origin being the bottom left corner cell.}
	\label{fig:u_task_1}
\end{minipage}
\begin{minipage}{0.45\columnwidth}
%\end{figure}
%
%\begin{figure}[!t]
	\centering
	\begin{tikzpicture}[scale=0.6]
	\definecolor{oxford_blue}{rgb}{0,0.13,0.28}
	\begin{axis}[
	axis background/.style={fill=white},
	axis line style={black},
	tick align=outside,
	tick pos=left,
	x grid style={white!89.80392156862746!black},
	xlabel={Steps},
	xmajorgrids,
	xmin=-4.95, xmax=103.95,
	xtick style={color=white!33.33333333333333!black},
	y grid style={white!89.80392156862746!black},
	ylabel={Expected Discounted Reward},
	ymajorgrids,
	ymin=-0.0276098847900357, ymax=0.75,
	ytick style={color=white!33.33333333333333!black},
	legend style={draw=black, fill=white},
	]
	\addplot [line width=1.64pt, orange]
	table {%
		0 0
		1 0.0055391606874764
		2 0.0101741468533874
		3 0.0162445772439241
		4 0.0211450643837452
		5 0.0253632310777903
		6 0.031331978738308
		7 0.0365632697939873
		8 0.0417822897434235
		9 0.0563660226762295
		10 0.0617316029965878
		11 0.0725953802466393
		12 0.129080608487129
		13 0.171923384070396
		14 0.226868405938148
		15 0.267122447490692
		16 0.297746211290359
		17 0.342760384082794
		18 0.370725184679031
		19 0.388545960187912
		20 0.391941159963608
		21 0.393272280693054
		22 0.402005463838577
		23 0.412633329629898
		24 0.421711534261703
		25 0.421922028064728
		26 0.424282163381577
		27 0.434874624013901
		28 0.428990989923477
		29 0.433142304420471
		30 0.42646923661232
		31 0.431045889854431
		32 0.427598893642426
		33 0.427974909543991
		34 0.429705768823624
		35 0.433334589004517
		36 0.429769903421402
		37 0.427119135856628
		38 0.443396806716919
		39 0.438816249370575
		40 0.44710648059845
		41 0.454330325126648
		42 0.45017409324646
		43 0.460150241851807
		44 0.451624989509583
		45 0.449834525585175
		46 0.441674917936325
		47 0.454279690980911
		48 0.45059335231781
		49 0.443054050207138
		50 0.444962441921234
		51 0.446405529975891
		52 0.438038855791092
		53 0.434500157833099
		54 0.435018599033356
		55 0.438794016838074
		56 0.439934104681015
		57 0.445620119571686
		58 0.440281897783279
		59 0.435386747121811
		60 0.44429811835289
		61 0.458437383174896
		62 0.460712105035782
		63 0.467985689640045
		64 0.46846804022789
		65 0.465772777795792
		66 0.45034196972847
		67 0.451332092285156
		68 0.441737711429596
		69 0.429723739624023
		70 0.438534170389175
		71 0.438204497098923
		72 0.444971889257431
		73 0.449167430400848
		74 0.452163755893707
		75 0.463798433542252
		76 0.45759791135788
		77 0.456379622220993
		78 0.455846726894379
		79 0.453819185495377
		80 0.456498682498932
		81 0.450186789035797
		82 0.444735765457153
		83 0.441361576318741
		84 0.445408254861832
		85 0.452753156423569
		86 0.450239270925522
		87 0.453572809696198
		88 0.456322222948074
		89 0.449536442756653
		90 0.453307777643204
		91 0.46086797118187
		92 0.468745410442352
		93 0.467181593179703
		94 0.465776354074478
		95 0.473481416702271
		96 0.467015326023102
		97 0.475312620401382
		98 0.475285083055496
		99 0.469768226146698
	};
	\addplot [line width=1.64pt, oxford_blue]
	table {%
		0 0
		1 0.018956549741187
		2 0.00113102735453617
		3 0.00169743990317843
		4 0.016709977562589
		5 0.0147193997813705
		6 0.0133946080288044
		7 0.00616272915178288
		8 0.0121188833135692
		9 0.0121360346728168
		10 0.0116240803422401
		11 0.00316765740509611
		12 0.00861339280582537
		13 0.00787063640410743
		14 0.0144602416247493
		15 0.0198963912589949
		16 0.0189879094618649
		17 0.0108835409485864
		18 0.00889708377451707
		19 0.00536481483298656
		20 0.000718486587857152
		21 0.00054889714181638
		22 0.00929787724194624
		23 0.00636930255707355
		24 0.00760029843801423
		25 0.0178357891565657
		26 0.0105150553829206
		27 0.01121020722053
		28 0.00472246814230124
		29 0.000477161582815644
		30 0.0065028585752232
		31 0.00273394785972933
		32 0.010204476916744
		33 0.019973671363851
		34 0.0134895939469174
		35 0.00363686993646289
		36 0.0178714307316598
		37 0.0159351984284328
		38 0.0146880338378796
		39 0.0181318729979512
		40 0.0152577096766614
		41 0.0157949527492353
		42 0.00707573955683207
		43 0.0196195314614425
		44 0.0192380187579645
		45 0.00322369306608038
		46 0.0150800814330374
		47 0.0143030179647491
		48 0.00922813395483955
		49 0.0106071143224689
		50 0.00980027843700383
		51 0.0184966414418914
		52 0.0100168212526131
		53 0.0166304897958362
		54 0.00707848409737432
		55 0.0176570183716251
		56 0.0179940117751325
		57 0.00922024329763275
		58 0.0113541014084049
		59 0.0184066087838386
		60 0.0144754590774404
		61 0.0097321710972317
		62 0.00443622021982022
		63 0.00649334487537796
		64 0.0139914327614049
		65 0.00332139370988252
		66 0.0181588099325219
		67 0.00536275025799633
		68 0.0182275567173609
		69 0.00619126249898921
		70 0.0191472342311232
		71 0.0141241161273521
		72 0.0100849763396664
		73 0.010354955122971
		74 0.0130282879793358
		75 0.0117588942358897
		76 0.00623688649102
		77 0.00415636949075854
		78 0.0102378331671057
		79 0.0186830871826756
		80 0.0124653017345174
		81 0.00150750738148091
		82 0.0164079998942403
		83 0.014518985749546
		84 0.0181530724190264
		85 0.00382805466608235
		86 0.0148956544855471
		87 0.00117517792797311
		88 0.013058198548691
		89 0.00546199464674299
		90 0.00453233058489526
		91 0.0175098234289648
		92 0.0021253196529105
		93 0.0104472533071786
		94 0.0170788601436974
		95 0.00489663955938034
		96 0.00420957877391293
		97 0.0176116351873256
		98 0.00845835296779392
		99 0.0143392219780995
	};
	\addplot [line width=1.64pt, blue]
	table {%
		0 0.48
		1 0.48
		2 0.48
		3 0.48
		4 0.48
		5 0.48
		6 0.48
		7 0.48
		8 0.48
		9 0.48
		10 0.48
		11 0.48
		12 0.48
		13 0.48
		14 0.48
		15 0.48
		16 0.48
		17 0.48
		18 0.48
		19 0.48
		20 0.48
		21 0.48
		22 0.48
		23 0.48
		24 0.48
		25 0.48
		26 0.48
		27 0.48
		28 0.48
		29 0.48
		30 0.48
		31 0.48
		32 0.48
		33 0.48
		34 0.48
		35 0.48
		36 0.48
		37 0.48
		38 0.48
		39 0.48
		40 0.48
		41 0.48
		42 0.48
		43 0.48
		44 0.48
		45 0.48
		46 0.48
		47 0.48
		48 0.48
		49 0.48
		50 0.48
		51 0.48
		52 0.48
		53 0.48
		54 0.48
		55 0.48
		56 0.48
		57 0.48
		58 0.48
		59 0.48
		60 0.48
		61 0.48
		62 0.48
		63 0.48
		64 0.48
		65 0.48
		66 0.48
		67 0.48
		68 0.48
		69 0.48
		70 0.48
		71 0.48
		72 0.48
		73 0.48
		74 0.48
		75 0.48
		76 0.48
		77 0.48
		78 0.48
		79 0.48
		80 0.48
		81 0.48
		82 0.48
		83 0.48
		84 0.48
		85 0.48
		86 0.48
		87 0.48
		88 0.48
		89 0.48
		90 0.48
		91 0.48
		92 0.48
		93 0.48
		94 0.48
		95 0.48
		96 0.48
		97 0.48
		98 0.48
		99 0.48
	};
	\legend{DeepSynth, DQN, Optimal}
	\end{axis}
	\end{tikzpicture}
	\subcaption{Task3}
%	\caption{Training progress for Task 3 with DeepSynth and DQN on the same training set $\mathcal{E}$. The expected return is over state $s_0=[4,4]$ with origin being the bottom left corner cell.}
	\label{fig:u_task_3}
\end{minipage}
\caption{Training progress for Tasks 1 and 3 with DeepSynth and DQN on the same training set $\mathcal{E}$. The expected return is over state $s_0=[4,4]$ with origin being the bottom left corner cell.}
\end{figure*}

\begin{table*}[!t]
	\centering
		\begin{tabular}{|l|r|p{10.8cm}|}
			\hline
			\hline
			\colorbox{white}{Hyperparameter}& Value   & Description                                                                                                                                                                      \\ \hline
			minibatch size                  & 32      & Number of training cases over which each Stochastic Gradient Descent (SGD) update is computed                                                                                    \\ \hline
			replay memory size              & 150000  & SGD updates are sampled from this number of most recent frames.                                                                                                                  \\ \hline
			agent history length            & 4       & The number of most recent frame experienced by the agent that are given as input to the Q network                                                                                \\ \hline
			target network update frequency & 10000   & The frequency (measured in the number of parameter updates) with which the target network is updated                          \\ \hline
			discount factor                 & 0.99    & Discount factor gamma used in the Q-learning update                                                                                                                              \\ \hline
			%		action repeat                   & 4       & Repeat each action selected by the agent this many times. Using a value of 4 results in the agent seeing only every 4th input frame.                                             \\ \hline
			%		update frequency                & 4       & The number of actions selected by the agent between successive SGD updates. Using a value of 4 results in the agent selecting 4 actions between each pair of successive updates. \\ \hline
			learning rate                   & 0.00025 & The learning rate used by RMSProp                                                                                                                                                \\ \hline
			%		gradient momentum               & 0.95    & Gradient momentum used by RMSProp                                                                                                                                                \\ \hline
			%		squared gradient momentum       & 0.95    & Squared gradient (denominator) momentum used by RMSProp                                                                                                                          \\ \hline
			%		min squared gradient            & 0.01    & Constant added to the squared gradient in the denominator of the RMSProp update                                                                                                  \\ \hline
			initial exploration             & 1       & Initial value of $\epsilon$ in $\epsilon$-greedy exploration                                                                                           \\ \hline
			final exploration               & 0.1     & Final value of $\epsilon$ in $\epsilon$-greedy exploration                                                                                             \\ \hline
			final exploration frame         & 150,000 & The number of frames over which the initial value of $\epsilon$ is linearly annealed to its final value                                                             \\ \hline
			replay start size               & 8000    & A uniform random policy is run for this number of frames before learning starts and the resulting experience is used to populate the replay memory.                              \\ \hline
			no-op max                       & 30      & Maximum number of ``do nothing'' actions to be performed by the agent at the start of an episode.                                                                                  \\ \hline
			\hline
		\end{tabular}
	\caption{Hyper-parameters of the DQN Modules for Montezuma's Revenge}
	\label{tab:hyper_par}
\end{table*}

\begin{algorithm2e}[ht]
	\DontPrintSemicolon
	\SetKw{return}{return}
	\SetKwRepeat{Do}{do}{while}
	%\SetKwFunction{assume}{assume}
	%\SetKwFunction{isf}{isFeasible}
	\SetKwData{conflict}{conflict}
	\SetKwData{safe}{safe}
	\SetKwData{sat}{sat}
	\SetKwData{unsafe}{unsafe}
	\SetKwData{unknown}{unknown}
	\SetKwData{true}{true}
	\SetKwInOut{Input}{input}
	\SetKwInOut{Output}{output}
	\SetKwFor{Loop}{Loop}{}{}
	\SetKw{KwNot}{not}
	\begin{footnotesize}
		\Input{automaton $\mathfrak{A}$ from the Synth step}
		\Output{approximated optimal $Q$-function: $\hat{Q}^*$}
		initialise all neural nets $B_{q_i}$ with random weights\;
		\Repeat{end of trial}
		{
			initialise the state to $(s_0, q_0)$\;
			\For{$t = 1, total\_time\_steps$}
			{
				$a_t=\arg\max_a B_{q_t}$ with $\epsilon$-greedy\;
				execute action $a_t$ and observe the total reward $r^T_t$\;
				observe the next image $s_{t+1}$\;
				semantic segmentation outputs $L(s_{t+1})$\;
				update the automaton state from $q_t$ to $q_{t+1}$\;
				preprocess images $s_t$ and $s_{t+1}$ to $\mathcal{P}_t$ and $\mathcal{P}_{t+1}$\;
				store transition $(\mathcal{P}_t, a_t, \mathcal{P}_{t+1}, r^T_t, L(s_{t+1}))$ in $\mathcal{E}_{q_t}$\;
				sample minibatch from $\mathcal{E}_{q_t}$\;
				target value from target network $\hat{B}_{q_t}$, $y_i=r^T_i+\gamma \max\limits_{a'}Q(\mathcal{P}_{i+1},a'|\theta^{\hat{B}_{q_t}})$\;
				update $B_{q_i}$ weights using $y_i$\; 
				clone the current network $B_{q_i}$ to $\hat{B}_{q_t}$ every $C$ steps
			}
		}
	\end{footnotesize}
	\caption{DeepSynth with Temporal DQN}
	\label{dqnal}
\end{algorithm2e} 

\subsection{Montezuma's Revenge}

In Montezuma's Revenge, the emulator’s internal state is not visible to the
agent.  The agent only observes an image, which is a matrix of pixel values
that represent the current screen.  Inspired by~\cite{deepql}, we apply a
basic preprocessing step to the Atari 2600 frames, designed to reduce the
dimensionality of the input.  We extract the luminance from the RGB frame
and rescale it to $84 \times 84$.  Further, a grey-scaling pre-processing is
applied and then the $4$ most recent frames are stacked together to form
the input to a DQN module.

The agent selects and executes actions according to an $\epsilon$-greedy
policy.  Similar to the NFQ, the DQN module also uses experience replay,
which averages the behaviour distribution over many of previous states,
smoothing out learning and avoiding oscillations or divergence in the
parameters~\cite{nfq,deepql}.  Each DQN module stores a finite number of
last experience tuples in the replay memory, and samples minibatches
uniformly at random.  Further, to improve the stability of the deep nets, we
use a separate network in each module for generating the target $y$ values
in~\eqref{loss_function}.  Specifically, after a finite number of steps we
clone the module network to obtain a target network, which is used to
generate the Q-learning targets~$y$.  This modification makes the algorithm
more stable compared to standard QL and NFQ updates~\cite{deepql}.

An overview of the algorithm is presented as Algorithm~\ref{dqnal}.  The
values of all the hyper-parameters and descriptions of all hyper-parameters
are provided in Table~\ref{tab:hyper_par}.

\clearpage
\begin{figure*}[t]
\section{Automata Synthesised for Other Benchmarks}
\end{figure*}

\begin{figure*}[!ht]
\centering
\ttfamily
\small
\begin{minipage}{0.45\columnwidth}
		\centering
			\begin{tikzpicture}[->,>=stealth',shorten >=1pt,auto,node distance=2cm, thick] 
			\node[state,initial] (q_1)   {$q_1$}; 
			\node[state,accepting,fill=green] (q_2) [right=of q_1] {$q_2$}; 
			\node[state] (q_3) [below of=q_1,fill=red]   {$q_3$};
			
			\path[->] 
			(q_1) edge  node {t} (q_2)
			(q_1) edge[loop above]  node {$\emptyset$} (q_1)
			(q_1) edge  node {u} (q_3)
			(q_3) edge [loop left]  node {u} (q_3)
			;
			\end{tikzpicture}
\end{minipage}
\begin{minipage}{0.45\columnwidth}
		\centering
			\begin{tikzpicture}[->,>=stealth',shorten >=1pt,auto,node distance=2cm, thick] 
			\node[state,initial] (q_1)   {$q_1$}; 
			\node[state] (q_2) [right=of q_1] {$q_2$}; 
			\node[state,accepting,fill=green] (q_3) [right=of q_2] {$q_3$}; 
			\node[state] (q_4) [below=of q_1,xshift=1.5cm,yshift=1.5cm,fill=red] {$q_4$}; 
			\path[->] 
			(q_1) edge [loop above] node {$\emptyset$} ()   	
			(q_1) edge  node {t1} (q_2)
			(q_2) edge [loop above] node {$\emptyset$} ()
			(q_2) edge node {t2} (q_3)
			(q_1) edge node [anchor=east]{u} (q_4)
			(q_2) edge node [anchor=west]{u} (q_4)
			(q_4) edge [loop left] node {u} () ; 
			\end{tikzpicture}
\end{minipage}
\caption{Automata synthesised for {\ttfamily mars-rover-1} and {\ttfamily mars-rover-2}}
\label{fig:mars1}
\end{figure*}

\begin{figure*}
\centering
\ttfamily
\small
	\begin{minipage}{0.35\textwidth}
		\centering
			\begin{tikzpicture}[->,>=stealth',shorten >=1pt,auto,node distance=2cm, thick] 
			\node[state,initial] (q_1)   {$q_1$}; 
			\node[state,accepting,fill=green] (q_2) [right=of q_1] {$q_2$}; 
			
			\path[->] 
			(q_1) edge  node {obj1} (q_2)
			(q_1) edge[loop above]  node {$\emptyset$} (q_1)
			;
			\end{tikzpicture}
	\end{minipage}
	\begin{minipage}{0.55\textwidth}
		\centering
			\begin{tikzpicture}[->,>=stealth',shorten >=1pt,auto,node distance=3cm, thick]
			\tikzstyle{every state}=[fill=white,draw=black,text=black]
			\node[initial,state] (A)                    {$q_1$};
			\node[state]         (B) [right of=A] {$q_2$};
			\node[state,accepting,fill=green]         (E) [below of=B] {$q_5$};
			\node[state]         (D) [right of=E] {$q_4$};
			\node[state]         (C) [right of=B] {$q_3$};

			\path (A) edge [right]      node[anchor=south]{obj3} (B)
			(B) edge [bend left]            node[anchor=south]{obj2} (C)
			(C) edge [left]            node[anchor=north]{obj3} (B)
			(C) edge [bend left]            node[anchor=west]{obj1} (D)
			(D) edge [left]            node[anchor=east]{obj2} (C)
			(D) edge [left]            node[anchor=north]{obj4} (E)
			(B) edge [left]            node[anchor=east]{obj4} (E)
			(D) edge [bend left]            node[anchor=north,sloped]{obj3} (B)
			(B) edge [left]            node[anchor=south,sloped,xshift=-5mm]{obj1} (D)
			
			(A) edge [loop above]            node[anchor=south]{$\emptyset$} (A)
			(B) edge [loop above]            node[anchor=south]{$\emptyset$} (B)
			(C) edge [loop above]            node[anchor=south]{$\emptyset$} (C)
			(D) edge [loop below]            node[anchor=north]{$\emptyset$} (D)	
			;
			\end{tikzpicture}
\end{minipage}
\caption{Automata synthesised for {\ttfamily slp-easy} and {\ttfamily slp-hard}.\label{fig:slp-hard}}
\end{figure*}

\begin{figure*}[!ht]
	\centering
	\ttfamily
	\small
	\begin{minipage}{\columnwidth}
	\centering
		\begin{tikzpicture}[->,>=stealth',shorten >=1pt,auto,node distance=2cm, thick] 
		\node[state,initial] (q_1)   {$q_1$}; 
		\node[state] (q_2) [right=of q_1] {$q_2$}; 
		\node[state,accepting,fill=green] (q_3) [right=of q_2] {$q_3$}; 
		\path[->] 
		(q_1) edge [loop above] node {$\emptyset$} ()   	
		(q_1) edge  node {A} (q_2)
		(q_2) edge [loop above] node {$\emptyset$} ()
		(q_2) edge node {B} (q_3);
		\end{tikzpicture}
	\caption{Automaton synthesised for {\ttfamily robot-surve}.}
	\label{fig:robotsurve}
	\end{minipage}
\end{figure*}

\begin{figure*}[!ht]
\centering
\ttfamily
\small
%	\medskip
	\begin{minipage}{\columnwidth}
		\centering
			\begin{tikzpicture}[->,>=stealth',shorten >=1pt,auto,node distance=2cm, thick] 
			\node[state,initial] (q_1)   {$q_1$}; 
			\node[state,accepting,fill=green] (q_2) [right=of q_1] {$q_2$}; 
			\node[state] (q_3) [below of=q_1,fill=red]   {$q_3$};
			
			\path[->] 
			(q_1) edge  node {obj1} (q_2)
			(q_1) edge[loop above]  node {$\emptyset$} (q_1)
			(q_1) edge  node {u} (q_3)
			(q_3) edge [loop left]  node {u} (q_3)
			;
			\end{tikzpicture}
	\end{minipage}
	\medskip
	\begin{minipage}{\columnwidth}
		\centering
		\begin{figure}[H]
			\caption{Automaton synthesised for {\ttfamily frozen-lake-1,2,3}.}
			\label{fig:frozen123}
		\end{figure}
	\end{minipage}
	\medskip
	\begin{minipage}{\columnwidth}
		\centering
			\begin{tikzpicture}[->,>=stealth',shorten >=1pt,auto,node distance=3cm, thick]
			\tikzstyle{every state}=[fill=white,draw=black,text=black]
			\node[initial,state] (A)                    {$q_1$};
			\node[state]         (B) [right of=A] {$q_2$};
			\node[state,accepting,fill=green]         (E) [below of=B] {$q_5$};
			\node[state]         (D) [right of=E] {$q_4$};
			\node[state]         (C) [right of=B] {$q_3$};
			\node[state,fill=red]         (F) [right of=C] {$q_6$};
			\path (A) edge [right]      node[anchor=south]{obj3} (B)
			(B) edge [bend left]            node[anchor=south]{obj2} (C)
			(C) edge [left]            node[anchor=north]{obj3} (B)
			(C) edge [bend left]            node[anchor=west]{obj1} (D)
			(D) edge [left]            node[anchor=east]{obj2} (C)
			(D) edge [left]            node[anchor=north]{obj4} (E)
			(B) edge [left]            node[anchor=east]{obj4} (E)
			(D) edge [bend left]            node[anchor=north,sloped]{obj3} (B)
			(B) edge [left]            node[anchor=south,sloped,xshift=-5mm]{obj1} (D)
			(A) edge [bend left=45]      node[anchor=south]{u} (F)
			(B) edge [bend left=45]      node[anchor=south]{u} (F)
			(C) edge [left]      node[anchor=south]{u} (F)
			(D) edge [bend right]      node[anchor=west]{u} (F)
			
			(A) edge [loop above]            node[anchor=south]{$\emptyset$} (A)
			(B) edge [loop above]            node[anchor=south]{$\emptyset$} (B)
			(C) edge [loop above]            node[anchor=south]{$\emptyset$} (C)
			(D) edge [loop below]            node[anchor=north]{$\emptyset$} (D)
			(F) edge [loop right]            node[anchor=west]{u} (F)	
			;	
			\end{tikzpicture}
	\end{minipage}
	\medskip
	\begin{minipage}{\columnwidth}
		\centering
		\begin{figure}[H]
			\caption{Automaton synthesised for {\ttfamily frozen-lake-4,5,6}.}
			\label{fig:frozen456}
		\end{figure}
	\end{minipage}
\end{figure*}

\clearpage
\bibliography{paper_extended_bib} 
\end{document}
\else
\maketitle
\if\doctype0
\linenumbers
\fi
\begin{abstract}
This paper proposes DeepSynth, a method for effective training of deep
Reinforcement Learning (RL) agents when the reward is sparse and non-Markovian,
but at the same time progress towards the reward requires achieving an unknown
\textit{sequence} of high-level objectives. Our method employs a novel algorithm
for synthesis of compact automata to uncover this sequential structure
automatically. We synthesise a human-interpretable automaton from trace data
collected by exploring the environment. The state space of the environment is
then enriched with the synthesised automaton so that the generation of a control
policy by deep RL is guided by the discovered structure encoded in the
automaton. The proposed approach is able to cope  with both high-dimensional,
low-level features and unknown sparse non-Markovian rewards. We have evaluated
DeepSynth's performance in a set of experiments that includes the Atari game
\textit{Montezuma's Revenge}. Compared to existing approaches, we obtain a
reduction of \textit{two} orders of magnitude in the number of iterations
required for policy synthesis, and also a significant improvement in
scalability.
\end{abstract}
\section{Introduction}

Reinforcement Learning (RL) is the key enabling technique for a variety of
applications of artificial intelligence, including advanced
robotics~\cite{polydoros2017survey}, resource and traffic
management~\cite{rlresource,dorsa}, drone control~\cite{ng}, chemical
engineering~\cite{chemistry}, and gaming~\cite{deepql}.  While RL is a very
general architecture, many advances in the last decade have been achieved using
specific instances of RL that employ a deep neural network to synthesise optimal
policies.  A deep RL algorithm, AlphaGo~\cite{alphago}, played moves in the game
of Go that were initially considered glitches by human experts, but secured
victory against the world champion.  Similarly, AlphaStar~\cite{alphastar} was
able to defeat the world's best players at the real-time strategy game
StarCraft~II, and to reach top 0.2\% in scoreboards with an ``unimaginably
unusual'' playing style.

While deep RL can autonomously solve many problems in complex environments,
tasks that feature extremely sparse, non-Markovian rewards or other long-term
sequential structures are often difficult or impossible to solve by unaided~RL. 
A~well-known example is the Atari game \emph{Montezuma's Revenge}, in which
DQN~\cite{deepql} failed to score. Interestingly, Montezuma's Revenge and other
hard problems often require learning to accomplish, possibly in a specific
sequence, a set of high-level objectives to obtain the reward.  These objectives
can often be identified with passing through designated and semantically
distinguished states of the system.  This insight can be leveraged to obtain a
manageable, high-level model of the system's behaviour and its dynamics.

\textbf{Contribution:} In this paper we propose DeepSynth, a new algorithm that
automatically infers unknown sequential dependencies of a reward on high-level
objectives and exploits this to guide a deep RL agent when the reward signal is
history-dependent and significantly delayed.  We assume that these sequential
dependencies have a \emph{regular} nature, in formal language theory
sense~\cite{Gulwani2012SynthesisFE}.  The identification of dependency on a
sequence of high-level objectives is the key to breaking down a complex task
into a series of Markovian ones.  In our work, we use automata expressed in
terms of high-level objectives to orchestrate sequencing of low-level actions in
deep RL and to guide the learning towards sparse rewards.  Furthermore, the
automata representation allows a human observer to easily interpret the deep RL
solution in a high-level manner, and to gain more insight into the optimality of
that solution.

At the heart of DeepSynth is a \emph{model-free} deep RL algorithm that is
synchronised in a closed-loop fashion with an automaton inference algorithm,
enabling our method to learn a policy that discovers and follows high-level
sparse-reward structures.  The synchronisation is achieved by a product
construction that creates a hybrid architecture for the deep RL.  When dealing
with raw image input, we assume that an off-the-shelf unsupervised image
segmentation method, e.g.~\cite{obj_det_o}, can provide enough object candidates
\NewEdit{in order to identify semantically distinguished states.} We evaluate
the performance of DeepSynth on a selection of benchmarks with unknown
sequential high-level structures.  These experiments show that DeepSynth is able
to automatically discover and formalise unknown, sparse, and non-Markovian
high-level reward structures, and then to efficiently synthesise successful
policies in various domains where other related approaches fail.  DeepSynth
represents a better integration of deep RL and formal automata synthesis than
previous approaches, making learning for non-Markovian rewards more scalable.

\if\doctype3\noindent\textbf{Related Work:~}\fi\if\doctype2\section{Related Work}
\fi
Our research employs formal methods to deal with the sparse reward problem in
RL. In the RL literature, the dependency of rewards on objectives is often
tackled with \emph{options}~\cite{sutton}, or, in general, the dependencies are
structured \emph{hierarchically}.  Current approaches to Hierarchical RL (HRL) very
much depend on state representations and whether they are structured enough for
a suitable reward signal to be effectively engineered manually. HRL
therefore often requires detailed supervision in the form of explicitly
specified high-level actions or intermediate supervisory signals~\cite{precup,
	options_1, options_h_1, kulkarni2016hierarchical, options_h_2, bacon2017option}.
A key difference between our approach and HRL is that our method
produces a modular, human-interpretable and succinct graph to represent the
sequence of tasks, as opposed to complex and comparatively sample-inefficient
structures, e.g.~RNNs.

The closest line of work to ours, which aims to avoid HRL
requirements, are model-based~\cite{topku, dorsa, fulton3, cai2020receding} or model-free RL approaches that constrain the agent with a temporal
logic property~\cite{arxiv, toro, toro2, plmdp, deeplcrl, nonmarkov2, bolts,
	lcnfq, lcrl_j, cautiousRL, hasanbeig2020deep, kazemi2020formal,
	lavaei2020formal}.  These approaches are limited to finite-state systems, or
more importantly require the temporal logic formula to be known a~priori.  The latter
assumption is relaxed in \cite{toronto, rens2020online, rens2020learning,
	furelos2020induction, gaon2020reinforcement, xu2020joint}, by inferring automata from exploration traces.

Automata inference in \cite{toronto} uses a local-search based algorithm, Tabu
search~\cite{Tabu}.  \NewEdit{The automata inference algorithm that we employ
	uses SAT, where the underlying search algorithm is a backtracking search method
	called DPLL~\cite{DPLL}.} \NewEdit{In
	comparison with Tabu search, the DPLL algorithm is complete and explores the
	entire search space efficiently~\cite{DPLLvsTabu}, producing more accurate
	representations of the trace.}\if\doctype2~A detailed
comparison of the two approaches is provided in Section~\ref{sec:synth_vs_tabu}
of the Appendix. \fi~The Answer Set Programming (ASP) based algorithm used to
learn automata in~\cite{furelos2020induction}, also uses DPLL but assumes a
known upper bound for the maximum finite distance between automaton states.  We
further relax this restriction and assume that the task and its automaton are
entirely unknown.

A classic automata learning technique is the L* algorithm~ \cite{lstar}. This is
used to infer automata in~\cite{rens2020online, rens2020learning,
	gaon2020reinforcement, ckks2020}.  It employs a series of equivalence and
membership queries from an oracle, the results of which are used to construct
the automaton.  The absence of an oracle in our setting prevents the use of L*
in our method.

Another common approach for synthesising automata from traces is
\emph{state-merge}~\cite{ktails}.\if\doctype2~Variants of the state-merge
algorithm, e.g. Evidence-Driven State Merge (EDSM)~\cite{edsm}, use both
positive and negative instances of behaviour to determine equivalence of states
to be merged based on statistical evidence.\fi~\if\doctype2 To avoid
over-generalisation in the absence of labelled data, the EDSM algorithm was
improved to incorporate inherent temporal
behaviour~\cite{Walkinshaw:2007:RES:1339262.1339495,state_merge}, which needs to
be known a~priori. \fi \if\doctype3State-merge and some of its
variants~\cite{edsm, Walkinshaw:2007:RES:1339262.1339495}\fi \if\doctype2These
approaches, however,\fi~do not always produce the most succinct automaton but
generate an approximation that conforms to the trace~\cite{exact_fsm}. The
comparative succinctness of our inferred automaton allows DeepSynth to be
applied to large high-dimensional problems, including Montezuma's Revenge.
A~detailed comparison of these approaches can be found in\if\doctype3~the
extended version of this
work~\cite{arxiv_deepsynth}.\fi\if\doctype2~Section~\ref{sec:lrmvssynth} of the
Appendix.\fi

\NewEdit{A number of approaches combine SAT with state-merge to generate minimal
	automata from
	traces\if\doctype3~\cite{efsm_state_merge,Heule2013}\fi\if\doctype2~\cite{efsm_state_merge,Heule2013,exact_fsm,model_SAT, Buzhinsky2017ModularPM}\fi. A similar SAT based algorithm is employed in~\cite{xu2020joint} to generate reward machines. Although this approach generates succinct automata that accurately capture a rewarding sequence of events, it is not ideal for hard exploration problems such as Montezuma's Revenge where reaching a rewarding state, e.g. collecting the key, requires the agent to follow a sequence of non-rewarding steps that are difficult to discover via exploration. The automata learning algorithm we use is able to capture these non-rewarding sequences and leverage it to guide exploration towards the rewarding states.}

\if\doctype2 Inferred automata have been also used to learn strategies for
infinite two-person games where strategies are a function of previously visited
states. The construction of \emph{chain automata} for these games provides a
means to implement memory-less strategies~\cite{rabin_game}, though these chain
automata have a disproportionally large number of states as compared to the size
of the actual automaton the game is played on. \fi

Further related work is \emph{policy sketching}~\cite{pol-sketch}, which learns
feasible tasks first and then stitches them together to accomplish a complex
task.  The key difference to our work is that the method assumes policy
sketches,~i.e.~temporal instructions, to be available to the agent. There is
also recent work on learning underlying non-Markovian objectives when an optimal
policy or human demonstration is
available~\cite{koul2018learning,memarian2020active}.

\if\doctype2 The rest of this paper is organised as follows. We first review
fundamental notation and concepts in Section~\ref{sec:rl_background} and
Section~\ref{sec:synth}. DeepSynth technical details are then presented in
Section~\ref{sec:deepsynth}, where we illustrate our method with Montezuma's
Revenge as the running example. Finally, Section~\ref{sec:results} evaluates
DeepSynth performance in a set of numerical experiments including Montezuma's Revenge. \fi

\section{Background on Reinforcement Learning}\label{sec:rl_background}
We consider a conventional RL setup, consisting of an agent
interacting with an environment, which is modelled as an unknown general
Markov Decision Process (MDP):

\begin{definition}[General MDP]\label{mdpdef}
The tuple $\mathfrak{M}=(\mathcal{S}, \allowbreak \mathcal{A}, \allowbreak
s_0,\allowbreak P,\allowbreak \Sigma,\allowbreak L)$ is a general MDP over a
set of continuous states $\mathcal{S}$, where $\mathcal{A}$ is
a finite set of actions and $s_0 \in \mathcal{S}$ is the initial state. 
$P:\mathcal{B}(\mathcal{S})\times\mathcal{S}\times\mathcal{A}\rightarrow
[0,1]$ is a Borel-measurable conditional transition kernel that assigns to
any pair of state $s \in \mathcal{S}$ and action~$a \in \mathcal{A}$ a
probability measure $P(\cdot|s,a)$ on the Borel space
$(\mathcal{S},\mathcal{B}(\mathcal{S}))$, where $\mathcal{B}(\mathcal{S})$ is the Borel sigma-algebra on the state space~\cite{shreve}. $\Sigma$ is called
the \textit{vocabulary set} and it is a finite set of atomic
propositions. There exists a labelling function $L: \mathcal{S}
\rightarrow 2^\Sigma$ that assigns to each state $s \in \mathcal{S}$ a set
of atomic propositions $L(s) \in 2^\Sigma$. 
\end{definition} 

\begin{definition}[Path] 
In a general MDP $\mathfrak{M}$, an infinite path $\rho$ starting at $s_0$
is an \NewEdit{infinite} sequence of state \NewEdit{transitions} $s_0 \xrightarrow{a_0} s_1
\xrightarrow{a_1}~... ~$ such that every transition $s_i \xrightarrow{a_i}
s_{i+1}$ is possible in $\mathfrak{M}$, i.e. $s_{i+1}$ belongs to the
smallest Borel set $B$ such that $P(B|s_i,a_i)=1$. \NewEdit{Similarly,} a finite path is a \NewEdit{finite} sequence of state \NewEdit{transitions} $\rho_n=
s_0 \xrightarrow{a_0} s_1 \xrightarrow{a_1} ... \xrightarrow{a_{n-1}} s_n$.
The set of infinite paths is
$(\mathcal{S}\times\mathcal{A})^\omega$ and the set of finite paths
is $(\mathcal{S}\times\mathcal{A})^*\times\mathcal{S}$.
\end{definition}

At each state $s \in \mathcal{S}$, an agent action is determined by a policy
$\pi$, which is a mapping from states to a probability distribution over the
actions. That is, $\pi: \mathcal{S} \rightarrow \mathcal{P}(\mathcal{A})$. 
Further, a random variable
$R(s,a)\sim\varUpsilon(\cdot|s,a)\in\mathcal{P}(\mathds{R})$ is defined over
the MDP~$\mathfrak{M}$, to represent the Markovian reward obtained when
action $a$ is taken in a given state~$s$, where $\mathcal{P}(\mathds{R})$ is
the set of probability distributions on subsets of $\mathds{R}$, and
$\varUpsilon$ is the reward distribution. Similarly, a non-Markovian reward
$\widehat{R}: (\mathcal{S}\times\mathcal{A})^*\times\mathcal{S}\rightarrow
\mathds{R}$ is a mapping from the set of finite paths to real numbers and
one possible realisation of $R$ and $\widehat{R}$ at time step $n$ is
denoted by $r_n$ and $\widehat{r}_n$ respectively.

\if\doctype2

\fi\if\doctype3
Due to space limitations we present the formal background on RL
in\if\doctype3~\cite{arxiv_deepsynth}~\fi\if\doctype2~the Appendix~\fi and
we only introduce the notation we use.  The expected discounted
return for a policy $\pi$ and state $s$ is denoted by ${U}^{\pi}(s)$, which
is maximised by the optimal policy $\pi^*$.  Similarly, at each state the
optimal policy maximises the Q-function ${Q}(s,a)$ over the set of actions. 
The Q-function can be parameterised using a parameter set $\theta^Q$ and
updated by minimising a loss $\mathfrak{L}(\theta^Q)$.
\fi

\section{Background on Automata Synthesis}\label{sec:synth}

\if\doctype2{}This section describes the algorithm used for automatic inference of unknown
high-level sequential structures as automata.\fi{} The automata synthesis
algorithm extracts information from trace sequences over finite paths in order to construct a succinct automaton that
represents the behaviour exemplified by these traces. This architecture is an instance of the general \emph{synthesis from
examples} approach~\cite{Gulwani2012SynthesisFE,synth}.\if\doctype2~It is a scalable method for learning finite-state automaton from trace data that produces abstract, concise models.\fi~Our synthesis method scales to long traces by employing a segmentation approach, achieving
automata learning in close-to-polynomial runtime~\cite{t2m}.\if\doctype2{} Experimental evidence indicates a linear growth in runtime for increasing
trace length in the case of segmented trace inputs, as compared to 
exponential growth for non-segmented trace inputs. Our algorithm learns succinct models
from traces without the additional information that is typically required by the
state-merge algorithm~\cite{Biermann:1972:SFM:1638603.1638997}.
\fi

The synthesis algorithm takes as input a trace sequence and generates an
$N$-state automaton conforming to the trace input.  Starting with $N=1$, the
algorithm systematically searches for the required automaton, 
incrementing $N$ by $1$ each time the search fails.  This ensures that the
smallest automaton conforming to the input trace is generated.
\NewEdit{The algorithm additionally uses a hyper-parameter $w$ to
tackle growing algorithm complexity for long trace input.}
The synthesis algorithm divides the trace into segments
using a sliding window of size $w$ and \NewEdit{only} unique segments are used for further
processing.  \NewEdit{In this way}, the algorithm exploits the presence of
patterns in traces.  Multiple occurrences of these patterns are processed
only once, thus reducing the size of the input to the algorithm.

\begin{figure}[!t]
	\centering
	\includegraphics[width=\columnwidth]{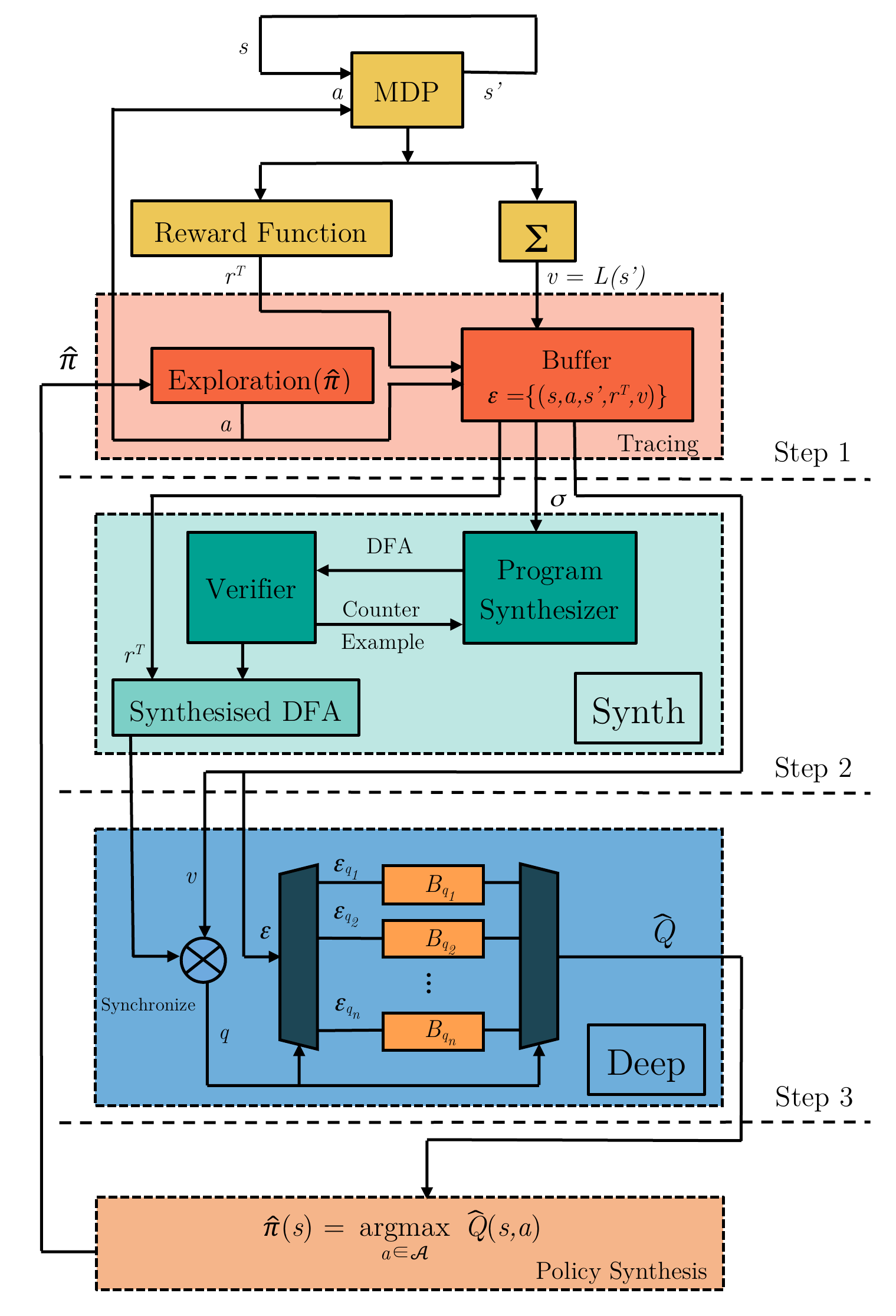}
	\caption{The DeepSynth Algorithm}
	\label{fig:bird_eye_view}
\end{figure}

Automata generated using only \emph{positive} trace samples tend to
overgeneralise~\cite{Gold1978ComplexityOA}.  This is mitigated by
performing a compliance check of the automaton against the trace input to
eliminate any transition sequences of length $l$ that are accepted by the
generated automaton but do not appear in the trace.  The hyper-parameter $l$
therefore controls the degree of generalisation in the generated automaton. 
A higher value for $l$ yields more exact representations of the trace.  The
correctness of the generated automaton is verified by checking if the
automaton accepts the input trace.  If the check fails, missing trace data
is incrementally added to refine the generated model, until the check
passes. Further details on tuning the hyper-parameters $w$ and $l$ are given in
the next section.

%An automaton can be expressed as a computer program, as illustrated in Fig.~\ref{fig:automaton}, and in what follows we use the terms ``program'' and ``automaton'' interchangably.

%\comment{The automaton obtained by the synthesis framework is deterministic. because at any given point in time the agent reads a single label from the current MDP state. These labels get recorded in the traces and are eventually expressed as transition predicate on the edges of the automaton.}

\section{DeepSynth}\label{sec:deepsynth}

\begin{figure}[!t]\centering
	\begin{minipage}{0.45\columnwidth}
	
	{\begin{tikzpicture}
			\node at (0,0) {\includegraphics[width=0.9\columnwidth]{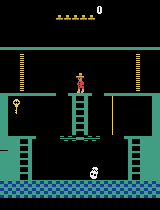}};
			\end{tikzpicture}}
	\subcaption{}
	\end{minipage}
	\quad
	\begin{minipage}{0.45\columnwidth}
	{\begin{tikzpicture}
			\node at (0,0) {\includegraphics[width=0.9\columnwidth]{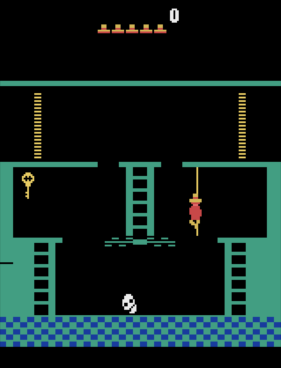}};
			\fill[color=yellow!30!white,opacity=0.4] (0.69,-0.35) circle (0.5cm);
			\end{tikzpicture}}
	\subcaption{}
	\end{minipage}
	\caption{(a) the first level of Atari 2600 Montezuma's Revenge; 
		(b) pixel overlap of two segmented objects.}
	\label{mont_initial}
\end{figure}

A schematic of the DeepSynth algorithm
is provided in Fig.~\ref{fig:bird_eye_view} and the algorithm is described
step-by-step in this section. We~begin by introducing the first level of Montezuma's Revenge as a running
example~\cite{ale}.  Unlike other Atari games where the primary goal is
limited to avoiding obstacles or collecting items with no particular order,
Montezuma's Revenge requires the agent to perform a long, complex sequence
of actions before receiving any reward.  The agent must find a key and open
either door in Fig.~\ref{mont_initial}.a.  To~this end, the agent has to
climb down the middle ladder, jump on the rope, climb down the ladder on the
right and jump over a skull to reach the key.  The reward given by the Atari
emulator for collecting the key is $100$ and the reward for opening one of the
doors is another $300$.  Owing to the sparsity of the rewards the existing
deep RL algorithms either fail to learn a policy that can even reach the
key, e.g.~DQN~\cite{deepql}, or the learning process is computationally
heavy and sample inefficient,~e.g.~FeUdal~\cite{feudal}, and
Go-Explore~\cite{goexplore}.

Existing techniques to solve this problem mostly
hinge on intrinsic motivation and object-driven guidance. 
% However, none of these approaches are able to incorporate prior domain knowledge into the learning process. As shown in details later, thanks to the intuitive structure of the automaton, DeepSynth can embed such information to facilitate the learning process.
Unsupervised object detection (or unsupervised semantic segmentation) from
raw image input has seen substantial progress in recent years, and became
comparable to its supervised
counterpart~\cite{obj_det_o,obj_det,obj_det_2,obj_det_3}.  In this work, we
assume that an off-the-shelf image segmentation algorithm can provide plausible
object candidates, e.g.~\cite{obj_det_o}. The key to solving a complex task
such as Montezuma's Revenge is finding the semantic correlation between the
objects in the scene.  When a human player tries to solve
this game the semantic correlations, such as ``keys open doors'', are partially known and the player's behaviour is driven by exploiting
these known correlations when exploring unknown objects.  This \NewEdit{drive to explore unknown objects} has been a
subject of study in psychology, where animals and humans seem to have
general motivations (often referred to as intrinsic motivations) that push
them to explore and manipulate their environment, encouraging curiosity and
cognitive growth~\cite{intrinsic2, intrinsic3, intrinsic}.

\NewEdit{As explained later, DeepSynth encodes these correlations as an automaton,
which is an intuitive and modular structure, and guides the
exploration so that previously unknown correlations are discovered.}
This exploration scheme imitates biological cognitive growth
in a formal and explainable way, and is driven by an intrinsic motivation to
explore as many objects as possible in order to find the optimal sequence
of extrinsically-rewarding high-level objectives. To \NewEdit{showcase} the full potential
of DeepSynth, in all the experiments and examples of this paper we assume
that semantic correlations are unknown to the agent.  The agent starts with
no prior knowledge of the sparse reward task or the correlation of the
high-level objects.

Let us write $\Sigma$ for the set of detected objects. Note that the
semantics of the names for individual objects is of no relevance
to the algorithm and $\Sigma$ can thus contain any distinct identifiers,
e.g. $\Sigma=\{\mathtt{obj}_1, \mathtt{obj}_2, \ldots\}$.
But for the sake of exposition we name the
objects according to their appearance in~Fig.~\ref{mont_initial}.a, i.e.
$\Sigma=\{\,\mathtt{red\_character},\,\, \mathtt{middle\_ladder},
\allowbreak\mathtt{rope}, \allowbreak\mathtt{right\_ladder},
\allowbreak\mathtt{left\_ladder}, \allowbreak\mathtt{key},
\allowbreak\mathtt{door}\}$.  Note that there can be any number of detected
objects, as long as the input image is segmented into enough objects whose
correlation can guide the agent to achieve the task.

\if\doctype3\noindent\textbf{Tracing (Step 1 in Fig.~\ref{fig:bird_eye_view}):~}\fi\if\doctype2
\subsection{Tracing (Step 1 in Fig.~\ref{fig:bird_eye_view})}
\fi
Note that the task is unknown initially and the extrinsic reward is
non-Markovian and extremely sparse. The agent receives a
reward $\widehat{R}: (\mathcal{S}\times\mathcal{A})^*\times\mathcal{S}\rightarrow \mathds{R}$
only when a correct sequence of state-action pairs and their associated
object \NewEdit{correlations} are visited.  \NewEdit{In order to guide the
agent to find the optimal sequence, DeepSynth uses
the following reward transformation:}

%Thus, at each time step the agent is free to select any object as its intrinsic goal \comment{Wait a minute - now we have intrinsic `goals'? What are \underline{those}?}  in the hope \comment{If the `hope' is to get the reward, then selecting this object is driven by \textit{extrinsic motivation} not intrinsic motivation. Be precise.} to find the optimal extrinsic rewarding task sequence. The total reward is composed of the extrinsic non-Markovian reward and an intrinsic automaton-based reward \comment{Hang on - now we have `intrinsic' \underline{rewards}. The language is very sloppy here.}
%
\begin{equation}\label{eq:reward}
	r^T = \widehat{r} + \mu ~r^{i},
\end{equation}
where \NewEdit{$\widehat{r}$ is the extrinsic reward,} $\mu>0$ is a positive
\NewEdit{regulatory} coefficient, \NewEdit{and $r^i$ is the intrinsic
reward}.  \NewEdit{The role of the intrinsic reward is to guide the
exploration and also to drive the exploration towards the discovery of
unknown object correlations.} The underlying mechanism of
intrinsic rewards depends on the inferred automaton and is explained in
detail later.  The only extrinsic rewards in Montezuma's Revenge are the
reward for reaching the key~$\widehat{r}_\mathit{key}$ and for reaching one of
the doors~$\widehat{r}_\mathit{door}$. Note that the lack of intrinsic
motivation as shown in \NewEdit{Section~\ref{sec:results}}, prevents other methods,
e.g.~\cite{toronto,rens2020online,gaon2020reinforcement,xu2020joint}, to succeed
in extremely-spare reward, high-dimensional and large problems such as
Montezuma's Revenge.
\if\doctype2~DeepSynth pseudo-code is in the Appendix.\fi

In Montezuma's Revenge, states consist of raw pixel images. 
Each state is a stack of four consecutive frames $84 \times 84 \times 4$
that are preprocessed to reduce input dimensionality \cite{deepql}. The
labelling function \NewEdit{employs} the object vocabulary set $\Sigma$ to detect object
pixel overlap in a particular state frame. For example, if the pixels of
$\mathtt{red\_character}$ collide with the pixels of $\mathtt{rope}$ in any
of the stacked frames, the labelling function for that particular state $s$
is $L(s){=}\{\mathtt{red\_character},\, \mathtt{rope}\}$
(Fig.~\ref{mont_initial}.b). In this specific example,
the only moving object is the character. So for sake of succinctness,
we omit the character from the label set,
e.g., the above label is $L(s){=}\{\mathtt{rope}\}$.

\if\doctype2
\begin{figure*}[!t]
\else
\begin{figure}[!t]
\fi
	\centering
\if\doctype2
	\includegraphics[width=0.6\columnwidth]{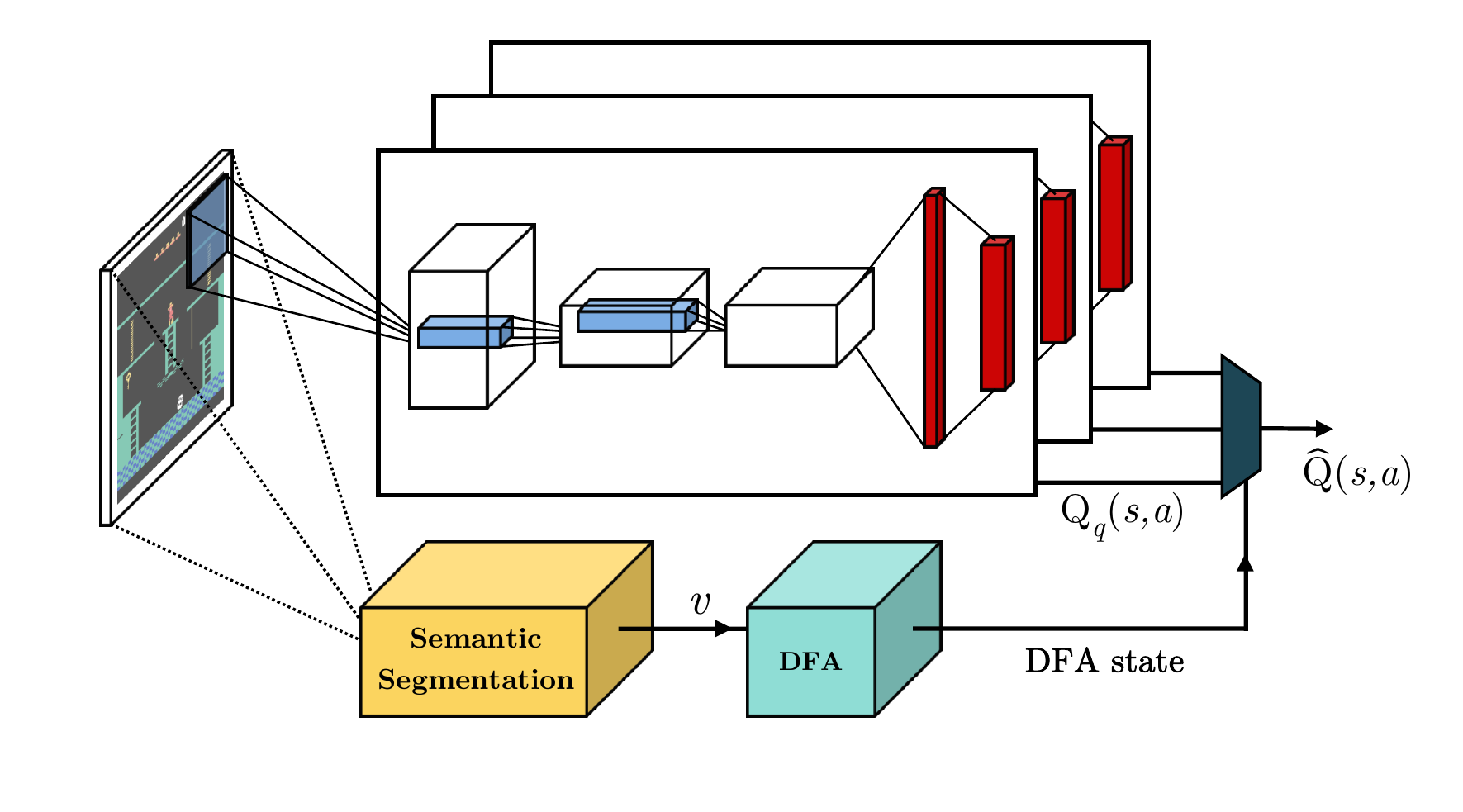}
\else
	\includegraphics[width=\columnwidth]{bird_eye_view_dqn.pdf}
\fi
	\caption{DeepSynth for Montezuma's Revenge: each DQN module is
		forced by the DFA to focus on the correlation of semantically distinct objects.
		%\if\doctype3~The structure of the DQN modules is discussed further in \cite{arxiv_deepsynth}.\fi
		\if\doctype3~The input to
		the first layer of the DQN modules is the input image which is convolved by $32$
		filters of $8 \times 8$ with stride $4$ and a ReLU. The second hidden
		layer convolves $64$ filters of $4 \times 4$ with stride $2$, followed by a
		ReLU. This is followed by another convolutional layer that convolves $64$
		filters of $3 \times 3$ with stride $1$ followed by a rectifier. The final
		hidden layer is fully connected and consists of $512$ ReLUs and the output
		layer is a fully-connected linear layer with a single output for each
		action~\cite{deepql}.\fi}
	\label{fig:bird_eye_view_dqn}
\if\doctype2
\end{figure*}
\else
\end{figure}
\fi

Given this labelling function, Tracing (\NewEdit{Step 1}) records the sequence of detected
objects $L(s_i)L(s_{i+1}) \ldots$ as the agent explores the MDP.  The labelling
function, as per Definition~\ref{mdpdef}, is a mapping from the state space
to the power set of objects in the vocabulary $L:\mathcal{S}\rightarrow
2^\Sigma$ and thus, the label of a state could be the empty set or a set of objects. 

All transitions with their corresponding labels are stored as 5-tuples
$\langle s,a,{s}',r^T,L(s') \rangle$, where $s$ is the current state, $a$~is the
executed action, $s'$ is the resulting state, $r^T$ is the total reward
received after performing action $a$ at state $s$, and $L(s')$ is the label
corresponding to the set of atomic propositions in $\Sigma$ that hold in state~$s'$. 
The set of past experiences is called the experience replay buffer
$\mathcal{E}$. The exploration process generates a set of \textit{traces},
defined as follows:

\begin{definition}[Trace] 
In a general MDP $\mathfrak{M}$, and over a finite path $\rho_n= s_0
\xrightarrow{a_0} s_1 \xrightarrow{a_1} \ldots \xrightarrow{a_{n-1}} s_n$, a
trace $\sigma$ is defined as a sequence of labels \NewEdit{$\sigma = v_1, v_2, \dots, v_n $}, where $v_{i}=L(s_i)$ is a trace event.
\end{definition}

\noindent The set of traces associated with $\mathcal{E}$ is denoted by
$\mathcal{T}$. The tracing scheme is the Tracing box
in Fig.~\ref{fig:bird_eye_view}.

\if\doctype2
\newcommand\myresizebox[3]{\resizebox{\textwidth}{!}{#3}}
\else
\newcommand\myresizebox[3]{\resizebox{#1}{#2}{#3}}
\fi

\begin{figure}[!t]
	\centering
	\ttfamily
	\small
	\begin{minipage}{\columnwidth}
	{{
			\myresizebox{0.7\textwidth}{!}{
				\begin{tikzpicture}[->,>=stealth',shorten >=1pt,auto,node distance=3cm, thick]
				\tikzstyle{every state}=[fill=white,draw=black,text=black]
				\node[initial,state] (A)                    {$q_1$};
				\node[state]         (B) [below right of=A] {$q_2$};
				\node[state]         (C) [above right of=B] {$q_3$};
				\node[state]         (D) [right of=C] {$q_4$};
				
				\path (A) edge [right]      node[anchor=north,sloped]{middle\_ladder} (B)
				(A) edge [right]            node[anchor=south]{rope} (C)
				(B) edge [bend left=20]            node[anchor=south,sloped]{rope} (C)
				(C) edge [bend left=20]            node[anchor=north,sloped]{middle\_ladder} (B)
				(C) edge [right]            node[anchor=south]{right\_ladder} (D)
				
				(A) edge [loop above]            node[anchor=south]{$\emptyset$} (A)
				(B) edge [loop below]            node[anchor=north]{$\emptyset \, \vee \,$middle\_ladder} (B)
				(C) edge [loop above]            node[anchor=south]{$\emptyset \, \vee \,$rope} (C)
				(D) edge [loop above]            node[anchor=south]{$\emptyset \, \vee \,$right\_ladder} (D)
				;
				\end{tikzpicture}}}}
	\subcaption{The right ladder is often discovered by random exploration}
	\end{minipage}

	\medskip
	
	\begin{minipage}{\columnwidth}
	{{
			\myresizebox{0.9\textwidth}{!}{
				\begin{tikzpicture}[->,>=stealth',shorten >=1pt,auto,node distance=3cm, thick]
				\tikzstyle{every state}=[fill=white,draw=black,text=black]
				\node[initial,state] (A)                    {$q_1$};
				\node[state]         (B) [below right of=A] {$q_2$};
				\node[state]         (C) [above right of=B] {$q_3$};
				\node[state]         (D) [right of=C] {$q_4$};
				\node[state]         (E) [below of=D] {$q_5$};
				\node[state]         (F) [left of=E] {$q_6$};
				
				\path (A) edge [right]      node[anchor=north,sloped]{middle\_ladder} (B)
				(A) edge [right]            node[anchor=south]{rope} (C)
				(B) edge [bend left=20]            node[anchor=south,sloped]{rope} (C)
				(C) edge [bend left=20]            node[anchor=north,sloped]{middle\_ladder} (B)
				(C) edge [right]            node[anchor=south]{right\_ladder} (D)
				(D) edge [bend right]            node[anchor=east]{left\_ladder} (E)
				(E) edge [bend right]            node[anchor=west]{right\_ladder} (D)
				(E) edge [bend left]            node[anchor=north]{key} (F)
				(F) edge [bend left]            node[anchor=south]{left\_ladder} (E)
				
				(A) edge [loop above]            node[anchor=south]{$\emptyset$} (A)
				(B) edge [loop below]            node[anchor=north]{$\emptyset \, \vee \,$middle\_ladder} (B)
				(C) edge [loop above]            node[anchor=south]{$\emptyset \, \vee \,$rope} (C)
				(D) edge [loop above]            node[anchor=south]{$\emptyset \, \vee \,$right\_ladder} (D)
				(E) edge [loop below]            node[anchor=north]{$\emptyset \, \vee \,$left\_ladder} (E)
				(F) edge [loop below]            node[anchor=north]{$\emptyset$} (F)
				%(G) edge [loop below]            node[anchor=north]{$\emptyset$} (G)
				;
				\end{tikzpicture}}}}\\
	\subcaption{The key is found with an extrinsic reward of $\widehat{r}_\mathit{key}=+100$}
	\end{minipage}

	\medskip
	
	\begin{minipage}{\columnwidth}
	{{
			\myresizebox{0.9\textwidth}{!}{
				\begin{tikzpicture}[->,>=stealth',shorten >=1pt,auto,node distance=3cm, thick]
				\tikzstyle{every state}=[fill=white,draw=black,text=black]
				\node[initial,state] (A)                    {$q_1$};
				\node[state]         (B) [below right of=A] {$q_2$};
				\node[state]         (C) [above right of=B] {$q_3$};
				\node[state]         (D) [right of=C] {$q_4$};
				\node[state]         (E) [below of=D,yshift=-10mm] {$q_5$};
				\node[state]         (F) [left of=E] {$q_6$};
				\node[state,accepting,fill=green]         (G) [left of=F] {$q_7$};
				
				\path (A) edge [right]      node[anchor=north,sloped]{middle\_ladder} (B)
				(A) edge [right]            node[anchor=south]{rope} (C)
				(B) edge [bend left=20]            node[anchor=south,sloped]{rope} (C)
				(C) edge [bend left=20]            node[anchor=north,sloped]{middle\_ladder} (B)
				(C) edge [right]            node[anchor=south]{right\_ladder} (D)
				(D) edge [bend right]            node[anchor=east]{left\_ladder} (E)
				(E) edge [bend right]            node[anchor=west]{right\_ladder} (D)
				(E) edge [bend left]            node[anchor=north]{key} (F)
				(F) edge [bend left]            node[anchor=south]{left\_ladder} (E)
				(F) edge [left]            node[anchor=north]{door} (G)
				
				(A) edge [loop above]            node[anchor=south]{$\emptyset$} (A)
				(B) edge [loop below]            node[anchor=north]{$\emptyset \, \vee \,$middle\_ladder} (B)
				(C) edge [loop above]            node[anchor=south]{$\emptyset \, \vee \,$rope} (C)
				(D) edge [loop above]            node[anchor=south]{$\emptyset \, \vee \,$right\_ladder} (D)
				(E) edge [loop below]            node[anchor=north]{$\emptyset \, \vee \,$left\_ladder} (E)
				(F) edge [loop below]            node[anchor=north]{$\emptyset$} (F)
				%(G) edge [loop below]            node[anchor=north]{$\emptyset$} (G)
				;
				\end{tikzpicture}}}}
	\subcaption{The door is unlocked with an extrinsic reward of $\widehat{r}_\mathit{door}=+300$}
	\end{minipage}
	\caption{Illustration of the evolution of the automaton synthesised
        for Montezuma's Revenge.  Note that the agent found a
        short-cut to reach the key by skipping the middle ladder and
        directly jumping over the rope, which is not obvious even to a human
        player.  Such observations are difficult to extract from other
        hierarchy representations, e.g.~LSTMs.}
	\label{fig:montezuma_dfa}
\end{figure}

\if\doctype3\noindent\textbf{Synth (Step 2 in Fig.~\ref{fig:bird_eye_view}):~}\fi\if\doctype2
\subsection{Synth (Step 2 in Fig.~\ref{fig:bird_eye_view})}
\fi
The automata synthesis algorithm described in Section~\ref{sec:synth} is used to
generate an automaton that conforms to the trace sequences generated
by Tracing (\NewEdit{Step~1}). Given a trace sequence \NewEdit{$\sigma = v_1, v_2, \dots, v_n $}, the labels $v_{i}$ serve as transition predicates in the
generated automaton. The synthesis algorithm further constrains the
construction of the automaton so that no two transitions from a given state in the
generated automaton have the same predicates. The automaton obtained by the
synthesis algorithm is thus deterministic. The learned automaton follows
the standard definition of a Deterministic Finite Automaton (DFA) with the
alphabet $\Sigma_\mathfrak{A}$, where a symbol of the alphabet $v \in
\Sigma_\mathfrak{A}$ is given by the labelling function $L: \mathcal{S}
\rightarrow 2^\Sigma$ defined earlier. Thus, given a trace sequence \NewEdit{$\sigma = v_1, v_2, \dots, v_n $} over a finite path $\rho_n= s_0 \xrightarrow{a_0} s_1
\xrightarrow{a_1} ... \xrightarrow{a_{n-1}} s_n ~$ in the MDP, the symbol
$v_{i} \in \Sigma_\mathfrak{A}$ is given by $v_{i}=L(s_i)$.

The Atari emulator provides the number of lives left in the game, which is
used to reset $\Sigma_\mathfrak{a} \subseteq
\Sigma_\mathfrak{A}$, where $\Sigma_\mathfrak{a}$ is the set of labels that
appeared in the trace so far.  Upon losing a life,
$\Sigma_\mathfrak{a}$ is reset to the empty set.

\begin{definition}[Deterministic Finite Automaton] \label{gba_definition}
A~DFA $\mathfrak{A}\allowbreak=(\allowbreak\mathcal{Q},\allowbreak
q_0,\allowbreak\Sigma_\mathfrak{A}, \allowbreak{F}, \allowbreak\delta)$ is a
5-tuple, where $\mathcal{Q}$ is a finite set of states, $q_0 \in
\mathcal{Q}$ is the initial state, $\Sigma_\mathfrak{A}$ is the alphabet,
${F}\subset\mathcal{Q}$ is the set of accepting states, and $\delta:
\mathcal{Q} \times \Sigma_\mathfrak{A} \rightarrow \mathcal{Q}$ is the
transition function.
\end{definition}

Let $\Sigma_\mathfrak{A}^*$ be the set of all finite words over
$\Sigma_\mathfrak{A}$.  A finite word $w = v_{1},v_{2},\ldots,v_{m} \in
\Sigma_\mathfrak{A}^*$ is accepted by a DFA $\mathfrak{A}$ if there exists a
finite run $\theta \in \mathcal{Q}^*$ starting from $\theta_0 = q_0$, where
$\theta_{i+1} = \delta(\theta_i,v_{i+1})$ for $i \geq 0$ and $\theta_m \in
{F}$.  Given the collected traces $\mathcal{T}$ we construct a DFA using the
method described in Section~\ref{sec:synth}.  The algorithm first divides
the trace into segments using a sliding window of size equal to the
hyper-parameter $w$ introduced earlier.  This determines the size of the
input to the search procedure, and consequently the algorithm runtime.  Note
that choosing $w=1$ will not capture any sequential behaviour.  In
DeepSynth, we would like to have a value for $w$ that results in the smallest input size. 
\NewEdit{In our experiments, we tried different values for $w$ in increasing
order, ranging within $1 < w \le |\sigma|$, and have obtained the same
automaton in all setups.}

%The strategy we adopt for our experiments is to fix a segment length $w=2$ to ensure quick results. 

As discussed in Section~\ref{sec:synth}, the hyper-parameter $l$ controls
the degree of generalisation in the learnt automaton.  Learning exact
automata from trace data is known to be
NP-complete~\cite{Gold1978ComplexityOA}.  Thus, a higher value for $l$
increases the algorithm runtime.  We optimise over the hyper-parameters and
choose $w=3$ and $l=2$ as the best fit for our setting.  This ensures that
the automata synthesis problem is not too complex for the synthesis
algorithm to solve but at the same time it does not over-generalise to fit
the trace.

The generated automaton provides deep insight into the correlation of
the objects detected in Step 1 and shapes the intrinsic reward.  The output of
this stage is a DFA, from the set of succinct DFAs obtained earlier. 
Fig.~\ref{fig:montezuma_dfa} gives three exemplars of the evolution of
the synthesised automata for Montezuma's Revenge. 
Most of the deep RL approaches are able to reach the states that correspond
to the DFA in Fig.~\ref{fig:montezuma_dfa}.a via random exploration.  However, reaching
the key and further the doors as in Fig.~\ref{fig:montezuma_dfa}.b and
Fig.~\ref{fig:montezuma_dfa}.c is challenging and is achieved \NewEdit{by DeepSynth} using a
hierarchical curiosity-driven learning method described next. 
The automata synthesis is the Synth box in Fig.~\ref{fig:bird_eye_view}
and implementation details can be found in~\cite{arxiv_deepsynth}.

\if\doctype3\noindent\textbf{Deep Temporal Neural Fitted RL (Step 3 in Fig.~\ref{fig:bird_eye_view}):~}\fi\if\doctype2
\subsection{Deep Temporal Neural Fitted RL (Step 3 in Fig.~\ref{fig:bird_eye_view})}
\fi
We propose a deep-RL-based \NewEdit{architecture} inspired by DQN~\cite{deepql}
and Neural Fitted $Q$-iteration (NFQ)~\cite{nfq} when the input is in
vector form, not a raw image. DeepSynth is able to
synthesise a policy whose traces are accepted by the DFA and it encourages the
agent to explore under the DFA guidance. More importantly, \NewEdit{the agent is guided and encouraged} to expand
the DFA towards task satisfaction. 

Given the constructed DFA, at each time step during the learning episode, if a new
label is observed during exploration, the intrinsic reward in \eqref{eq:reward} becomes
positive. Namely,
\begin{equation}\label{gamma}
R^i(s,a) = \left\{
\begin{array}{ll}
\eta & $ if $ L(s')\notin\Sigma_\mathfrak{a},\\
0 & $ otherwise, 
$
\end{array}
\right.
\end{equation}
where $\eta$ is an arbitrarily finite and positive reward, and
$\Sigma_\mathfrak{a}$, as discussed in the Synth step, is the set of
labels that the agent has observed \emph{in the current learning episode}. 
Further, once a new label that does not belong to $\Sigma_\mathfrak{A}$ is
observed during exploration (Step~1) it is then passed to the automaton
synthesis \NewEdit{step} (Step~2).  The automaton synthesis algorithm then synthesises a new DFA
that complies with the new label. 

\if\doctype2
For example, in Montezuma’s revenge, the agent will get intrinsic reward every time it reaches the middle ladder for the first time during a trace, not just on the first trace it reaches that ladder. Furthermore, note that the intrinsic reward is used to motivate exploration and to discover new label objects. The overall task is initially unknown, thus \emph{bad} high-level actions cannot be inferred solely from the labels. Note that whenever the agent is intrinsically rewarded towards a bad behaviour, the contribution from the extrinsic reward is much higher than that from the intrinsic reward (adjusted by $\mu$ in \eqref{gamma}). Thus, the agent is able to escape local intrinsic optima and backpropagate the extrinsic reward to rule out bad behaviours and find the extrinsic optimal policy. In principle, if a high-level objective could be labelled as \emph{bad}, namely if the task is partially known, then the intrinsic reward could as well account for that.
\fi

\begin{theorem}\label{thm:policy_inv}\textbf{\textup{(Formalisation of the Intrinsic Reward)}} The optimal policies are invariant under the reward transformation in \eqref{eq:reward} and \eqref{gamma}. 
\end{theorem}
\if\doctype2
\noindent\textbf{Proof.~}
We leverage the concept of reward shaping to prove that the generated policies under the intrinsic reward transformation are guaranteed to be optimal with respect to the original extrinsic reward function \cite{reward_shaping_2,reward_shaping}. \cite{reward_shaping_2} proved that given any MDP $\mathfrak{M}=(\mathcal{S}, \allowbreak \mathcal{A}, \allowbreak
s_0,\allowbreak P,\allowbreak \Sigma,\allowbreak L)$, if there exists a potential function $\varPhi:\mathcal{S}\rightarrow\mathds{R}$, such that it transforms the reward function $r$ of MDP $\mathfrak{M}$ to
$$
r'(s,a,s')=r(s,a,s')+\gamma\varPhi(s')-\varPhi(s),
$$
then the set of optimal policies are invariant under $r'$. We define the potential function as 
\begin{equation}\label{eq:potential}
\varPhi(s) = \left\{
\begin{array}{ll}
\mu(\eta/\gamma) & $ if $ L(s)\notin\Sigma_\mathfrak{a},\\
0 & $ otherwise.
$
\end{array}
\right.
\end{equation}
Note that the potential function is bounded, and \eqref{eq:reward} can be
formalised as a formal reward shaping, in which the intrinsic reward $r^i$
is proved not to be part of the original optimisation problem.
\hfill$\Box$
\fi

\if\doctype3{}The proof of Theorem~\ref{thm:policy_inv} is presented in \cite{arxiv_deepsynth}.~\fi In the following, in order to
explain the core ideas underpinning the algorithm, we temporarily assume
that the MDP graph and the associated transition probabilities are fully
known. Later we relax these assumptions, and we stress that the algorithm
can be run \textit{model-free} over any black-box MDP environment.
%The recorded sequences play a critical role in breaking down a non-Markovian task into a set of Markovian, history-independent goals. 
% However, once a high-level task is completed, what the agent records into the replay buffer $\mathcal{E}$ is just the final reward $R$. As stated before, along the way of performing that high-level task, the agent also records state-action pairs and their corresponding label. The sequence of labels acts as a memory for the trace-dependent reward and allows one to convert it to a Markovian reward with which we can employ RL. Further, the final reward categorises the traces into different sets, each associated to a high-level task. 
Specifically, we relate the black-box MDP and the automaton by synchronising them
\emph{on-the-fly} to create a new structure
that breaks down a non-Markovian task into a set of Markovian,
history-independent sub-goals.

\begin{definition} [Product MDP]
	\label{product_mdp_def} Given an MDP $\mathfrak{M}=(\mathcal{S}, \allowbreak \mathcal{A}, \allowbreak s_0,\allowbreak P,\allowbreak \Sigma)$ and
	a DFA $\mathfrak{A}\allowbreak=(\allowbreak\mathcal{Q},\allowbreak
	q_0,\allowbreak\Sigma_\mathfrak{A}, \allowbreak{F}, \allowbreak\delta)$, the product MDP is defined as $(\mathfrak{M}\otimes
	\mathfrak{A}) = \mathfrak{M}_\mathfrak{A}=(\mathcal{S}^\otimes,\allowbreak
	\mathcal{A},\allowbreak s^\otimes_0,P^\otimes,\allowbreak
	\Sigma^\otimes,\allowbreak
	{F}^\otimes)$, where $\mathcal{S}^\otimes =
	\mathcal{S}\times\mathcal{Q}$, $s^\otimes_0=(s_0,q_0)$,
        $\Sigma^\otimes = \mathcal{Q}$, and ${F}^\otimes=\mathcal{S}\times F$.
The transition kernel $P^\otimes$ is such that given the current state
$(s_i,q_i)$ and action~$a$, the new state $(s_j,q_j)$ is given by $s_j\sim
P(\cdot|s_i,a)$ and $q_j=\delta(q_i,L(s_j))$.
\end{definition} 

By synchronising MDP states with the DFA states by means of the product MDP, we
can evaluate the satisfaction of the associated high-level task. Most
importantly, as shown in~\cite{nonmarkov}, for any MDP $\mathfrak{M}$
with finite-horizon non-Markovian reward, e.g.~Montezuma's Revenge,
there exists a Markov reward MDP $\mathfrak{M}'=(\mathcal{S}, \allowbreak
\mathcal{A}, \allowbreak s_0,\allowbreak P,\allowbreak \Sigma)$ that is
equivalent to $\mathfrak{M}$ such that the states of $\mathfrak{M}$ can be
mapped into those of $\mathfrak{M}'$. The corresponding states yield
the same transition probabilities, and corresponding traces have the same
rewards. Based on this result,~\cite{nonmarkov2} showed that the product
MDP $\mathfrak{M}_\mathfrak{A}$ is $\mathfrak{M}'$ defined above. 
Therefore, the non-Markovianity of the extrinsic reward is resolved by synchronising
the DFA with the original MDP, where the DFA represents the history of state
labels that has led to that reward. 
%This allows one to run RL over theproduct MDP and to find the optimal policy that maximises the corresponding Markovian reward.

Note that the DFA transitions can be executed just by observing the labels
of the visited states, which makes the agent aware of the automaton state
without explicitly constructing the product MDP.  This means that the
proposed approach can run \emph{model-free}, and as such it does not require
a~priori knowledge about the MDP. 

Each state of the DFA in the synchronised product MDP divides the general
sequential task so that each transition between the states represents an
achievable Markovian sub-task. Thus, given a synthesised DFA
$\mathfrak{A}\allowbreak=(\allowbreak\mathcal{Q},\allowbreak
q_0,\allowbreak\Sigma_\mathfrak{A}, \allowbreak{F}, \allowbreak\delta)$, we
propose a hybrid architecture of $n=|\mathcal{Q}|$ separate deep RL modules
(Fig.~\ref{fig:bird_eye_view_dqn} and Deep in
Fig.~\ref{fig:bird_eye_view}). \NewEdit{For each state in the DFA, there is a dedicated deep RL module, where each deep RL module is an instance of a deep RL algorithm with distinct neural networks and replay buffers. The modules are interconnected, in the sense that} modules act as a global
\emph{hybrid} deep RL architecture to approximate the $Q$-function in the
product MDP. \NewEdit{As explained in the following,} this allows the agent to jump from one sub-task to another by
just switching between these modules as prescribed by the
DFA.

In the running example, the agent exploration scheme is $\epsilon$-greedy
with diminishing $\epsilon$ where the rate of decrease also depends on the
DFA state so that each module has enough chance to explore. For each automaton state $q_i \in \mathcal{Q}$ in the
the product MDP,  the associated deep RL module is called $B_{q_i}(s,a)$. Once the agent is at state
$s^\otimes=(s,q_i)$, the neural net $B_{q_i}$ is active and explores the MDP. 
Note that the modules are \NewEdit{interconnected}, as discussed above. For example, assume that by taking
action $a$ in state $s^\otimes=(s,q_i)$ the label $v=L(s')$ has been
observed and as a result the agent is moved to state ${s^\otimes}'=(s',q_j)$,
where $ q_i\neq q_j $. By minimising the loss function $\mathfrak{L}$
%(Equation \eqref{loss_function} in the Appendix)
the weights of $B_{q_i}$
are updated such that $B_{q_i}(s,a)$ has minimum possible error to
$R^T(s,a)+\gamma\max_{a'} B_{q_j}({s}',a')$ while $B_{q_i} \neq B_{q_j}$. 
As such, the output of $B_{q_j}$ directly affects $B_{q_i}$ when the
automaton state is changed. This allows the extrinsic reward to
back-propagate efficiently, e.g. from modules $B_{q_7}$ and $B_{q_6}$
associated with $q_7$ and $q_6$ in Fig.~\ref{fig:montezuma_dfa}.c, to the
initial module $B_{q_1}$% corresponding to $q_1$
.

Define $\mathcal{E}_{q_i}$ as the projection of the general replay buffer $\mathcal{E}$ onto $q_i$. 
The size of the replay buffer for each module is limited and in the case of
our running example $|\mathcal{E}_{q_i}|=15000$. This includes the most recent
frames that are observed when the product MDP state was
$s^\otimes=(s,q_i)$. In the running example we used RMSProp for each module
with uniformly sampled mini-batches of size~$32$. When the state is in
vector form and no convolutional layer is involved we resort to NFQ deep RL
modules instead of DQN modules. 

\section{Experimental Results}\label{sec:results}
\if\doctype3\noindent\textbf{Benchmarks and Setup:~}\fi\if\doctype2
\subsection{Benchmarks and Setup}
\fi
We evaluate and compare the performance of DeepSynth with DQN on a comprehensive set of
benchmarks, given in Table~\ref{tab:benchmarks}.  The Minecraft environment
($\mathtt{minecraft\text{-}tX}$) taken from~\cite{pol-sketch} requires
solving challenging low-level control tasks, and features highly sequential
high-level goals. The two $\mathtt{mars\text{-}rover}$ benchmarks are taken
from~\cite{lcnfq}, and the models have uncountably infinite
state spaces.  The example $\mathtt{robot\text{-}surve}$ is adopted
from~\cite{dorsa}, and the task is to visit two regions in sequence.  Models $\mathtt{slp\text{-}easy}$ and
$\mathtt{slp\text{-}hard}$ are inspired by the noisy MDPs of Chapter~6
in~\cite{sutton}.  The goal in $\mathtt{slp\text{-}easy}$ is to reach a
particular region of the MDP and the goal in $\mathtt{slp\text{-}hard}$ is
to visit four distinct regions sequentially in proper order.  The
$\mathtt{frozen\text{-}lake}$ benchmarks are similar: the first three are
simple reachability problems and the last three require sequential visits of
four regions, except that now there exist unsafe regions as well.  The
$\mathtt{frozen\text{-}lake}$ MDPs are stochastic and are adopted from
the OpenAI Gym~\cite{gym}.

The \emph{task DFA} column in Table~\ref{tab:benchmarks} gives the number of
states in the automaton that can be generated from the high-level objective
sequences of the ground-truth task.  The \emph{synth DFA} column gives the
number of states of the automaton synthesised by DeepSynth, and
\emph{prod.~MDP} gives the number of states in the resulting product MDP
(Definition~\ref{product_mdp_def}).  Finally, \emph{max sat.~prob.~at $s_0$}
is the maximum probability of achieving the extrinsic reward from the
initial state. In all experiments the high-level objective sequences are initially unknown
to the agent.  Furthermore, the
extrinsic reward is only given when completing the task and reaching the objectives in the
correct order.  The details of all experiments, including hyperparameters, are given
in\if\doctype3~\cite{arxiv_deepsynth,deepsynth}\fi\if\doctype2~the
Appendix\fi.

%All simulations have been carried out on a machine with an Intel Xeon 3.5\,GHz processor and 16\,GB of RAM, running Ubuntu~18.

\newcolumntype{R}{>{$}r<{$}}
\newcolumntype{T}{>{\ttfamily}l<{}}

\if\doctype2
\newcommand\mytableresizebox[3]{#3}
\else
\newcommand\mytableresizebox[3]{\resizebox{#1}{#2}{#3}}
\fi

\if\doctype2
\begin{table*}[!t]
\else
\begin{table}[!t]
\fi
	\centering
	{
		\small
		\mytableresizebox{\textwidth}{!}{%
			\begin{tabular}{|T|R|c|c|R|c|c|c|}
				\hline
				\multirow{ 2}{*}{\textrm{experiment}} &\multirow{ 2}{*}{~~$|\mathcal{S}|$~~} & task & synth & \textrm{prod.}& \textrm{max sat.}& \textrm{DeepSynth} & \textrm{DQN} \\
				& &\textrm{DFA} & \textrm{DFA} & \textrm{MDP} & prob. at $s_0$ & conv. ep$.^*$ & conv. ep$.^*$ \\ \hline
				minecraft-t1  &  100 & 3 & 6 &  600 & 1 & 25 & 40\\
				minecraft-t2  &  100 & 3 & 6 &  600 & 1 & 30 & 45\\
				minecraft-t3  &  100 & 5 & 5 &  500 & 1 &  40 & t/o\\
				minecraft-t4  &  100 & 3 & 3 &  300 & 1 & 30 & 50\\
				minecraft-t5  &  100 & 3 & 6 &  600 & 1 & 20 & 35\\
				minecraft-t6  &  100 & 4 & 5 &  500 & 1 & 40 & t/o\\
				minecraft-t7  &  100 & 5 & 7 &  800 & 1 & 70 & t/o\\
				mars-rover-1  &\infty& 3 & 3 &\infty& n/a & 40 & 50\\
				mars-rover-2  &\infty& 4 & 4 &\infty& n/a & 40 & t/o\\
				robot-surve   &   25 & 3 & 3 &   75 & 1 & 10 & 10\\
				slp-easy-sml  &  120 & 2 & 2 &  240 & 1 & 10 & 10\\
				slp-easy-med  &  400 & 2 & 2 &  800 & 1 & 20 & 20\\
				slp-easy-lrg  & 1600 & 2 & 2 & 3200 & 1 & 30 & 30 \\
				slp-hard-sml  &  120 & 5 & 5 &  600 & 1 & 80 & t/o\\
				slp-hard-med  &  400 & 5 & 5 & 2000 & 1 & 100  & t/o \\
				slp-hard-lrg  & 1600 & 5 & 5 & 8000 & 1 & 120 & t/o\\
				frozen-lake-1 &  120 & 3 & 3 &  360 & 0.9983 & 100 & 120\\
				frozen-lake-2 &  400 & 3 & 3 & 1200 & 0.9982 & 150 & 150\\
				frozen-lake-3 & 1600 & 3 & 3 & 4800 & 0.9720 & 150 & 150\\
				frozen-lake-4 &  120 & 6 & 6 &  720 & 0.9728 & 300 & t/o\\
				frozen-lake-5 &  400 & 6 & 6 & 2400 & 0.9722 & 400 & t/o\\
				frozen-lake-6 & 1600 & 6 & 6 &9600 & 0.9467 & 450 & t/o\\
				\hline              
	\end{tabular}}}
	\caption{Comparison between DeepSynth and DQN\if\doctype3\vspace{-3mm}\fi}
	{\vspace{2mm} * average number of episodes to convergence over 10 runs}
	\label{tab:benchmarks}
\if\doctype2
\end{table*}
\else
\end{table}
\fi

\if\doctype3\noindent\textbf{Results:~}\fi\if\doctype2
\subsection{Results}
\fi

\input{figure_MR} 
\if\doctype3
\input{figure_MC}
\fi
\if\doctype2
\begin{figure}[!t]
\centering
	\begin{tikzpicture}[scale=0.7]
	%	\definecolor{color0}{rgb}{0.75,0,0.75}
	\definecolor{oxford_blue}{rgb}{0,0.13,0.28}
	\begin{axis}[
	axis background/.style={fill=white},
	axis line style={black},
	tick align=outside,
	tick pos=left,
	x grid style={white!89.80392156862746!black},
	xlabel={Steps},
	xmajorgrids,
	xmin=-4.95, xmax=103.95,
	xtick style={color=white!33.33333333333333!black},
	y grid style={white!89.80392156862746!black},
	ylabel={Loss},
	ymajorgrids,
	ymin=-0.04, ymax=1.05,
	ytick style={color=white!33.33333333333333!black},
	legend style={draw=black, fill=white}
	]
	\addplot [line width=1.64pt, orange]
	table {%
		0 0.0219974718391983
		1 0.00189497959765653
		2 0
		3 0.000906162101425027
		4 0.000872798931744083
		5 0.0019594045874813
		6 0.00311768502190989
		7 0.0188322896285152
		8 0.113640830067244
		9 0.282939709388352
		10 0.514819577410236
		11 0.902620645685925
		12 0.835450913744719
		13 0.849803306636028
		14 1
		15 0.620136526650263
		16 0.622401557756216
		17 0.421188243172349
		18 0.254522780763907
		19 0.183077500323138
		20 0.158314537471929
		21 0.13570968464083
		22 0.0900510701765381
		23 0.085301504654935
		24 0.0785587585289398
		25 0.076890180418157
		26 0.0667220483414757
		27 0.0640463432890918
		28 0.0610492970911719
		29 0.0531263653718212
		30 0.0575015767838292
		31 0.0658844327113103
		32 0.0647737209341279
		33 0.0625981330586676
		34 0.0603434366554289
		35 0.0623908516852282
		36 0.0483382486209102
		37 0.0531218923151253
		38 0.0494563516378463
		39 0.0506000958767886
		40 0.0453129112725503
		41 0.0499927982590995
		42 0.050977240388877
		43 0.0500230635977737
		44 0.0431982182136362
		45 0.0416703964844275
		46 0.0431192940499365
		47 0.0424342106581494
		48 0.0387003345091131
		49 0.0377931715028839
		50 0.0358631220976944
		51 0.0383032224140282
		52 0.0440072245381724
		53 0.0498154770339403
		54 0.0447095508602017
		55 0.0442563974061295
		56 0.0442055585940536
		57 0.0500565138153318
		58 0.050265637350969
		59 0.0498373975037875
		60 0.0494242453641012
		61 0.0452957448991756
		62 0.0450192555414657
		63 0.0428252525639273
		64 0.0389235560144666
		65 0.0387726646706917
		66 0.0444168791834289
		67 0.0431930307616336
		68 0.0501070771110487
		69 0.0467723423767068
		70 0.0442317177188028
		71 0.0448645843278006
		72 0.039080538460197
		73 0.0379157184875176
		74 0.0354189726036569
		75 0.0337190280188751
		76 0.0341470760773157
		77 0.0325897120668148
		78 0.0334643249921898
		79 0.032437826188372
		80 0.0340336762825764
		81 0.0368644868403279
		82 0.0351405806510906
		83 0.036124459297301
		84 0.0311453514655808
		85 0.0345026220561112
		86 0.0390174682162321
		87 0.0437615862893644
		88 0.0416413520697665
		89 0.0392125564706872
		90 0.0361700354931935
		91 0.0360068769353516
		92 0.0333918841504097
		93 0.0333089540562063
		94 0.0325783453240001
		95 0.035580800171766
		96 0.0313448835891514
		97 0.0321000403750567
		98 0.0330433178687179
		99 0.0288823756984654
	};
	\addplot [line width=1.64pt, red]
	table {%
		0 0.00864234024647059
		1 0
		2 0.000937900527175273
		3 0.00922995030990268
		4 0.035611850579697
		5 0.135163420985722
		6 0.419153623949563
		7 0.797534878193328
		8 0.99
		9 0.84053344959494
		10 0.826132022582378
		11 0.848511565640329
		12 0.790496145499474
		13 0.679491393605454
		14 0.612921872534113
		15 0.530914669353344
		16 0.465949783666173
		17 0.355167643437352
		18 0.248900471794269
		19 0.21685432690641
		20 0.149896711067824
		21 0.149158953727419
		22 0.128409165561266
		23 0.101953134496922
		24 0.0921271821795723
		25 0.0921534792251195
		26 0.0920660435225799
		27 0.112195245172139
		28 0.110496938063392
		29 0.11138539156858
		30 0.108497230657226
		31 0.0981153332465544
		32 0.0938279565379797
		33 0.108616140694634
		34 0.110051440548829
		35 0.110718485633159
		36 0.0898074189877533
		37 0.0884879710730735
		38 0.0787864620719034
		39 0.084078445379975
		40 0.0816682292388798
		41 0.0789294494301557
		42 0.0647601127137202
		43 0.0662696470795122
		44 0.0706504298529633
		45 0.0776342868176092
		46 0.0784252642778984
		47 0.0871079631583549
		48 0.0788834791209375
		49 0.0850482224868128
		50 0.0783825999225139
		51 0.0856720505435502
		52 0.0844467977589292
		53 0.0991792801183931
		54 0.106245202512411
		55 0.0991873544113814
		56 0.099099374540341
		57 0.102803356209641
		58 0.0920220078052647
		59 0.0874467054677075
		60 0.0834110781247223
		61 0.0817337649547796
		62 0.0945820923843558
		63 0.0943829889869492
		64 0.10399634536155
		65 0.0884138192904141
		66 0.078195921097295
		67 0.0736568764454254
		68 0.0808085678459566
		69 0.0751716044666618
		70 0.0766513465456401
		71 0.0765912558166009
		72 0.0731632840945614
		73 0.068455784149889
		74 0.067996245325212
		75 0.0708335143282486
		76 0.0740539184566304
		77 0.0803249977258467
		78 0.0861652102663627
		79 0.0919329168256159
		80 0.0986635241916202
		81 0.0944403171225853
		82 0.0979482767739276
		83 0.101701797152881
		84 0.0957601427903384
		85 0.0819880012734439
		86 0.0837920480593645
		87 0.0783633364135052
		88 0.0844907658862384
		89 0.0827970281234514
		90 0.0792853567596844
		91 0.0843023790409799
		92 0.0763227290105166
		93 0.0661412905736569
		94 0.0666261457088366
		95 0.0626032048565481
		96 0.0762272600580626
		97 0.0738764312223583
		98 0.0743838808890086
		99 0.0676258535514504
	};
	\addplot [line width=1.64pt, cyan]
	table {%
		0 0.00989081825592384
		1 0.608118612920054
		2 0.522941483267005
		3 0.45463353494612
		4 0.245923120731229
		5 0.9
		6 0.810314097877729
		7 0.565930363252504
		8 0.659335651707327
		9 0.797586895706883
		10 0.62024078121105
		11 0.47700583245478
		12 0.324409528312561
		13 0.213854183967685
		14 0.129046449372162
		15 0.0935290597160753
		16 0.0719829252422544
		17 0.0562216836156985
		18 0.0471376353986296
		19 0.0387763446321561
		20 0.0353787021512887
		21 0.034726254385767
		22 0.0266689832718646
		23 0.0251885955181299
		24 0.0206333531701117
		25 0.0136106951345889
		26 0.0123706503973877
		27 0.0139494861419395
		28 0.0157359859697341
		29 0.0202313821191468
		30 0.0212002705538863
		31 0.0183062540687078
		32 0.0134824477607028
		33 0.0125979356535329
		34 0.0131668445074561
		35 0.0092809721216348
		36 0.0129408207790234
		37 0.0106927354088025
		38 0.0128588424276185
		39 0.00674995595959702
		40 0.00813737618403789
		41 0.00536911451374637
		42 0.00492225530877994
		43 0.00543576464084643
		44 0.00914207994389291
		45 0.00223786547592836
		46 0.00879176901731022
		47 0.00629358249831715
		48 0.0193635147356342
		49 0.0136090813966036
		50 0.0095607846269669
		51 0.0135331140937718
		52 0.0132083405127988
		53 0.0142254317332874
		54 0.0115030255984427
		55 0.00854474947155487
		56 0.0145788964673339
		57 0.016380868766126
		58 0.0102973824187202
		59 0.0107746616594539
		60 0.014696133265908
		61 0.00787354312575404
		62 0.00262598901760689
		63 0.00404869712859842
		64 0.00934680487105591
		65 0.016721393227874
		66 0.0127269210966935
		67 0.00504876368822966
		68 0.00710362747504549
		69 0.00383818181063979
		70 0.00214456986933007
		71 0.00451895271285959
		72 0.00654656787065781
		73 0.00897447745494359
		74 0.00857421394047558
		75 0.0110199916779951
		76 0.0114552270734137
		77 0.00358699228935473
		78 0.00329501733227772
		79 0
		80 0.0139690145226309
		81 0.00863288814515055
		82 0.00992152834276661
		83 0.00701505053850568
		84 0.00783861322814526
		85 0.00978021067216244
		86 0.00739782690407003
		87 0.00970138440959947
		88 0.00528180082222758
		89 0.0104435037959008
		90 0.00385055852457489
		91 0.00583207251200344
		92 0.00617423896421866
		93 0.00790271773329749
		94 0.0146347393747345
		95 0.00773475553682871
		96 0.0164343387497172
		97 0.0150721640852567
		98 0.00916299668245827
		99 0.00480466064613712
	};
	\addplot [line width=1.64pt, teal]
	table {%
		0 0.846507738562942
		1 1
		2 0.381818709247165
		3 0.038548378129984
		4 0.0172899675734227
		5 0.0103963042932547
		6 0.00731274763888972
		7 0.00817566374033483
		8 0.00791815477350432
		9 0.00562333335600643
		10 0.00731512431319416
		11 0.00421180781308664
		12 0.00123195112056568
		13 0.000706751712891235
		14 0.000471383157830484
		15 0.000377287724407175
		16 0.000355932011933475
		17 0.000260288750414126
		18 0.000216204102242873
		19 0.00018761742996316
		20 0.000174779328373982
		21 0.000144983589979023
		22 0.000139048034425897
		23 0.000104955187376041
		24 9.1493267931788e-05
		25 8.64575285430004e-05
		26 9.32657015715194e-05
		27 7.33202772318895e-05
		28 6.78173526817374e-05
		29 5.39698998126336e-05
		30 3.82789409261336e-05
		31 4.70811726080277e-05
		32 3.2132875937684e-05
		33 2.87978734161124e-05
		34 2.7593481955694e-05
		35 3.40402072431342e-05
		36 4.25673553216888e-05
		37 1.26848260573879e-05
		38 8.24680312983383e-05
		39 9.91152154979975e-06
		40 7.81937718437213e-06
		41 5.0390191053606e-06
		42 0
		43 3.82760635910285e-06
		44 3.04929387083954e-05
		45 4.82001250033122e-05
		46 7.4038575138734e-05
		47 0.000191431021237572
		48 0.000235179750314184
		49 0.000151309344670031
		50 0.000124915176927642
		51 0.000151629824784591
		52 0.000231483813482166
		53 0.000154083922270951
		54 0.000166357214167888
		55 0.000199715285291695
		56 0.000146024954865104
		57 0.000263141281141484
		58 0.000289419916709104
		59 0.000576681229544678
		60 0.000370036492719829
		61 0.000272752453863325
		62 0.000634780909139024
		63 0.000689768957233849
		64 0.000685422449831545
		65 0.000274914337106199
		66 0.000118750024494032
		67 0.000208876720343558
		68 0.000282605947029629
		69 0.000297521587334832
		70 0.000273918001377953
		71 0.000295224086052632
		72 0.000365462700814418
		73 0.00132347766064677
		74 0.000303848436898333
		75 0.000118737475179447
		76 0.000145456721441961
		77 8.47324318426414e-05
		78 0.000311626251153642
		79 0.000382609482887321
		80 0.000296542790919266
		81 0.00069376176419825
		82 0.000485507255325707
		83 0.000420455206949497
		84 0.00047123621341189
		85 0.000537101044682343
		86 0.000156748806072103
		87 0.000478775344878417
		88 0.000431475095433162
		89 0.000189947028844631
		90 0.00028317853848082
		91 0.000243859434811655
		92 0.000164918203675997
		93 0.000312189307349088
		94 0.000498452973043782
		95 0.000514775483066015
		96 0.000426015343641151
		97 0.00014536498003537
		98 0.000394372971752186
		99 0.000220766211859343
	};
	\legend{$q_1$, $q_2$,$q_3$, $q_4$}
	\end{axis}
	\end{tikzpicture}
	
	\begin{tikzpicture}[scale=0.7]
	\definecolor{oxford_blue}{rgb}{0,0.13,0.28}
	\begin{axis}[
	axis background/.style={fill=white},
	axis line style={black},
	tick align=outside,
	tick pos=left,
	x grid style={white!89.80392156862746!black},
	xlabel={Steps},
	xmajorgrids,
	xmin=-4.95, xmax=103.95,
	xtick style={color=white!33.33333333333333!black},
	y grid style={white!89.80392156862746!black},
	ylabel={Expected Discounted Reward},
	ymajorgrids,
	ymin=-0.0276098847900357, ymax=0.75,
	ytick style={color=white!33.33333333333333!black},
	legend style={draw=black, fill=white},
	]
	\addplot [line width=1.64pt, orange]
	table {%
		0 0
		1 0.0055391606874764
		2 0.0101741468533874
		3 0.0162445772439241
		4 0.0211450643837452
		5 0.0253632310777903
		6 0.031331978738308
		7 0.0365632697939873
		8 0.0417822897434235
		9 0.0563660226762295
		10 0.0617316029965878
		11 0.0725953802466393
		12 0.129080608487129
		13 0.171923384070396
		14 0.226868405938148
		15 0.267122447490692
		16 0.297746211290359
		17 0.342760384082794
		18 0.370725184679031
		19 0.388545960187912
		20 0.391941159963608
		21 0.393272280693054
		22 0.402005463838577
		23 0.412633329629898
		24 0.421711534261703
		25 0.421922028064728
		26 0.424282163381577
		27 0.434874624013901
		28 0.428990989923477
		29 0.433142304420471
		30 0.42646923661232
		31 0.431045889854431
		32 0.427598893642426
		33 0.427974909543991
		34 0.429705768823624
		35 0.433334589004517
		36 0.429769903421402
		37 0.427119135856628
		38 0.443396806716919
		39 0.438816249370575
		40 0.44710648059845
		41 0.454330325126648
		42 0.45017409324646
		43 0.460150241851807
		44 0.451624989509583
		45 0.449834525585175
		46 0.441674917936325
		47 0.454279690980911
		48 0.45059335231781
		49 0.443054050207138
		50 0.444962441921234
		51 0.446405529975891
		52 0.438038855791092
		53 0.434500157833099
		54 0.435018599033356
		55 0.438794016838074
		56 0.439934104681015
		57 0.445620119571686
		58 0.440281897783279
		59 0.435386747121811
		60 0.44429811835289
		61 0.458437383174896
		62 0.460712105035782
		63 0.467985689640045
		64 0.46846804022789
		65 0.465772777795792
		66 0.45034196972847
		67 0.451332092285156
		68 0.441737711429596
		69 0.429723739624023
		70 0.438534170389175
		71 0.438204497098923
		72 0.444971889257431
		73 0.449167430400848
		74 0.452163755893707
		75 0.463798433542252
		76 0.45759791135788
		77 0.456379622220993
		78 0.455846726894379
		79 0.453819185495377
		80 0.456498682498932
		81 0.450186789035797
		82 0.444735765457153
		83 0.441361576318741
		84 0.445408254861832
		85 0.452753156423569
		86 0.450239270925522
		87 0.453572809696198
		88 0.456322222948074
		89 0.449536442756653
		90 0.453307777643204
		91 0.46086797118187
		92 0.468745410442352
		93 0.467181593179703
		94 0.465776354074478
		95 0.473481416702271
		96 0.467015326023102
		97 0.475312620401382
		98 0.475285083055496
		99 0.469768226146698
	};
	\addplot [line width=1.64pt, oxford_blue]
	table {%
		0 0
		1 0.018956549741187
		2 0.00113102735453617
		3 0.00169743990317843
		4 0.016709977562589
		5 0.0147193997813705
		6 0.0133946080288044
		7 0.00616272915178288
		8 0.0121188833135692
		9 0.0121360346728168
		10 0.0116240803422401
		11 0.00316765740509611
		12 0.00861339280582537
		13 0.00787063640410743
		14 0.0144602416247493
		15 0.0198963912589949
		16 0.0189879094618649
		17 0.0108835409485864
		18 0.00889708377451707
		19 0.00536481483298656
		20 0.000718486587857152
		21 0.00054889714181638
		22 0.00929787724194624
		23 0.00636930255707355
		24 0.00760029843801423
		25 0.0178357891565657
		26 0.0105150553829206
		27 0.01121020722053
		28 0.00472246814230124
		29 0.000477161582815644
		30 0.0065028585752232
		31 0.00273394785972933
		32 0.010204476916744
		33 0.019973671363851
		34 0.0134895939469174
		35 0.00363686993646289
		36 0.0178714307316598
		37 0.0159351984284328
		38 0.0146880338378796
		39 0.0181318729979512
		40 0.0152577096766614
		41 0.0157949527492353
		42 0.00707573955683207
		43 0.0196195314614425
		44 0.0192380187579645
		45 0.00322369306608038
		46 0.0150800814330374
		47 0.0143030179647491
		48 0.00922813395483955
		49 0.0106071143224689
		50 0.00980027843700383
		51 0.0184966414418914
		52 0.0100168212526131
		53 0.0166304897958362
		54 0.00707848409737432
		55 0.0176570183716251
		56 0.0179940117751325
		57 0.00922024329763275
		58 0.0113541014084049
		59 0.0184066087838386
		60 0.0144754590774404
		61 0.0097321710972317
		62 0.00443622021982022
		63 0.00649334487537796
		64 0.0139914327614049
		65 0.00332139370988252
		66 0.0181588099325219
		67 0.00536275025799633
		68 0.0182275567173609
		69 0.00619126249898921
		70 0.0191472342311232
		71 0.0141241161273521
		72 0.0100849763396664
		73 0.010354955122971
		74 0.0130282879793358
		75 0.0117588942358897
		76 0.00623688649102
		77 0.00415636949075854
		78 0.0102378331671057
		79 0.0186830871826756
		80 0.0124653017345174
		81 0.00150750738148091
		82 0.0164079998942403
		83 0.014518985749546
		84 0.0181530724190264
		85 0.00382805466608235
		86 0.0148956544855471
		87 0.00117517792797311
		88 0.013058198548691
		89 0.00546199464674299
		90 0.00453233058489526
		91 0.0175098234289648
		92 0.0021253196529105
		93 0.0104472533071786
		94 0.0170788601436974
		95 0.00489663955938034
		96 0.00420957877391293
		97 0.0176116351873256
		98 0.00845835296779392
		99 0.0143392219780995
	};
	\addplot [line width=1.64pt, blue]
	table {%
		0 0.48
		1 0.48
		2 0.48
		3 0.48
		4 0.48
		5 0.48
		6 0.48
		7 0.48
		8 0.48
		9 0.48
		10 0.48
		11 0.48
		12 0.48
		13 0.48
		14 0.48
		15 0.48
		16 0.48
		17 0.48
		18 0.48
		19 0.48
		20 0.48
		21 0.48
		22 0.48
		23 0.48
		24 0.48
		25 0.48
		26 0.48
		27 0.48
		28 0.48
		29 0.48
		30 0.48
		31 0.48
		32 0.48
		33 0.48
		34 0.48
		35 0.48
		36 0.48
		37 0.48
		38 0.48
		39 0.48
		40 0.48
		41 0.48
		42 0.48
		43 0.48
		44 0.48
		45 0.48
		46 0.48
		47 0.48
		48 0.48
		49 0.48
		50 0.48
		51 0.48
		52 0.48
		53 0.48
		54 0.48
		55 0.48
		56 0.48
		57 0.48
		58 0.48
		59 0.48
		60 0.48
		61 0.48
		62 0.48
		63 0.48
		64 0.48
		65 0.48
		66 0.48
		67 0.48
		68 0.48
		69 0.48
		70 0.48
		71 0.48
		72 0.48
		73 0.48
		74 0.48
		75 0.48
		76 0.48
		77 0.48
		78 0.48
		79 0.48
		80 0.48
		81 0.48
		82 0.48
		83 0.48
		84 0.48
		85 0.48
		86 0.48
		87 0.48
		88 0.48
		89 0.48
		90 0.48
		91 0.48
		92 0.48
		93 0.48
		94 0.48
		95 0.48
		96 0.48
		97 0.48
		98 0.48
		99 0.48
	};
	\legend{DeepSynth, DQN, Optimal}
	\end{axis}
	\end{tikzpicture}
\caption{Minecraft Task 3 Experiment: (a)~Training progress with four hybrid deep NFQ modules, (b)~Training progress with DeepSynth and DQN on the same training set $\mathcal{E}$. 
%The expected return is at $s_0=[4,4]$.
%\vspace{-1.5mm}}
}
\label{fig:result_3}
\end{figure}
\fi

%As discussed before, we let the agent randomly explore the environment to
%find possible rewards. Each episode of exploration starts with the agent
%initialised. Every time we see a reward, e.g.~$R_i$, we save the observed
%trace in the buffer under the task~$i$. The \emph{Synth} box then outputs a
%DFA for each of the discovered rewards. This means that even a single
%occurrence of task completion is enough for our framework to find a policy
%that accomplishes this task.
%The resulting DFA is then employed to guide the learning process. 
%Note that there may be a number of ways to accomplish a particular task in the synthesised DFAs. This phenomenon however causes no harm to the learning since there is only one valid way to receive the extrinsic reward. Hence, once the extrinsic reward is back-propagated the non-optimal options automatically fall out.
%Additionally, since the initial position during the training is random, once the training is done at any given state the agent is able to find the optimal policy to satisfy the property.
%\subsection{Results}
The training progress for Montezuma's Revenge and Task~3 in Minecraft is
plotted in Fig.~\ref{fig:montezuma_results} and Fig.~\ref{fig:result_3}.  In
Fig.~\ref{fig:result_3}.a the orange line gives the loss for the very first deep net
associated to the initial state of the DFA, the red and blue ones are of the
intermediate states in the DFA and the green line is associated to the final
state.  This shows an efficient back-propagation of extrinsic reward from
the final high-level state to the initial state, namely once the last deep
net converges the expected reward is back-propagated to the second and so
on.  Each NFQ module has $2$ hidden layers and $128$ ReLUs. Note that there may be a number of ways to accomplish a particular task in
the synthesised DFAs.  This, however, causes no harm since when
the extrinsic reward is back-propagated, the non-optimal options
fall out.  
\if\doctype2Further, emphasise that if the task DFA, in any of
the experiments, is known or even partially known a~priori, then the Synth
step can exploit this partial automaton by incrementally adding any new
labels or subtask sequences discovered during exploration. As an example,
if the automaton in Fig.~\ref{fig:montezuma_dfa}.b is given initially, the
agent is able to utilise the semantic correlation of the objects to
facilitate its explorations and find the key faster.\fi

%in each layer, and the training is done using the Adam optimizer with a discount factor of $0.95$. 

%After the training, by starting from any initial point in the crafting environment (Fig.~\ref{minecraft}), the agent is able to accomplish Task 1 and Task 3 with 100\% success rate. Further, each trained neural nets can be individually employed to accomplish any arbitrary event such as grass, wood, etc.~in transfer learning scenarios.

\section{Conclusions}

We have proposed a fully-unsupervised approach for training deep RL
agents when the reward is extremely sparse and non-Markovian. 
We~\emph{automatically} infer a high-level structure from observed
exploration traces using automata synthesis.
%
%This allows us to recall how to achieve any high-level task, even when it has been observed only once. This high-level structure is then synchronised with a hybrid deep neural fitted $Q$-iteration to convert the reward into a Markovian reward and also fill in low-level policy generation. 
%
%
%Finally, we would like to emphasise that DeepSynth is the first deep RL architecture to synthesise policies for unknown non-Markovian sequential objectives over continuous-state MDPs. However, note that the actual satisfaction probability can not be computed when the MDP has uncountably infinite states.
%
The inferred automaton is a formal, un-grounded, human-interpretable
representation of a complex task and its steps.  We showed that we are able
to efficiently learn policies that achieve complex high-level objectives
using fewer training samples as compared to alternative algorithms.
Owing to the modular structure of the automaton, the overall task can be
segmented into easy Markovian sub-tasks.  Therefore, any segment of the
proposed network that is associated with a sub-task can be used as a
separate trained module in transfer learning.
Another major contribution of the proposed method is that in problems where
domain knowledge is available, this knowledge can be easily encoded as an automaton to
guide learning.  This enables the agent to solve complex tasks and saves the
agent from an exhaustive exploration in the beginning.

% no matter how rare the required sequence of actions, they can still be delineated abstractly in terms of an automaton over high-level objectives that can guide deep RL.
%To the best of our knowledge, our work is also the first to enable deep RL to synthesise policies for unknown non-Markovian sequential objectives over \emph{continuous-state} MDPs. Many real world problems require actions to be taken in response to high-dimensional and real-valued state observations. It is also well-known that discretisation schemes generally suffer from the trade off between accuracy and curse of dimensionality~\cite{lcnfq}, and hence it is more efficient to treat the infinite state space directly.

\clearpage

\if\doctype0
\bibliography{paper_extended_bib} 
\fi

\if\doctype1
\bibliography{paper_extended_bib} 
\fi

\if\doctype2
\bibliography{paper_extended_bib} 
\fi

\if\doctype3
\section*{Acknowledgements}

The authors would like to thank Hadrien Pouget for interesting discussions
and the anonymous reviewers for their insightful suggestions.
This work was supported by a grant from the UK NCSC, and Balliol College, Jason Hu scholarship.

\bibliography{paper_limited_bib}

\begin{thebibliography}{83}
\providecommand{\natexlab}[1]{#1}
\providecommand{\url}[1]{\texttt{#1}}
\providecommand{\urlprefix}{URL }
\expandafter\ifx\csname urlstyle\endcsname\relax
  \providecommand{\doi}[1]{doi:\discretionary{}{}{}#1}\else
  \providecommand{\doi}{doi:\discretionary{}{}{}\begingroup
  \urlstyle{rm}\Url}\fi

\bibitem[{Abbeel et~al.(2007)Abbeel, Coates, Quigley, and Ng}]{ng}
Abbeel, P.; Coates, A.; Quigley, M.; and Ng, A.~Y. 2007.
\newblock An Application of Reinforcement Learning to Aerobatic Helicopter
  Flight.
\newblock In \emph{NeurIPS}, 1--8. MIT Press.

\bibitem[{Andreas, Klein, and Levine(2017)}]{pol-sketch}
Andreas, J.; Klein, D.; and Levine, S. 2017.
\newblock Modular Multitask Reinforcement Learning with Policy Sketches.
\newblock In \emph{ICML}, volume~70, 166--175.

\bibitem[{Angluin(1987)}]{lstar}
Angluin, D. 1987.
\newblock Learning Regular Sets from Queries and Counterexamples.
\newblock \emph{Inf. Comput.} 75(2): 87--106.

\bibitem[{Bacon, Harb, and Precup(2017)}]{bacon2017option}
Bacon, P.-L.; Harb, J.; and Precup, D. 2017.
\newblock The Option-critic Architecture.
\newblock In \emph{AAAI}.

\bibitem[{Bellemare et~al.(2013)Bellemare, Naddaf, Veness, and Bowling}]{ale}
Bellemare, M.~G.; Naddaf, Y.; Veness, J.; and Bowling, M. 2013.
\newblock The Arcade Learning Environment: An Evaluation Platform for General
  Agents.
\newblock \emph{Artificial Intelligence Research} 47: 253--279.

\bibitem[{Berlyne(1960)}]{intrinsic2}
Berlyne, D.~E. 1960.
\newblock \emph{Conflict, Arousal, and Curiosity}.
\newblock McGraw-Hill Book Company.

\bibitem[{Bertsekas and Shreve(2004)}]{shreve}
Bertsekas, D.~P.; and Shreve, S. 2004.
\newblock \emph{Stochastic Optimal Control: The Discrete-time Case}.
\newblock Athena Scientific.

\bibitem[{Bertsekas and Tsitsiklis(1996)}]{NDP}
Bertsekas, D.~P.; and Tsitsiklis, J.~N. 1996.
\newblock \emph{Neuro-dynamic Programming}, volume~1.
\newblock Athena Scientific.

\bibitem[{Biermann and Feldman(1972{\natexlab{a}})}]{ktails}
Biermann, A.~W.; and Feldman, J.~A. 1972{\natexlab{a}}.
\newblock On the Synthesis of Finite-State Machines from Samples of Their
  Behavior.
\newblock \emph{IEEE Trans. Comput.} 21(6): 592--597.

\bibitem[{Biermann and
  Feldman(1972{\natexlab{b}})}]{Biermann:1972:SFM:1638603.1638997}
Biermann, A.~W.; and Feldman, J.~A. 1972{\natexlab{b}}.
\newblock On the Synthesis of Finite-State Machines from Samples of Their
  Behavior.
\newblock \emph{IEEE Trans. Comput.} 21(6): 592--597.

\bibitem[{Brafman, De~Giacomo, and Patrizi(2018)}]{nonmarkov}
Brafman, R.~I.; De~Giacomo, G.; and Patrizi, F. 2018.
\newblock {LTLf/LDLf} Non-{Markovian} Rewards.
\newblock In \emph{AAAI}.

\bibitem[{Brockman et~al.(2016)Brockman, Cheung, Pettersson, Schneider,
  Schulman, Tang, and Zaremba}]{gym}
Brockman, G.; Cheung, V.; Pettersson, L.; Schneider, J.; Schulman, J.; Tang,
  J.; and Zaremba, W. 2016.
\newblock {OpenAI} Gym.
\newblock \emph{arXiv} 1606.01540.

\bibitem[{{Buzhinsky} and {Vyatkin}(2017)}]{model_SAT}
{Buzhinsky}, I.; and {Vyatkin}, V. 2017.
\newblock Automatic Inference of Finite-State Plant Models From Traces and
  Temporal Properties.
\newblock \emph{IEEE Trans. Ind. Informat.} 13(4): 1521--1530.

\bibitem[{Buzhinsky and Vyatkin(2017)}]{Buzhinsky2017ModularPM}
Buzhinsky, I.; and Vyatkin, V. 2017.
\newblock Modular Plant Model Synthesis from Behavior Traces and Temporal
  Properties.
\newblock In \emph{Emerging Technologies and Factory Automation}, 1--7. IEEE.

\bibitem[{{Cai} et~al.(2021){Cai}, {Peng}, {Li}, {Gao}, and
  {Kan}}]{cai2020receding}
{Cai}, M.; {Peng}, H.; {Li}, Z.; {Gao}, H.; and {Kan}, Z. 2021.
\newblock Receding Horizon Control-Based Motion Planning With Partially
  Infeasible LTL Constraints.
\newblock \emph{IEEE Control Systems Letters} 5(4): 1279--1284.

\bibitem[{Camacho et~al.(2019)Camacho, Toro~Icarte, Klassen, Valenzano, and
  McIlraith}]{toro2}
Camacho, A.; Toro~Icarte, R.; Klassen, T.~Q.; Valenzano, R.; and McIlraith,
  S.~A. 2019.
\newblock {LTL and Beyond: Formal Languages for Reward Function Specification
  in Reinforcement Learning}.
\newblock In \emph{IJCAI}, 6065--6073.

\bibitem[{Chockler et~al.(2020)Chockler, Kesseli, Kroening, and
  Strichman}]{ckks2020}
Chockler, H.; Kesseli, P.; Kroening, D.; and Strichman, O. 2020.
\newblock Learning the Language of Software Errors.
\newblock \emph{Artificial Intelligence Research} 67: 881--903.

\bibitem[{Cook and Mitchell(1996)}]{DPLLvsTabu}
Cook, S.; and Mitchell, D. 1996.
\newblock Finding Hard Instances of the Satisfiability Problem: A Survey.
\newblock In \emph{Satisfiability Problem: Theory and Applications}.

\bibitem[{Csikszentmihalyi(1990)}]{intrinsic3}
Csikszentmihalyi, M. 1990.
\newblock \emph{Flow: The Psychology of Optimal Experience}.
\newblock Harper \& Row New York.

\bibitem[{Daniel, Neumann, and Peters(2012)}]{options_h_1}
Daniel, C.; Neumann, G.; and Peters, J. 2012.
\newblock Hierarchical relative entropy policy search.
\newblock In \emph{Artificial Intelligence and Statistics}, 273--281.

\bibitem[{Davis and Putnam(1960)}]{DPLL}
Davis, M.; and Putnam, H. 1960.
\newblock A Computing Procedure for Quantification Theory.
\newblock \emph{J. ACM} 7(3): 201–215.

\bibitem[{De~Giacomo et~al.(2020)De~Giacomo, Favorito, Iocchi, and
  Patrizi}]{bolts}
De~Giacomo, G.; Favorito, M.; Iocchi, L.; and Patrizi, F. 2020.
\newblock Imitation Learning over Heterogeneous Agents with Restraining Bolts.
\newblock In \emph{International Conference on Automated Planning and
  Scheduling}, 517--521.

\bibitem[{De~Giacomo et~al.(2019)De~Giacomo, Iocchi, Favorito, and
  Patrizi}]{nonmarkov2}
De~Giacomo, G.; Iocchi, L.; Favorito, M.; and Patrizi, F. 2019.
\newblock Foundations for Restraining Bolts: Reinforcement Learning with
  {LTLf/LDLf} Restraining Specifications.
\newblock In \emph{International Conference on Automated Planning and
  Scheduling}, volume~29, 128--136.

\bibitem[{Ecoffet et~al.(2021)Ecoffet, Huizinga, Lehman, Stanley, and
  Clune}]{goexplore}
Ecoffet, A.; Huizinga, J.; Lehman, J.; Stanley, K.~O.; and Clune, J. 2021.
\newblock First Return, Then Explore.
\newblock \emph{Nature} 590(7847): 580--586.

\bibitem[{Fu and Topcu(2014)}]{topku}
Fu, J.; and Topcu, U. 2014.
\newblock Probably Approximately Correct {MDP} Learning and Control With
  Temporal Logic Constraints.
\newblock In \emph{Robotics: Science and Systems X}.

\bibitem[{Fulton and Platzer(2018)}]{fulton3}
Fulton, N.; and Platzer, A. 2018.
\newblock Safe Reinforcement Learning via Formal Methods: Toward Safe Control
  Through Proof and Learning.
\newblock In \emph{AAAI}, 6485--6492.

\bibitem[{Furelos-Blanco et~al.(2020)Furelos-Blanco, Law, Russo, Broda, and
  Jonsson}]{furelos2020induction}
Furelos-Blanco, D.; Law, M.; Russo, A.; Broda, K.; and Jonsson, A. 2020.
\newblock Induction of Subgoal Automata for Reinforcement Learning.
\newblock In \emph{AAAI}, 3890--3897.

\bibitem[{Gaon and Brafman(2020)}]{gaon2020reinforcement}
Gaon, M.; and Brafman, R. 2020.
\newblock Reinforcement Learning with Non-Markovian Rewards.
\newblock In \emph{Proceedings of the AAAI Conference on Artificial
  Intelligence}, volume~34, 3980--3987.

\bibitem[{Glover and Laguna(1998)}]{Tabu}
Glover, F.; and Laguna, M. 1998.
\newblock Tabu Search.
\newblock In \emph{Handbook of Combinatorial Optimization}, volume 1--3,
  2093--2229. Springer.

\bibitem[{Gold(1978)}]{Gold1978ComplexityOA}
Gold, E.~M. 1978.
\newblock Complexity of Automaton Identification from Given Data.
\newblock \emph{Information and Control} 37: 302--320.

\bibitem[{Grze{\'s}(2017)}]{reward_shaping}
Grze{\'s}, M. 2017.
\newblock Reward Shaping in Episodic Reinforcement Learning.
\newblock In \emph{Autonomous Agents and Multiagent Systems}, 565--573.
  International Foundation for Autonomous Agents and Multiagent Systems.

\bibitem[{Gulwani(2012)}]{Gulwani2012SynthesisFE}
Gulwani, S. 2012.
\newblock Synthesis from Examples.
\newblock In \emph{WAMBSE}.

\bibitem[{Hahn et~al.(2019)Hahn, Perez, Schewe, Somenzi, Trivedi, and
  Wojtczak}]{hahn}
Hahn, E.~M.; Perez, M.; Schewe, S.; Somenzi, F.; Trivedi, A.; and Wojtczak, D.
  2019.
\newblock Omega-Regular Objectives in Model-Free Reinforcement Learning.
\newblock In \emph{TACAS}, 395--412. Springer.

\bibitem[{Hasanbeig, Abate, and Kroening(2018)}]{arxiv}
Hasanbeig, M.; Abate, A.; and Kroening, D. 2018.
\newblock Logically-Constrained Reinforcement Learning.
\newblock \emph{arXiv} 1801.08099.

\bibitem[{Hasanbeig, Abate, and Kroening(2019{\natexlab{a}})}]{lcrl_j}
Hasanbeig, M.; Abate, A.; and Kroening, D. 2019{\natexlab{a}}.
\newblock Certified Reinforcement Learning with Logic Guidance.
\newblock \emph{arXiv} 1902.00778.

\bibitem[{Hasanbeig, Abate, and Kroening(2019{\natexlab{b}})}]{lcnfq}
Hasanbeig, M.; Abate, A.; and Kroening, D. 2019{\natexlab{b}}.
\newblock Logically-Constrained Neural Fitted {Q}-Iteration.
\newblock In \emph{Autonomous Agents and Multiagent Systems}, 2012--2014.
  International Foundation for Autonomous Agents and Multiagent Systems.

\bibitem[{Hasanbeig, Abate, and Kroening(2020)}]{cautiousRL}
Hasanbeig, M.; Abate, A.; and Kroening, D. 2020.
\newblock Cautious Reinforcement Learning with Logical Constraints.
\newblock In \emph{Autonomous Agents and Multiagent Systems}, 483--491.
  International Foundation for Autonomous Agents and Multiagent Systems.

\bibitem[{Hasanbeig et~al.(2019{\natexlab{a}})Hasanbeig, Kantaros, Abate,
  Kroening, Pappas, and Lee}]{plmdp}
Hasanbeig, M.; Kantaros, Y.; Abate, A.; Kroening, D.; Pappas, G.~J.; and Lee,
  I. 2019{\natexlab{a}}.
\newblock {Reinforcement Learning for Temporal Logic Control Synthesis with
  Probabilistic Satisfaction Guarantees}.
\newblock In \emph{CDC}, 5338--5343. IEEE.

\bibitem[{Hasanbeig, Kroening, and Abate(2020)}]{hasanbeig2020deep}
Hasanbeig, M.; Kroening, D.; and Abate, A. 2020.
\newblock Deep Reinforcement Learning with Temporal Logics.
\newblock In \emph{Formal Modeling and Analysis of Timed Systems}, volume 12288
  of \emph{LNCS}, 1--22. Springer.

\bibitem[{Hasanbeig et~al.(2019{\natexlab{b}})Hasanbeig, Yogananda~Jeppu,
  Abate, Melham, and Kroening}]{arxiv_deepsynth}
Hasanbeig, M.; Yogananda~Jeppu, N.; Abate, A.; Melham, T.; and Kroening, D.
  2019{\natexlab{b}}.
\newblock {DeepSynth}: Program Synthesis for Automatic Task Segmentation in
  Deep Reinforcement Learning [Extended Version].
\newblock \emph{arXiv} 1911.10244.

\bibitem[{Heule and Verwer(2013)}]{Heule2013}
Heule, M. J.~H.; and Verwer, S. 2013.
\newblock Software Model Synthesis Using Satisfiability Solvers.
\newblock \emph{Empirical Software Engineering} 18(4): 825--856.

\bibitem[{Hwang et~al.(2019)Hwang, Yu, Shi, Collins, Yang, Zhang, and
  Chen}]{obj_det_2}
Hwang, J.-J.; Yu, S.~X.; Shi, J.; Collins, M.~D.; Yang, T.-J.; Zhang, X.; and
  Chen, L.-C. 2019.
\newblock {SegSort}: Segmentation by Discriminative Sorting of Segments.
\newblock In \emph{ICCV}, 7334--7344.

\bibitem[{Jeppu(2020)}]{t2m}
Jeppu, N.~Y. 2020.
\newblock \emph{Trace2Model Github repository}.
\newblock \urlprefix\url{https://github.com/natasha-jeppu/Trace2Model}.

\bibitem[{Jeppu et~al.(2020)Jeppu, Melham, Kroening, and O'Leary}]{synth}
Jeppu, N.~Y.; Melham, T.; Kroening, D.; and O'Leary, J. 2020.
\newblock {Learning Concise Models from Long Execution Traces}.
\newblock In \emph{Design Automation Conference}, 1--6. ACM/IEEE.

\bibitem[{Ji, Henriques, and Vedaldi(2019)}]{obj_det}
Ji, X.; Henriques, J.~F.; and Vedaldi, A. 2019.
\newblock Invariant Information Clustering for Unsupervised Image
  Classification and Segmentation.
\newblock In \emph{ICCV}, 9865--9874.

\bibitem[{Kazemi and Soudjani(2020)}]{kazemi2020formal}
Kazemi, M.; and Soudjani, S. 2020.
\newblock Formal Policy Synthesis for Continuous-Space Systems via
  Reinforcement Learning.
\newblock \emph{arXiv} 2005.01319.

\bibitem[{Kearns and Singh(2002)}]{options_1}
Kearns, M.; and Singh, S. 2002.
\newblock Near-optimal Reinforcement Learning in Polynomial Time.
\newblock \emph{Machine Learning} 49(2-3): 209--232.

\bibitem[{Kingma and Ba(2015)}]{adam}
Kingma, D.~P.; and Ba, J. 2015.
\newblock Adam: A Method for Stochastic Optimization.
\newblock In \emph{International Conference on Learning Representations}.

\bibitem[{Koul, Fern, and Greydanus(2019)}]{koul2018learning}
Koul, A.; Fern, A.; and Greydanus, S. 2019.
\newblock Learning Finite State Representations of Recurrent Policy Networks.
\newblock In \emph{International Conference on Learning Representations}.

\bibitem[{Krishnan et~al.(1995)Krishnan, Puri, Brayton, and
  Varaiya}]{rabin_game}
Krishnan, S.~C.; Puri, A.; Brayton, R.~K.; and Varaiya, P.~P. 1995.
\newblock The {Rabin} Index and Chain Automata, with Applications to Automata
  and Games.
\newblock In \emph{Computer-Aided Verification (CAV)}, LNCS, 253--266.
  Springer.

\bibitem[{Kulkarni et~al.(2016)Kulkarni, Narasimhan, Saeedi, and
  Tenenbaum}]{kulkarni2016hierarchical}
Kulkarni, T.~D.; Narasimhan, K.; Saeedi, A.; and Tenenbaum, J. 2016.
\newblock Hierarchical Deep Reinforcement Learning: Integrating Temporal
  Abstraction and Intrinsic Motivation.
\newblock In \emph{NeurIPS}, 3675--3683.

\bibitem[{Lang, Pearlmutter, and Price(1998)}]{edsm}
Lang, K.~J.; Pearlmutter, B.~A.; and Price, R.~A. 1998.
\newblock Results of the {Abbadingo} {One} {DFA} Learning Competition and a New
  Evidence-driven State Merging Algorithm.
\newblock In \emph{Grammatical Inference}, volume 1433 of \emph{LNCS}, 1--12.
  Springer.

\bibitem[{Lavaei et~al.(2020)Lavaei, Somenzi, Soudjani, Trivedi, and
  Zamani}]{lavaei2020formal}
Lavaei, A.; Somenzi, F.; Soudjani, S.; Trivedi, A.; and Zamani, M. 2020.
\newblock Formal Controller Synthesis for Continuous-space MDPs via Model-free
  Reinforcement Learning.
\newblock In \emph{International Conference on Cyber-Physical Systems},
  98--107. IEEE.

\bibitem[{Liu et~al.(2019)Liu, Wei, Sharpnack, and Owens}]{obj_det_o}
Liu, W.; Wei, L.; Sharpnack, J.; and Owens, J.~D. 2019.
\newblock Unsupervised Object Segmentation with Explicit Localization Module.
\newblock \emph{arXiv} 1911.09228.

\bibitem[{Mao et~al.(2016)Mao, Alizadeh, Menache, and Kandula}]{rlresource}
Mao, H.; Alizadeh, M.; Menache, I.; and Kandula, S. 2016.
\newblock Resource Management with Deep Reinforcement Learning.
\newblock In \emph{ACM Workshop on Networks}, 50--56. ACM.

\bibitem[{Memarian et~al.(2020)Memarian, Xu, Wu, Wen, and
  Topcu}]{memarian2020active}
Memarian, F.; Xu, Z.; Wu, B.; Wen, M.; and Topcu, U. 2020.
\newblock Active Task-Inference-Guided Deep Inverse Reinforcement Learning.
\newblock In \emph{59th {IEEE} Conference on Decision and Control, {CDC}},
  1932--1938. {IEEE}.

\bibitem[{Mnih et~al.(2015)}]{deepql}
Mnih, V.; et~al. 2015.
\newblock Human-level Control Through Deep Reinforcement Learning.
\newblock \emph{Nature} 518(7540): 529--533.

\bibitem[{Ng, Harada, and Russell(1999)}]{reward_shaping_2}
Ng, A.~Y.; Harada, D.; and Russell, S. 1999.
\newblock Policy Invariance Under Reward Transformations: Theory and
  Application to Reward Shaping.
\newblock In \emph{ICML}, volume~99, 278--287.

\bibitem[{Polydoros and Nalpantidis(2017)}]{polydoros2017survey}
Polydoros, A.~S.; and Nalpantidis, L. 2017.
\newblock Survey of Model-based Reinforcement Learning: Applications on
  Robotics.
\newblock \emph{Journal of Intelligent \& Robotic Systems} 86(2): 153--173.

\bibitem[{Precup(2001)}]{precup}
Precup, D. 2001.
\newblock \emph{Temporal Abstraction in Reinforcement Learning}.
\newblock Ph.D. thesis, University of Massachusetts Amherst.

\bibitem[{Rens and Raskin(2020)}]{rens2020learning}
Rens, G.; and Raskin, J.-F. 2020.
\newblock Learning Non-Markovian Reward Models in {MDPs}.
\newblock \emph{arXiv} 2001.09293.

\bibitem[{Rens et~al.(2020)Rens, Raskin, Reynouad, and Marra}]{rens2020online}
Rens, G.; Raskin, J.-F.; Reynouad, R.; and Marra, G. 2020.
\newblock Online Learning of Non-{Markovian} Reward Models.
\newblock \emph{arXiv} 2009.12600.

\bibitem[{Riedmiller(2005)}]{nfq}
Riedmiller, M. 2005.
\newblock Neural Fitted {Q} iteration -- First Experiences with a Data
  Efficient Neural Reinforcement Learning Method.
\newblock In \emph{ECML}, volume 3720, 317--328. Springer.

\bibitem[{Ryan and Deci(2000)}]{intrinsic}
Ryan, R.~M.; and Deci, E.~L. 2000.
\newblock Intrinsic and Extrinsic Motivations: Classic Definitions and New
  Directions.
\newblock \emph{Contemporary Educational Psychology} 25(1): 54--67.

\bibitem[{Sadigh et~al.(2014)Sadigh, Kim, Coogan, Sastry, and Seshia}]{dorsa}
Sadigh, D.; Kim, E.~S.; Coogan, S.; Sastry, S.~S.; and Seshia, S.~A. 2014.
\newblock A Learning Based Approach to Control Synthesis of {Markov} Decision
  Processes for Linear Temporal Logic Specifications.
\newblock In \emph{Conference on Decision and Control}, 1091--1096. IEEE.

\bibitem[{Silver et~al.(2016)Silver, Huang, Maddison, Guez, Sifre, van~den
  Driessche, Schrittwieser, Antonoglou, Panneershelvam, Lanctot, Dieleman,
  Grewe, Nham, Kalchbrenner, Sutskever, Lillicrap, Leach, Kavukcuoglu, Graepel,
  and Hassabis}]{alphago}
Silver, D.; Huang, A.; Maddison, C.~J.; Guez, A.; Sifre, L.; van~den Driessche,
  G.; Schrittwieser, J.; Antonoglou, I.; Panneershelvam, V.; Lanctot, M.;
  Dieleman, S.; Grewe, D.; Nham, J.; Kalchbrenner, N.; Sutskever, I.;
  Lillicrap, T.; Leach, M.; Kavukcuoglu, K.; Graepel, T.; and Hassabis, D.
  2016.
\newblock Mastering the Game of {Go} with Deep Neural Networks and Tree Search.
\newblock \emph{Nature} 529: 484--503.

\bibitem[{Sutton and Barto(1998)}]{sutton}
Sutton, R.~S.; and Barto, A.~G. 1998.
\newblock \emph{Reinforcement Learning: An Introduction}, volume~1.
\newblock MIT Press Cambridge.

\bibitem[{Toro~Icarte et~al.(2018)Toro~Icarte, Klassen, Valenzano, and
  McIlraith}]{toro}
Toro~Icarte, R.; Klassen, T.~Q.; Valenzano, R.; and McIlraith, S.~A. 2018.
\newblock Teaching Multiple Tasks to an {RL} Agent using {LTL}.
\newblock In \emph{Autonomous Agents and Multiagent Systems}, 452--461.

\bibitem[{Toro~Icarte et~al.(2019)Toro~Icarte, Waldie, Klassen, Valenzano,
  Castro, and McIlraith}]{toronto}
Toro~Icarte, R.; Waldie, E.; Klassen, T.; Valenzano, R.; Castro, M.; and
  McIlraith, S. 2019.
\newblock Learning Reward Machines for Partially Observable Reinforcement
  Learning.
\newblock In \emph{NeurIPS}, 15497--15508.

\bibitem[{Ulyantsev, Buzhinsky, and Shalyto(2018)}]{exact_fsm}
Ulyantsev, V.; Buzhinsky, I.; and Shalyto, A. 2018.
\newblock Exact Finite-state Machine Identification from Scenarios and Temporal
  Properties.
\newblock \emph{International Journal on Software Tools for Technology
  Transfer} 20(1): 35--55.

\bibitem[{{Ulyantsev} and {Tsarev}(2011)}]{efsm_state_merge}
{Ulyantsev}, V.; and {Tsarev}, F. 2011.
\newblock Extended Finite-State Machine Induction Using {SAT}-Solver.
\newblock In \emph{International Conference on Machine Learning and
  Applications and Workshops}, 346--349.

\bibitem[{Vezhnevets et~al.(2016)Vezhnevets, Mnih, Osindero, Graves, Vinyals,
  Agapiou et~al.}]{options_h_2}
Vezhnevets, A.; Mnih, V.; Osindero, S.; Graves, A.; Vinyals, O.; Agapiou, J.;
  et~al. 2016.
\newblock Strategic Attentive Writer for Learning Macro-actions.
\newblock In \emph{NeurIPS}, 3486--3494.

\bibitem[{Vezhnevets et~al.(2017)Vezhnevets, Osindero, Schaul, Heess,
  Jaderberg, Silver, and Kavukcuoglu}]{feudal}
Vezhnevets, A.~S.; Osindero, S.; Schaul, T.; Heess, N.; Jaderberg, M.; Silver,
  D.; and Kavukcuoglu, K. 2017.
\newblock {FeUdal} Networks for Hierarchical Reinforcement Learning.
\newblock In \emph{International Conference on Machine Learning}, 3540–3549.

\bibitem[{Vinyals et~al.(2019)Vinyals, Babuschkin, Czarnecki, Mathieu, Dudzik,
  Chung, Choi, Powell, Ewalds, Georgiev, Oh, Horgan, Kroiss, Danihelka, Huang,
  Sifre, Cai, Agapiou, Jaderberg, Vezhnevets, Leblond, Pohlen, Dalibard,
  Budden, Sulsky, Molloy, Paine, Gulcehre, Wang, Pfaff, Wu, Ring, Yogatama,
  W{\"u}nsch, McKinney, Smith, Schaul, Lillicrap, Kavukcuoglu, Hassabis, Apps,
  and Silver}]{alphastar}
Vinyals, O.; Babuschkin, I.; Czarnecki, W.~M.; Mathieu, M.; Dudzik, A.; Chung,
  J.; Choi, D.~H.; Powell, R.; Ewalds, T.; Georgiev, P.; Oh, J.; Horgan, D.;
  Kroiss, M.; Danihelka, I.; Huang, A.; Sifre, L.; Cai, T.; Agapiou, J.~P.;
  Jaderberg, M.; Vezhnevets, A.~S.; Leblond, R.; Pohlen, T.; Dalibard, V.;
  Budden, D.; Sulsky, Y.; Molloy, J.; Paine, T.~L.; Gulcehre, C.; Wang, Z.;
  Pfaff, T.; Wu, Y.; Ring, R.; Yogatama, D.; W{\"u}nsch, D.; McKinney, K.;
  Smith, O.; Schaul, T.; Lillicrap, T.; Kavukcuoglu, K.; Hassabis, D.; Apps,
  C.; and Silver, D. 2019.
\newblock Grandmaster Level in {StarCraft II} Using Multi-agent Reinforcement
  Learning.
\newblock \emph{Nature} 1--5.

\bibitem[{Walkinshaw(2018)}]{mint}
Walkinshaw, N. 2018.
\newblock \emph{MINT Framework Github Repository}.
\newblock \urlprefix\url{https://github.com/neilwalkinshaw/mintframework}.

\bibitem[{{Walkinshaw} and {Bogdanov}(2008)}]{state_merge}
{Walkinshaw}, N.; and {Bogdanov}, K. 2008.
\newblock Inferring Finite-State Models with Temporal Constraints.
\newblock In \emph{Automated Software Engineering}, 248--257. IEEE.

\bibitem[{Walkinshaw et~al.(2007)Walkinshaw, Bogdanov, Holcombe, and
  Salahuddin}]{Walkinshaw:2007:RES:1339262.1339495}
Walkinshaw, N.; Bogdanov, K.; Holcombe, M.; and Salahuddin, S. 2007.
\newblock Reverse Engineering State Machines by Interactive Grammar Inference.
\newblock In \emph{Working Conference on Reverse Engineering}, 209--218. IEEE.

\bibitem[{Walkinshaw, Taylor, and Derrick(2016)}]{Walkinshaw2016}
Walkinshaw, N.; Taylor, R.; and Derrick, J. 2016.
\newblock Inferring Extended Finite State Machine Models From Software
  Executions.
\newblock \emph{Empirical Software Engineering} 21(3): 811--853.

\bibitem[{Watkins and Dayan(1992)}]{watkins}
Watkins, C.~J.; and Dayan, P. 1992.
\newblock {Q}-learning.
\newblock \emph{Machine learning} 8(3-4): 279--292.

\bibitem[{Xu et~al.(2020)Xu, Gavran, Ahmad, Majumdar, Neider, Topcu, and
  Wu}]{xu2020joint}
Xu, Z.; Gavran, I.; Ahmad, Y.; Majumdar, R.; Neider, D.; Topcu, U.; and Wu, B.
  2020.
\newblock Joint Inference of Reward Machines and Policies for Reinforcement
  Learning.
\newblock In \emph{AAAI}, volume~30, 590--598.

\bibitem[{Yuan et~al.(2019)Yuan, Hasanbeig, Abate, and Kroening}]{deeplcrl}
Yuan, L.~Z.; Hasanbeig, M.; Abate, A.; and Kroening, D. 2019.
\newblock Modular Deep Reinforcement Learning with Temporal Logic
  Specifications.
\newblock \emph{arXiv} 1909.11591.

\bibitem[{Zheng and Yang(2021)}]{obj_det_3}
Zheng, Z.; and Yang, Y. 2021.
\newblock Rectifying Pseudo Label Learning via Uncertainty Estimation for
  Domain Adaptive Semantic Segmentation.
\newblock \emph{International Journal of Computer Vision} (to appear).

\bibitem[{Zhou et~al.(2017)}]{chemistry}
Zhou, Z.; et~al. 2017.
\newblock Optimizing Chemical Reactions with Deep Reinforcement Learning.
\newblock \emph{ACS Central Science} 3(12): 1337--1344.

\end{thebibliography}
\fi

\if\doctype2
\vfill
\clearpage

\fi
\end{document}
\fi